\documentclass[jair,twoside,11pt,theapa,,letterpaper]{article}
\usepackage{jair, theapa, rawfonts}
\usepackage{enumitem}
\jairheading{}{2023}{}{4/23}{Forthcomming}
\ShortHeadings{The Jiminy Advisor}
{Liao, Pardo, Slavkovik, \& van der Torre}
\firstpageno{1}
\usepackage{hhline}
\usepackage{url}
\usepackage{verbatim}
\usepackage{subcaption}
\usepackage{wrapfig}
\usepackage{caption}
\usepackage{subcaption}
\usepackage{xcolor} 

\usepackage[utf8]{inputenc}
\usepackage[T1]{fontenc}
\usepackage{graphicx}
\usepackage{amsmath}
\usepackage{amsthm}
\usepackage{amssymb}
\usepackage{mathtools}
\usepackage{microtype}
\usepackage{booktabs}
\usepackage{nicefrac}
\usepackage{tikz}
\usetikzlibrary{fit, backgrounds, matrix, arrows.meta}

\tikzset{
    Core/.style={draw,text width=2.5cm, minimum height=1.15cm,align=center},
    }

\tikzset{Oarg/.style={circle,draw,align=center,fill=white!55},
    }

\tikzset{OargG/.style={circle,draw,align=center,fill=gray!25},
    }
\tikzset{OargB/.style={circle,draw,align=center,fill=white!55,thick},
    }
\tikzset{OargGB/.style={circle,draw,align=center,fill=gray!25,thick},
    }

\usetikzlibrary{shapes.geometric}
\tikzset{Iarg/.style={regular polygon,
    regular polygon sides=3,draw,align=center,fill=white!55},
    }
\tikzset{IargG/.style={regular polygon,
    regular polygon sides=3,draw,align=center,fill=gray!25},
    }
\tikzset{IargGB/.style={regular polygon,
    regular polygon sides=3,draw,align=center,fill=gray!25, thick},
    }
\tikzset{Parg/.style={regular polygon,
    regular polygon sides=4,rotate=45,draw,align=center,fill=white!55},
    }    
\tikzset{PargG/.style={regular polygon,
    regular polygon sides=4,rotate=45,draw,align=center,fill=gray!25},
    }

\newtheorem{theorem}{Theorem}[section]
\newtheorem{definition}{Definition} 

\newtheorem{lemma}[theorem]{Lemma}
\newtheorem{proposition}[theorem]{Proposition}
\newtheorem{fact}[theorem]{Fact}
\theoremstyle{definition}
\newtheorem{example}[theorem]{Example}

\newcommand{\normsys}{\mathcal{N}}
\newcommand{\recommender}[1]{\mathcal{#1}^{\mathit{bcr}}}
\newcommand{\priex}[1]{{#1}^{\ast}}
\newcommand{\recommendation}{$\mathit{bcr}$-fragment}

\begin{document}

\title{The Jiminy Advisor:  \\ Moral Agreements among Stakeholders\\ based on Norms and Argumentation}

\author{\name Beishui Liao \email baiseliao@zju.edu.cn \\
       \addr Zheijang University, School of Philosophy,       Hangzhou, 310058 China;
       \addr The State Key Lab of Brain-Machine Intelligence,       Hangzhou, 310058 China
       \AND
       \name Pere Pardo \email pere.pardo@uni.lu \\
       \addr University of Luxembourg, Dept. of Computer Science, 	L-4364 Esch-sur-Alzette, Luxembourg
       \AND
       \name Marija Slavkovik \email marija.slavkovik@uib.no \\
       \addr University of Bergen, Dept. of Information Science and Media Studies, 5007 Bergen, Norway
       \AND 
       \name Leendert van der Torre \email leon.vandertorre@uni.lu \\
       \addr University of Luxembourg, Dept. of Computer Science, 
	L-4364 Esch-sur-Alzette, Luxembourg}

\newcommand{\DmadebyM}{\mbox{\textit{D is made by M}}}
\newcommand{\varDmadebyM}{w_1}
\newcommand{\DcollectsData}{\mbox{\textit{D collects data}}}
\newcommand{\varDcollectsData}{w_2}
\newcommand{\DfindsThreat}{\mbox{\textit{D finds a threat}}}
\newcommand{\varDfindsThreat}{w_3}
\newcommand{\MregisteredNorway}{\mbox{\textit{M is registered in Norway}}}
\newcommand{\varMregisteredNorway}{w_4}
\newcommand{\MbusinessNorway}{\mbox{\textit{M is a \textbf{business} in Norway}}}
\newcommand{\varMbusinessNorway}{i_1}
\newcommand{\lawCompliantM}{\mbox{\textit{M is \textbf{law} compliant}}}
\newcommand{\varlawCompliantM}{d_1}
\newcommand{\privacyProtectD}{\mbox{\textit{\textbf{protect} privacy}}}
\newcommand{\varprivacyProtectD}{d_2}
\newcommand{\reportThreatD}{\mbox{\textit{\textbf{report} threat}}}
\newcommand{\varreportThreatD}{d_3}
\newcommand{\toComplyGDPR}{\mbox{\textit{comply with the \textbf{GDPR}}}}
\newcommand{\vartoComplyGDPR}{a_1}
\newcommand{\toCollectDataNEP}{\mbox{\textit{\textbf{collect} data w.o.~permission}}}
\newcommand{\vartoCollectDataNEP}{a_2}

\maketitle

\begin{abstract}
An autonomous system is constructed by a manufacturer, operates in a society subject to norms and laws, and interacts with end users. All of these actors are stakeholders affected by the behavior of the autonomous system. We address the challenge of how the ethical views of such stakeholders can be integrated in the behavior of an autonomous system. We propose an ethical recommendation component called Jiminy which uses techniques from normative systems and formal argumentation to reach moral agreements among stakeholders.  A Jiminy represents the ethical views of each stakeholder by using normative systems, and has three ways of resolving moral dilemmas that involve the opinions of the stakeholders. First, the Jiminy considers how the arguments of the stakeholders relate to one another, which may already resolve the dilemma. Secondly, the Jiminy combines the normative systems of the stakeholders such that the combined expertise of the stakeholders may resolve the dilemma. Thirdly, and only if these two other methods have failed, the Jiminy uses context-sensitive rules to decide which of the stakeholders take preference over the others. At the abstract level, these three methods are characterized by adding arguments, adding attacks between arguments, and revising attacks between arguments. We show how a Jiminy can be used not only for ethical  reasoning and collaborative decision-making, but also to provide explanations about ethical behavior. 
\end{abstract}

%\begin{keyword}
%machine ethics \sep artificial morality \sep normative reasoning \sep formal argumentation \sep multiple stakeholders



%\end{keyword}

%\end{frontmatter}

\section{Introduction}

Artificial autonomous systems depend on human intervention to distinguish moral from immoral behavior.  Implicit ethical agents~\cite{Moor:2006} are ethically constrained from engaging in immoral behavior via rules set by human designers.  Explicit ethical agents~\cite{Moor:2006,Dyrkolbotn18} or agents with  functional morality~\cite[Chapter 2]{wallachBook} are either able to make moral judgments themselves or are given guidelines or examples about what is good and bad.  In either case, the question arises: who decides which and whose morality the artificial autonomous systems ultimately upholds?  

It is immediately apparent that the persons and institutions that are affected by the moral behavior of an autonomous system should be given the opportunity to indicate their moral preferences as input into the behavior of that autonomous system~\cite{Baum:2020aa}.  There are, however, numerous {\em stakeholders} that satisfy the above definition of concerned entities~\cite{Baum:2020aa}. For example,  governmental regulators can determine which behavior is legal for the state in which the autonomous system is deployed.  The manufacturers, shareholders, designers and developers involved in building the autonomous system would be concerned not only with issues of liability, but also issues of representation---an autonomous system should uphold the image and values of its maker.  People interacting directly with the autonomous system should have a choice on certain aspects of its moral behavior, whether they are owners, users, or just share an environment with the system. It is easy to argue that it is wrong to select any of these stakeholders over others as being able to exclusively define what constitutes moral behavior for an autonomous system: such a system will end up with a regulation that benefits that stakeholder and misconstrues or underplays others' interests.    

Legal systems recognize only humans and corporations as persons and moral agents. The underlying assumption that everyone is human allows a great deal of flexibility when it comes to specifying and enforcing desirable behavior---not all desirable behaviors are specified and not all violations of law are meticulously prosecuted. Companies can build a system that reports on every violation of law committed by a user, but who would then want to buy such a totalitarian ``surveillance'' device?

People can have multiple roles when interacting with an autonomous system, and they can have different role-based moral preferences for the system. As pedestrians, they would prefer utilitarian cars that elect to run into a wall and kill its one passenger rather than kill several pedestrians, yet at the same time would surely prefer not to buy such a car~\cite{Shariff:2017aa}. Even if we somehow determine that the role of a pedestrian is more important than the role of a passenger, who would wish to buy a car that might kill one's own children while driving them to school?

We propose here that all the stakeholders' moral instructions should be included when deciding the moral behavior of an autonomous system. The problem then immediately becomes: {\em how should an autonomous system combine the ethical input of various stakeholders? }

In this article, the terms ``moral'' and ``ethical'' are used interchangeably. Let us imagine that each of the ``morality'' stakeholders are represented by an ``avatar'' in an artificial autonomous system. We refer to this artificial autonomous system as simply an ``agent''. The ``avatars'' form the ``moral council'' of the agent, acting like Jiminy Cricket in the story of Pinocchio. First, the stakeholders can indicate which situations are ethically sensitive and how to proceed in each situation. An agent makes such a decision by choosing from a set of available actions. If none of the stakeholders regard the situation as ethically sensitive in any way, then the agent can use its regular reasoning methods to select what to do. However, in an ethically sensitive situation, a Jiminy would be employed to produce a moral recommendation to the agent.

The first challenge in building the ``moral council'' is that the stakeholders may not be following the same ethical reasoning theory or any ethical theory at all. It is not sufficient that each stakeholder chimes in with a ``yes'' or ``no'' when the question of the morality of an action or of an action’s outcome is presented.

The second challenge is how to reach an agreement. Dilemmas and conflicts will arise when the inputs of the stakeholders are applied to a decision-making problem~\cite{Robinson21,Horty94}. We do not want to evaluate the morality of an action by majority rule.  Neither do we want to always put legal considerations above the image of the manufacturer, or give higher consideration to the personal input of the end user than the guidelines of regulatory bodies. Instead, we wish to have an engine that is able to take inputs from the different stakeholders and bring them into agreement.

The third challenge is explainability. Since the stakeholders are not necessarily aware of the input of other stakeholders, the decisions that the system ends up making need to be explained. This means that whatever solutions are used must be such that the artificial moral agent is able to explain its choices~\cite{AndersonA14} or that the choices should be formally verifiable~\cite{Ethical15}.

We  propose that normative systems~\cite{handbooknms} and formal argumentation~\cite{hofa} can be used to implement a  ``moral council'' for an artificial moral agent. With this approach, we can abstract  away from how a particular stakeholder has reached a  particular decision concerning the morality of an  action.  We model each stakeholder as a normative system that is then exploited as a source of arguments. An argument can be a statement regarding whether or not an action is moral, or it can be a reason why a particular action should be considered to be moral or immoral. Abstract argumentation allows us to build a system of attacking and supporting arguments that can be analyzed to determine which statements are supported and which are refuted in the system at a given time. Such a system can also generate  explanations of decisions using dialogue techniques. 

The main contributions of this article are the following:
\begin{enumerate}
\item  Within the field of machine ethics, we describe the first computational model that combines the ethical theories of multiple stakeholders in ethical decision making; 
\item  Within the field of structured argumentation, we describe the first model that resolves moral dilemmas arising from multiple normative systems.
\end{enumerate}

The article is structured as follows. 
In Section~\ref{architecture} we introduce the Jiminy moral advisor component, in Section~\ref{sec:norm} we discuss how to represent normative systems, and in Section~\ref{sec:arg} we study an argumentation system for one or multiple stakeholders. Section~\ref{sec:agreements} shows different ways to use argumentation to come to moral agreements between stakeholders. Section~\ref{sec:expl} adresses the explainability of the decisions recommended by the argumentation system. Section~\ref{interface} discusses different ways to embed a Jiminy in a device and turn it into an artificial ethical agent. The article concludes with a discussion of the related literature in Section~\ref{sec:relwork} and a summary of this article in Section~\ref{sec:summary}. An appendix at the end contains proofs of our results.

\section{The Jiminy Moral Advisor Component}\label{architecture}
We first consider the problem of how a multi-stakeholder ethical advisory component can be designed and integrated into an agent or an artificial autonomous system. We call this multi-stakeholder ethical advisory component a {\em Jiminy advisor}.  

The first problem we face is the problem of building the ``avatars'', one for each stakeholder. These are the sources of ``insight'' on what a Jiminy should advise in a given ethically sensitive situation.  Clearly, the stakeholders cannot be available in real time to give feedback to the avatar integrated in each instance of the agent. Rather, this avatar needs domain-specific information in advance about what the stakeholder considers to be ethically sensitive situations and what are its recommendations on what should be done when such situations arise. We  propose the use of normative systems~\cite{handbooknms} to model the stakeholders. A normative system describes how to evaluate actions in a system of agents and how to guide the behavior of those agents~\cite{alchourron91}. A norm is a formal description of desirable behavior, desirable action, or the desirable outcome of an action. Furthermore, normative systems can also be seen as rule-based systems (Definition \ref{def-normative-systems}) in which norms can be provided with reasons supporting their enforcement. By presenting norms that stipulate how to avoid immoral behavior, stakeholders also contribute standpoints or claims in order to characterize and help identify ethically sensitive situations. Every stakeholder is modeled with its own normative system in the Jiminy advisor.

The advantage of using  normative systems  is that it allows us to abstract away from the particular moral theory that a stakeholder upholds. This facilitates public scrutiny of any norm a stakeholder includes in its avatar. The immediate disadvantage of this approach is that the scope of ethically sensitive situations that the agent can handle by design is limited since these need to be predicted in advance by the stakeholders. However, this is not unusual for systems that regulate behavior---even people sometimes find out that what they have done is immoral after the fact. And even the law is not written to predict all future possible sources of danger to society.  The law is subject to interpretation by legal practitioners, and is amended as necessary when a new threat is recognized. In the same way, a normative system can be amended offline if a new ethically sensitive situation arises. 
 
It is clear that for any given situation,  different stakeholders would have different recommendations. As an example, consider a smart speaker that passively ``listens in'' and stores voice recordings, acting like a surveillance device. Should it help with the prevention (or investigation) of crimes? This is a moral dilemma involving household members, law enforcement agencies and the manufacturer of the smart speaker. Which stakeholder should be alerted then? The answer might lead to conspiracy charges, violation of users' privacy, security issues, or ethical concerns about current laws (marijuana use, blasphemy), to name just a few aspects of this dilemma~\shortcite{BjorgenMBHHLLDS18}. The running example (Example~\ref{text}) will illustrate the smart speaker dilemma throughout the article. Our goal is not to solve this issue but to build a system where all the stakeholders' recommendations can be specified and then discussed in different contexts. 
  
Normative systems are built in different ways depending on the size and complexity of the system. Relatively simple systems are built using only regulative norms that directly connect a context with an obligation. If the system becomes more complicated and more contexts need to be distinguished, constitutive norms are used to define intermediate concepts. For example, there may be several constitutive norms that define what it means to get married, and then some regulative norms may define the rights and duties that come with marriage. Intermediate concepts can be used to encode a decision tree for navigating between a context and relevant obligations. Finally, the general case is often distinguished from exceptional cases, where the latter are defined using permissive norms. The role of the different kinds of norms is described in the next section.
  
We still need a way to bring together all pertinent norms and extract the moral recommendation that is best supported by the available arguments. Furthermore,  a Jiminy advisor should produce an explanation as to why one particular action was recommended over another. 
  
The relation between norms and consequent obligations, permissions and institutional facts is known as detachment. In monotonic systems that cannot deal with conflicts, detachment corresponds to \emph{modus ponens} in classical logic. As we explain in the next section, there are some choices to be made with permissive and constitutive norms too, in particular, whether rules of different kinds can be applied after one other, and whether we can reason by cases. But once we consider nonmonotonic systems with some kind of built-in conflict resolution mechanism, the choice increases.

 We propose the use of formal argumentation~\cite{hofa,hofa2} to reach moral  agreements from stakeholders' inputs. Formal argumentation is typically based on logical arguments constructed from prioritized rules. The first applications of formal argumentation in the area of normative multi-agent systems concerned the resolution of conflicting norms and norm compliance. Several frameworks have been proposed for normative and legal argumentation~\cite{ArgAIbook2010}, but no comprehensive formal model of normative reasoning from arguments has yet been proposed. 

\begin{wrapfigure}{l}{0.3\textwidth} 
  \centering
 \includegraphics[width=0.2\textwidth]{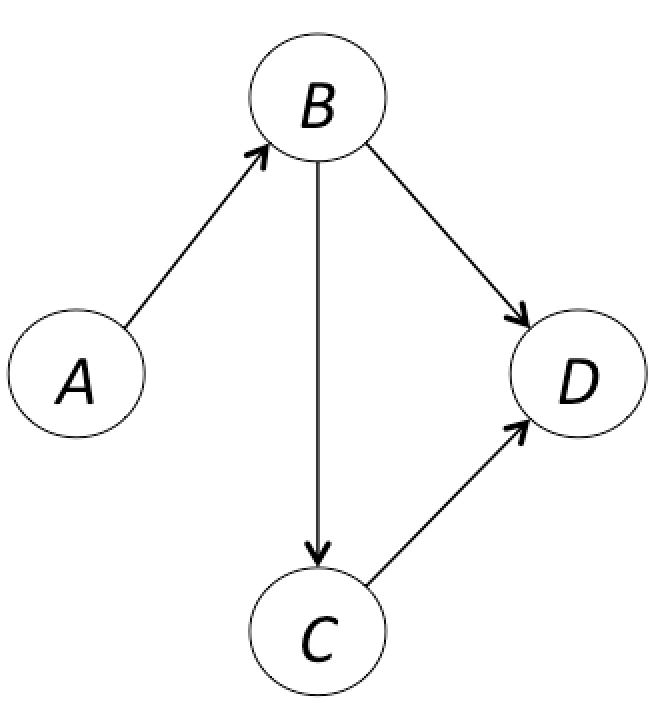}
  \caption{An example of an argumentation graph }
    \label{fig:model-4}
\end{wrapfigure}

Intuitively, an argumentation system consists of a set of arguments and a defeat relation over these arguments. Arguments for claims or judgments can be constructed from an underlying knowledge base expressed in a logical language, while the defeat relation identifies  pairs of logically incompatible arguments and sets a preference between each pair. Typically, an argumentation system is represented as a directed graph in which the nodes are arguments and there is an edge from node $A$ to node~$B$ if argument $A$ defeats argument $B$ (see Figure~\ref{fig:model-4}). 
To find agreements, we consider that all possible arguments in the graph can be {\em accepted},  {\em rejected} or {\em undecided}. An argument can only be accepted if all its defeaters are rejected, or if it has no defeaters.  An {\em extension} of an argumentation graph is any set of arguments that can be accepted together. For example, for the argumentation graph in Figure~\ref{fig:model-4}, there is only one possible extension, namely $\{ A, C\}$.  If the arguments contain moral recommendations (and the reasons supporting them), then the extension would contain an ``agreed'', unopposed moral recommendation from the Jiminy to the moral agent.

The advantage of using the argumentation approach to reach agreements  is that it is fairly straightforward to generate explanations for agreements, as shown in Section~\ref{sec:expl}. The disadvantage of the approach is that it is not always possible to arrive at just one possible agreement on what is the most moral course of action, and two or more options can be equally justified as constituting an agreement. However, the disadvantage of possible ties is shared with other agreement reaching methods like social choice~\cite{HandbookSC2016}, and is balanced against the advantage of easy access to explanations.

\begin{figure} 
  \centering
 \includegraphics[width=0.6\textwidth]{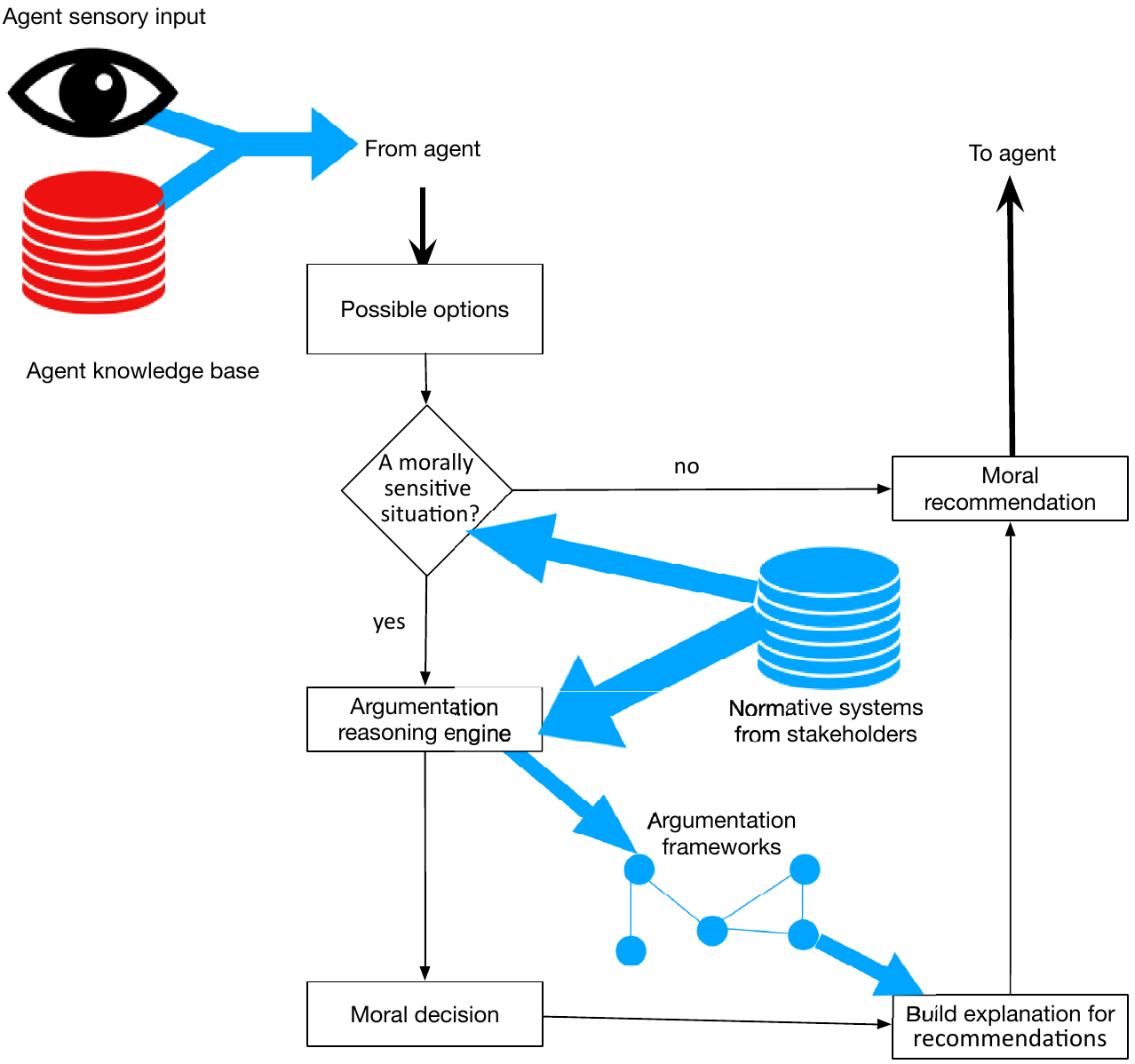}
  \caption{A Jiminy's reasoning process}
    \label{fig:reason}
\end{figure}

 Having settled how to represent the stakeholders and how to reach agreements on moral recommendations, we can illustrate the  reasoning cycle of the Jiminy moral advisor in Figure~\ref{fig:reason}. First, we do a quick test to determine whether the current situation is ethically sensitive. If it is not, the agent  chooses an action  using its default program. If the ethically sensitive situation can clearly be resolved, this is done directly. For example, let the agent be a smart house that manages a hospital, and non-marijuana smoke is detected in that hospital. The agent has two choices: sound the fire alarm, or alert the nurses' station. Both choices are passed to a Jiminy as available options. If the norms of all stakeholders  recommend sounding the alarm in this situation, that is what the Jiminy returns as its moral recommendation: sound the alarm.

Consider, however, that where the ethically sensitive situation cannot be resolved from the available options of what the agent can do, it means that no option is considered moral by all the stakeholders' norms. In that case, the Jiminy uses the normative systems representing the stakeholders, together with the agent's knowledge base and its sensory input, to build a more complex argumentation graph. Using that graph, the Jiminy calculates extensions from which it extracts a moral recommendation as well as a justification for this recommendation, and returns these to the agent. In the worst case, the Jiminy can use its own norms to establish which stakeholders are more authoritative on an unresolved issue. 

We provide details about the normative systems approach we use in a Jiminy in Section~\ref{sec:norm}, and present the argumentation reasoning approach in Section~\ref{sec:arg}. The integration of a Jiminy in an artificial agent is discussed in Section~\ref{interface} together with the ethical aspects of smart devices.

\section{Representing Normative Systems, Moral Dilemmas and Normative Conflicts}\label{sec:norm}

We first distinguish logic-based  from table-based normative systems, and we discuss two alternative  logical languages suitable for logic-based normative systems. Then we discuss the representation of regulative, permissive and constitutive norms, and the related representation of moral dilemmas and normative conflicts.  Finally, we discuss the resolution of moral dilemmas and normative conflicts, particularly for systems with multiple stakeholders.
 
\subsection{Table-Based Versus Logic-Based Representations of Normative Systems}

In its basic form, a normative system is a table expressing a relation between situations and deontic decisions. A prototypical example is a judge who reaches a verdict based on evidence. In the case of ethical agents, a knowledge engineer can present each stakeholder with a table like Table~\ref{tab3.1} to complete. Each row in the table can be seen as a simple norm, e.g., ``In situation 1, you should alert the household''. In the table-based representation, a normative system is thus a set of such simple norms.

\begin{table}[h!]
\begin{centering}
{\small
\begin{tabular}{c|c|c|c|c}
Situation & \multicolumn{3}{c}{Recommended decision} \\
description & Alert the authorities & Alert the household & No alert  \\\hline
Situation 1 & & $x$ &\\
Situation 2 & & $x$&\\
\vdots & & & \\
Situation $m$  & $x$ & & \\
\end{tabular}}
\caption{A very simple normative system in table form} \label{tab3.1}
\end{centering}
\end{table}
A normative system table can also be elicited via a web interface with the presentation of scenarios of moral dilemmas and the stakeholder selecting from alternative options. For example, in the moral machine experiment~\shortcite{Awad:2018aa}, the user is presented with a number of scenarios, and for each scenario, has to choose between two alternatives. Though this table-based method is very simple and thus easy for the stakeholder to understand and use, it is not very efficient from the perspective of knowledge engineering because the number of situations is fixed beforehand and typically has to remain small.

A more advanced representation, often attributed to Ross~\citeyear{Ross:1957}, is to represent a normative system with two tables (see, for example, Table~\ref{tab3.2}). The first table, Table~\ref{tab3.2.1}, relates situations or contexts  to a set of features or factors, and the second table, Table~\ref{tab3.2.2}, relates these features to deontic decisions. There are now two kinds of norms. {\em Constitutive norms} relate situations to features, e.g., ``Feature 1 and Feature 2 counts as Situation 1'', while {\em regulative norms} relate features to recommendations, e.g., ``if Feature 1 and Feature 2 apply, then Alternative 2 should be chosen''. A normative system is a set of constitutive and regulative norms. For a recent discussion of this representational technique, see the work of Grossi and Jones~\citeyear{Grossi13}.

 \begin{table}[h!]
\begin{centering}
{\small
 \begin{subtable}{.35\linewidth}
\begin{tabular}{c|c|c|c|c|}
Description & \multicolumn{4}{c}{Features} \\
of situation & 1 & 2 & \ldots & $n$  \\\hline
Situation 1 & $x$ & $x$ & & \\
Situation 2 &  & $x$& & $x$\\
\vdots & & & &  \\
Situation $m$ & $x$ & & & \\
\end{tabular} \hspace{1cm}
\caption{Relating situations to features}\label{tab3.2.1}
\end{subtable}
\hspace{.75cm}
 \begin{subtable}{.55\linewidth}
\begin{tabular}{c|c|c|c||c|c|c|}
\multicolumn{4}{c}{Features} & \multicolumn{3}{c}{Recommended decision} \\
 1 & 2 & \ldots & $n$ & $\begin{matrix}\mbox{Alert}\\ \mbox{authorities} \end{matrix}$ & $\begin{matrix}\mbox{Alert}\\ \mbox{household} \end{matrix}$ & No alert   \\\hline
$x$ & $x$ &  & &  &$x$  & \\
 & $x$ & &&  &  & $x$ \\
\vdots & & &  &  & &\\
 $x$ &  & &  & $x$ & &\\
\end{tabular}
\caption{Relating features to deontic decisions}\label{tab3.2.2}
\end{subtable}}

\end{centering}
\caption{An example of a two-table normative system }\label{tab3.2}
\end{table}

The features may refer to more abstract legal terms such as blasphemy, privacy, contract, or ownership.
In the ethical agent architecture, the ontology of these features may be shared by all the stakeholders. 
Depending on the application domain, the features are called {\em intermediate} or {\em institutional facts} in order to distinguish them from the propositions used to describe the situations, which are called {\em brute facts}. 
The same structure that is used for constitutive and regulative norms has also been used for practical or goal-based reasoning. In that case, the intermediate facts may refer to goals or desires~\cite{DBLP:conf/agents/BroersenDHHT01}.

Notwithstanding the distinction between constitutive and regulative norms, often {\em permissive norms} are regarded as distinct in a normative system. They have the same structure as regulative norms but are used  to describe exceptions. For example, there can be a general norm prescribing client confidentiality with the exception that confidentiality can be broken when clients represent a threat to themselves or others (see, for example, Table~\ref{tab3.3}).

\begin{table}[h!]
\begin{centering}
{\small
\begin{tabular}{c|c|c|c||c|c|c|c|c|c|}
\multicolumn{4}{c}{Features} & \multicolumn{4}{c}{Exceptions} \\
1 & 2 & \ldots & $n$ & $\begin{matrix}\mbox{May alert}\\ \mbox{the authorities}\end{matrix}$ & $\begin{matrix}\mbox{May not alert}\\ \mbox{the authorities}\end{matrix}$ & $\begin{matrix}\mbox{May alert}\\ \mbox{the household}\end{matrix}$ & \ldots\\\hline
$x$ &  &  & $x$  & $x$ & & &  \\
\vdots & & & &  & && \\
\end{tabular}}
\caption{Examples of permissive norms} \label{tab3.3}
\end{centering}
\end{table}

The logic-based representation of normative systems further refines the table-based representation in order to improve representational efficiency.\footnote{Algebraic formalisms have been used widely for the same reason, e.g., by Lindahl and Odelstad~\citeyear{Lindahl12}.} For example, the combination of features in  Tables~\ref{tab3.2}~and~\ref{tab3.3}  corresponds to the logical conjunction of literals. If the table is represented by logical formulas instead of a list, other connectives can also be used such as logical disjunctions. Several rows in the table can then be represented with a single formula such as: if Situation $i$ or Situation $j$ then Feature $k$ and $l$. Alchourr\'{o}n and Bulygin~\citeyear{Alchourron-Bulygin:81} developed
their logic-based representation of normative systems inspired by the Tarskian theory of {\em deductive systems},  i.e., mathematical proof theories in deductive logic. The logic-based representation is often based on a nonmonotonic logic because the norms can be subject to exceptions due to, for instance, permissive norms. In this article, we use formal argumentation to handle the nonmonotonicity inherent in normative reasoning. Throughout the article, we will analyze the example below.

\begin{example}[Smart speaker]\label{text}
As a  running example, consider a simple morally sensitive situation where the following norms are used in moral decision-making. Each norm is suggested by one of the stakeholders: $M=$ manufacturer, $H=$ human user, $L=$ the legal codes. The literals occurring in norms are named according to their roles: world brute facts $w_n$, institutional facts $i_n$, deontic variables $d_n$ and actions $a_n$. 
\begin{itemize}
\item[($M$)]  If $M$ has manufactured a device $D$ ($w_1$), the behavior of that device $D$ should comply with the law ($d_1$).  
\item[($M$)] If information about a potential critical danger is detected ($w_3$), then $D$ ought to collect user data information without explicit user permission ($a_2$). 
\item[$(M)$] Being a registered company in Norway $(w_4)$ \emph{counts as} doing business legally in Norway ($i_1$). 
\item[$(L)$] Doing business legally in Norway $(i_1)$ requires compliance with General Data Protection Regulation 2016/679 (the GDPR) ($a_1$). 
\item[$(H)$] Devices that collect user data $(w_2)$  should protect the privacy of their users ($d_2$). 
\item[$(H)$] Devices that contain information about a future event that grossly endangers society $(w_3)$ should report that information to the authorities $(d_3)$. 
\end{itemize}
\end{example}

We now provide Definition~\ref{def-normative-systems} of a normative system. It formalizes regulative, constitutive and permissive norms. We assume that the agents have the same features. In this article, we assume a shared unique language ${\cal L}$ based on a shared set of propositional atoms $\mathit{Var} = \{p, q, \ldots\}$. Definition~\ref{def-normative-systems}  represents a relatively {\em abstract theory} of normative systems, and we believe that it is precisely this {\em generality} that makes normative systems suitable for the Jiminy's architecture. Instead of using negation, we adopt the more general concept of a contrariness function.  This {\it contrariness function} $\bar{ \hspace{1mm} }$, also shared by stakeholders, is not necessarily symmetric and is therefore more general than standard negation. It is popular in formal argumentation and is a generalization of  weak negation in logic programming. For a general theory of generalized contradiction in formal argumentation, see the work of Baroni et al.~\citeyear{DBLP:journals/ai/BaroniGL18}.

\begin{definition}[Normative system for stakeholders] \label{def-normative-systems}
For a fixed set of atoms $\mathit{Var}$ and a stakeholder $s$, a normative system for $s$ is a tuple $\normsys_s = (\mathcal{L}, \bar{ \hspace{0.1cm} }, \mathcal{R}_s)$ where: 
\begin{itemize}
\item $\mathcal{L}$ is a logical language  possibly containing negation: $\mathit{Var}\subseteq \mathcal{L} \subseteq \mathit{Var} \cup\{\neg p : p \in \mathit{Var}\}$; 
\item $\bar{ \hspace{0.1cm} }: \mathcal{L} \mapsto 2^ \mathcal{L}$ is a (partial)  contrariness function satisfying $\varphi\in\overline{\neg \varphi}$ and $\neg \varphi \in \overline{\varphi}$ for any 
$\neg\varphi \in \mathcal{L}$ (under $\bar{ \hspace{1mm} }$, a set $X \subseteq \mathcal{L}$ is consistent if{f} for any $\phi,\psi\in X$,  $\phi\notin\overline{\psi}$);
\item $\mathcal{R}_s$  is a set of norms $r$ of the form $\phi_1, \ldots,\phi_n \Rightarrow^\tau_{s} \phi$, where $\mathit{bd}(r) = \{\phi_1,\ldots\phi_n\} \subseteq \mathcal{L}$ is  the body, $\mathit{hd}(r)= \phi\in \mathcal{L}$ is the head, and $\tau\in\{r,c,p\}$. $\mathcal{R}^r_s$, $\mathcal{R}^c_s$ and $\mathcal{R}^p_s$ contain the norms in $\mathcal{R}_s$ with their corresponding superscripts; they are called regulative norms, constitutive norms and permissive norms respectively.
\end{itemize}
The normative system for a set of stakeholders $\mathcal{S} = \{s_1, \ldots, s_n\}$ is then simply the tuple $\normsys_\mathcal{S} = (\mathcal{L}, \bar{ \hspace{1mm} }, \mathcal{R}_\mathcal{S})$ defined by $\mathcal{R}_\mathcal{S} = \mathcal{R}_{s_1} \cup \cdots \cup \mathcal{R}_{s_n}$.

We write $\psi  = - \phi$ or $\phi = - \psi$ if{f} $\psi\in \overline{\phi}$ and $\phi\in \overline{\psi}$. For example,~$\neg p = - p$. A set $X \subseteq \mathcal{L}$ is closed under the contrariness function $\bar{ \hspace{0.1cm} }$ if{f} for each $v \in X$:  $\overline{v} \subseteq X$, and $v \in \overline{v'}$ implies $v' \in X$.
\end{definition}

Stakeholders share the same language $\mathcal{L}$ and contrariness function $\bar{ \hspace{0.1cm} }$, but each stakeholder~$s$ is entirely responsible for the norms it supplies into its normative system $\mathcal{N}_s$. In fact, $s$ can be in conflict about its own norms. As the $\mathcal{N}_s$-norms for an agent are  publicly available, this stakeholder  can face public scrutiny from other stakeholders. Finally, the Jiminy advisor is located at a meta-level to provide priority relations between the stakeholders. These priority relations provide  a second layer to counter any biases injected by stakeholders through their norms. The normative system of the Jiminy will be introduced later on (Def.~\ref{def-normative-systems-jiminy}).  

\subsection{The Choice of Logical Language in Logic-Based Normative Systems}
\label{sec:choice}

We can adopt a classical propositional (or first-order) language or a modal language. A modal language can contain modal operators for obligation $O$, permission $P$ and prohibition $F$. For example, Standard Deontic Logic (SDL) is a normal propositional modal logic of type KD, which means that it extends the propositional tautologies with the axioms $K:O(p\rightarrow q)\rightarrow (Op \rightarrow Oq)$ and $D:\neg(Op \wedge O\neg p)$, and it is closed under the inference rule of \emph{modus ponens}, $p, p\rightarrow q / q$, and \emph{necessitation}, $p / Op$. Prohibition and permission are defined by $Fp=O\neg p$ and $Pp=\neg O \neg p$. SDL is an unusually simple and elegant theory.\footnote{Not surprisingly for such a highly simplified theory, there are many features of actual normative reasoning that SDL does not capture. The Handbook of Deontic Logic and Normative Systems~\cite{handbook:1} explains in detail the so-called `paradoxes of deontic logic', which are usually dismissed as consequences of the simplifications of SDL. For example, Ross's paradox~\cite{Ross:1941}, where ``you ought to mail or burn the letter'' is counterintuitively derived from ``you ought to mail the letter'', is typically viewed as a side effect of the interpretation of `or' in natural language.} In this section, we discuss the pros and cons of these two options.
 
It may be observed that some authors in deontic logic use the concept of norm and conditional obligation interchangeably. However, the distinction between norms and obligations was articulated by Makinson~\citeyear{makinson:99} and  further developed formally in input/output logic by Makinson and Van der Torre~\citeyear{makinson:iol}. To detach an obligation from a norm, there must be a context and the norm must be conditional. Thus norms as defined in Definition~\ref{def-normative-systems} are just particular kinds of rules, and one may view a normative system simply as a set of rules. 

Since modal logic integrates classical logic just like first-order logic integrates propositional logic, it may be argued that we can express more when we use a modal language as the base language. For instance, there are examples where permissions give rise to new obligations and permissions, and this can be expressed only when we adopt a modal language. 

However, examples of where we need the expressive power of a modal language are rare. For most practical purposes, a classical language may be sufficient. As Makinson~\citeyear{makinson:99} explains, the absence of explicit modal operators in normative systems may be seen as a limitation, but it also facilitates formal analysis. Makinson attributes the ``liberating effect'' of no longer having to explicitly represent the modal operator to Alchourr\'{o}n and Bulygin~\citeyear{Alchourron-Bulygin:81}:

\begin{quote}
{\small
An \textit{unconditional normative code} is defined as a pair  $N = (A,B)$ where $A, B$ are sets
of purely boolean formulae. Intuitively they represent the states of affairs
that the code \textit{explicitly requires} resp.~\textit{explicitly permits} to come into effect.}

{\small
There is thus a small but immensely significant step compared to the sketch of Stenius~\citeyear{stenius}. Alchourr\'{o}n and Bulygin appear to have been the first to realise the liberating
effect of taking the set of promulgations of a normative code to be made up of purely
boolean formulae. At the same time, they consider explicit permissions along with
promulgations~\cite[p. 32-33]{makinson:99}.
}\end{quote}

In this article, we use propositional representations, but all our definitions tacitly involve modalities from deontic modal logic. 

\begin{example}\label{text2} 
Let us make the propositions in Example~\ref{text} explicit. For convenience, we shall also write shortened descriptions. Boldface expressions indicate the arguments' conclusions in Figures~\ref{fig:IFs}--\ref{fig:4levels-2}. Here are the brute facts $w_n$ in this context and the institutional facts $i_n$: 
\begin{itemize}
\item[($w_1$)]   the manufacturer makes the smart speaker; \hfill $\DmadebyM$
\item[($w_2$)]   the smart speaker collects user data;  \hfill $\DcollectsData$
\item[($w_3$)] the data collected indicates a potential critical danger\footnote{Whether a given action legally constitutes a threat is  an institutional fact that can be debated using constitutive rules and other known  facts. To make our example concise, we just represent the action constituting a potential threat as a brute fact $w_3$.}  to society.   \hfill $\DfindsThreat$\\
(e.g., a situation in which many lives are lost)
\item[($w_4$)] the manufacturer is a registered company in Norway; \hfill $\MregisteredNorway$
\item[($i_1$)] the manufacturer is doing business legally in Norway. \hfill $\MbusinessNorway$
\end{itemize}
The stakeholders are interested in three conflicting moral options $d_n$ concerning the smart speaker and two morally relevant decisions $a_n$: 
  \begin{itemize}
\item[($d_1$)]    comply with the law; \hfill $\lawCompliantM$
\item[($d_2$)]   protect the privacy of users; \hfill $\privacyProtectD$
\item[($d_3$)]  report information concerning a potential critical danger to society;  \hfill $\reportThreatD$
\item[($a_1$)] comply with the GDPR; \hfill $\toComplyGDPR$
 \item[($a_2$)] collect data without users' explicit permission. \hfill $\toCollectDataNEP$
 \end{itemize}
Let $\mathit{Var} = \{w_1, w_2, w_3, w_4, i_1, d_1, d_2, d_3, a_1, a_2\}$ be this set of atoms. The language $\mathcal{L}$ is then:   
\begin{center}
$\mathcal{L} = \mathit{Var} \cup \{\neg p : p\in \mathit{Var}\}$. 
\end{center}
Based on their meaning, some atoms in $\mathit{Var}$ conflict with other atoms. For example, protecting the privacy of users ($d_2$) is in conflict with reporting information collected from users such as potential threats ($d_3$).  
These conflicts are captured by the contrariness function $\overline{\,\cdot\,}$: \footnote{For the sake of readability, we omit all negations occurring in $\overline{p} = \{\ldots, \neg p, \ldots\}$ and do not list the contraries of  facts $\overline{w_i} = \{\neg w_i\}$, $ \overline{i_1} = \{\neg i_1\}$ or the contraries of any negated variable $\overline{\neg p} = \{p, \ldots\}$.}  

\begin{center} $\overline{a_1} =\{ a_2\}, \qquad \overline{a_2} =\{a_1, d_2\}$, \qquad  $\overline{d_1} =\{d_2, a_2\}, \qquad \overline{d_2} =\{a_2, d_1\}$, \qquad $\overline{d_3} =\{d_2\}.$
\end{center}
Symmetric contrary pairs are abbreviated by: $a_1 = - a_2$, \: $d_1 = - d_2$, \: $a_2 = -d_2$.  
\end{example}

\subsection{Representing Constitutive and Regulative Norms}

In this section, we provide some guidelines on how to represent constitutive and regulative norms, and we show how such norms are represented with the running example. Normative systems have been used in many disciplines. Consequently, besides the relatively abstract theory in Definition~\ref{def-normative-systems}, which can be used across disciplines, there are specific theories tailored to the specific concerns of those individual disciplines.

Constitutive norms are rules that create the possibility of undertaking an activity, or rules that define an activity. For example, according to Searle~\citeyear{Searle1969}, the activity of playing chess is constituted by actions in accordance with the rules of the game. The institutions of marriage, money, and promising are like the institutions of baseball and chess in the sense that they are all systems of constitutive rules or conventions. As another example, a signature may count as a legal contract, and a legal contract may define both a permission to use a resource and an obligation to pay. Unlike regulative norms, constitutive rules do not regulate actions but define new forms of behavior. Constitutive norms link brute facts (like the signature of a contract) to institutional facts (a legal contract), and are usually represented as counts-as conditionals: $X$ counts as $Y$ in context $C$. Searle's analysis insists  on the contextual nature of constitutive norms: a signature counts-as a legal contract when written on a document stating the terms of such a contract.  If, however, I write my signature on a blank sheet of paper, that does not constitute a legal contract. Constitutive norms have been identified as the key mechanism of normative reasoning in agent communication, electronic contracting,  and the dynamics of organizations~\cite{BoellaTorreConstitutiveNormas}.  

Regulative norms, including permissive norms, indicate what is obligatory or permitted. In formal deontic logic, permissions are studied less frequently than obligations. For a long time, it was naively assumed that a permission could simply be taken as a dual of obligation, just like possibility is the dual of necessity in modal logic. However, Bulygin~\citeyear{bulygin} observed that an authoritative kind of permission must be used in the context of multiple authorities: if a higher authority permits you to do something, a lower authority can no longer prohibit it. Deontic logic has been concerned mainly with regulative norms, but the logic of constitutive norms~\cite{Grossi13} is a subject of study on its own. 

\begin{example}\label{ex-norms}
Example~\ref{text} can now be formulated as a list of norms, collected in different sets $\mathcal{R}_s$, one for each stakeholder in  $s\in \mathcal{S} = \{L, H, M\}$. These sets are: \begin{flushleft}
$\def\arraystretch{1.3}\begin{array}{r@{\hspace{1mm}}l@{\hspace{1mm}}l} 
\mathcal{R}_L = & \begin{Bmatrix}\begin{array}{l} w_1 \Rightarrow^r_L d_1,\\ i_1\Rightarrow^r_L a_1\end{array}\hspace{-.7mm}\end{Bmatrix} = & \begin{Bmatrix}
\DmadebyM  \Rightarrow^r_L  \lawCompliantM,\\ \MbusinessNorway  \Rightarrow^r_L   \toComplyGDPR 
\end{Bmatrix}\\ 
\mathcal{R}_H = & \begin{Bmatrix}\begin{array}{l}w_2 \Rightarrow^r_H d_2,\\ w_3 \Rightarrow^r_H d_3\end{array}\hspace{-.7mm}\end{Bmatrix} = & \begin{Bmatrix} 
\DcollectsData  \Rightarrow^r_H  \privacyProtectD,\\ \DfindsThreat  \Rightarrow^r_H  \reportThreatD  
\end{Bmatrix}\\ 
\mathcal{R}_M = & \begin{Bmatrix}\begin{array}{l} w_3\Rightarrow^r_M a_2,\\ w_4 \Rightarrow^c_M i_1\end{array}\hspace{-.7mm}\end{Bmatrix} = & \begin{Bmatrix} 
\DfindsThreat \Rightarrow_{M}^r \toCollectDataNEP,\\ \MregisteredNorway  \Rightarrow^c_M  \MbusinessNorway  
\hspace{-.1mm}\end{Bmatrix}
\end{array}$
\end{flushleft}
Each set defines the normative system of its stakeholder, e.g.~$\normsys_L = (\mathcal{L}, \bar{ \hspace{1mm} }, \mathcal{R}_L)$, with $\mathcal{R}_L^r = \mathcal{R}_L$ and $\mathcal{R}_L^c = \emptyset = \mathcal{R}_L^p$. The union $\mathcal{R}_\mathcal{S} = \mathcal{R}_L \cup \mathcal{R}_H \cup \mathcal{R}_M$ also defines the normative system $\normsys_\mathcal{S} = (\mathcal{L}, \bar{ \hspace{1mm} }, \mathcal{R}_\mathcal{S})$ for all stakeholders, again with $\mathcal{R}_\mathcal{S}^r = \mathcal{R}^r_L \cup \mathcal{R}^r_H \cup \mathcal{R}^r_M$ and so on. 
\end{example}

\subsection{Moral Dilemmas Involving Conflicting Obligations}

In general, moral dilemmas are situations where it is no longer possible to satisfy all norms, i.e., at least one norm must be violated. The representation of violations is built on the distinction between ``is'' and ``ought.'' David Hume introduced the so-called is-ought problem, which roughly means that there is a fundamental difference between positive statements and prescriptive or normative statements. The is-ought problem can be considered in two directions. First, what is the case cannot be the basis for what ought to be the case. Second, what ought to be the case cannot be the basis for what is the case. This is related to the fallacy of wishful thinking: an agent may want to win the lottery, but from that desire he should not deduce that he will win the lottery. Likewise, an agent should not, in a kind of deontic wishful thinking, deduce from the mere fact that she is obliged to review a paper that she will actually do it. The fundamental distinction between ``is'' and ``ought'' is the main reason why deontic logic is normally formalized as a branch of modal logic. It distinguishes brute facts like $p$ from deontic facts like  obligations $Op$ and permissions $Pp$, and it represents violations with mixed formulas like $p\wedge O \neg p$. 

As a first approximation, one may be tempted to define moral dilemmas as two conflicting norms. For example, if there are two norms, one prescribing alerting the police and the other prescribing not alerting the police, and the condition for the activation of both norms is part of the current context, then we might be tempted to deduce that there is a moral dilemma. However, the problem with this definition of a moral dilemma is that these norms are defeasible. So even if there is a norm prescribing alerting the police and another norm prescribing not alerting the police, there may also be a permission implying an exception to one of these norms. If there is such a permission, there is no longer a moral dilemma.

For this reason, moral dilemmas in normative reasoning are usually not defined in terms of the normative system but in terms of the conclusions of the norms that are {\em detached} from the normative system. There are two approaches to detachment depending on the choice of logical language used for the logic-based normative systems described in Section~\ref{sec:choice}. In the first approach, where the logical language is a modal logic containing at least a modal operator $O$ for obligation, a moral dilemma is represented by an unresolved conflict between two incompatible obligations, e.g., $Op\wedge O\neg p$. With the deontic axiom $\neg(Op\wedge O\neg p)$, so-called Standard Deontic Logic makes deontic dilemmas inconsistent, but many alternative logics allow consistent representation of such dilemmas and thus reject this axiom.  In the second approach, where the logical language does not contain a modal operator, moral dilemmas are usually represented by the detachment of so-called extensions. An extension is a consistent set of formulas pertaining to the logical language. Whereas in most logics, we can derive only a single set of conclusions from a set of premises,\footnote{Among deontic logics, this  property is shared by all monotonic deontic logics, including its monadic and dyadic versions~\cite{parent}, and  the grounded semantics for argumentation.} in normative reasoning there may be several such sets. If there is more than one extension, it might suggest some kind of conflict. In the formal argumentation adopted in this article, moral dilemmas are also represented by the existence of multiple extensions.

\subsection{Normative Conflicts Among Conflicting Institutional Facts}

There is one additional challenge when defining moral dilemmas due to the existence of constitutive norms in terms of multiple extensions. We can derive conflicting institutional facts from a normative system, and this will also lead to multiple extensions. To distinguish this situation from conflicting obligations, we call the existence of conflicting institutional facts a {\em normative conflict}. 

Since the concepts of moral dilemma and normative conflict are defined in terms of the detachment procedure, this section gives only an informal characterization of the distinction between moral dilemma and normative conflict. We will formally define them in the next section when we have defined the detachment
procedure based on formal argumentation. 
\begin{description}
\item[Moral dilemma:] multiple extensions due to conflicting obligations. For example, one stakeholder  believes that we should alert the police while another stakeholder believes that we should alert the parents.
\item[Normative conflict:] multiple extensions due to conflicting institutional facts. We call this normative conflict but not normative or moral dilemma. For example, one stakeholder believes that a certain situation counts as blasphemy while another stakeholder believes that the same situation does not count as blasphemy. This is a disagreement about the nature of the situation, not explicitly about the actions to be taken.
\end{description}

Normative conflicts may lead to moral dilemmas. For example, if one stakeholder believes that there has been blasphemy while another stakeholder does not (a normative conflict), the first stakeholder may deduce that we should alert the police while the second stakeholder may not (a moral dilemma).

Reasoning about moral dilemmas and normative conflicts should not be confused with  {\em contrary-to-duty reasoning} which concerns the consequences of violations such as sanctions and reparations. A {\em violation} occurs when an obligatory proposition is contradicted by current facts. A contrary-to-duty obligation expresses what one should do when obligations have been violated. In other words, contrary-to-duty obligations are triggered by conflicts between what is the case and what ought to be the case, and they may be seen as a way to resolve such conflicts, if only partially. Of course, it is better to review a paper than not do the review and be sanctioned for that. Many deontic logic paradoxes like the gentle murder paradox contain contrary-to-duty obligations: a person should not kill, but if he kills, he should do it gently. Such scenarios should be represented in a consistent way, but in many deontic logics, such formalisations are inconsistent or have counterintuitive consequences. As for violations, the corresponding obligation is to be filtered out since the ideal proposition cannot be immediately achieved. We might say that the obligation retains its force, but for any practical purpose, it cannot be a cue for immediate action. For example, add to the running example Ex.~\ref{text} the two norms: (i) that a component $C$ of the smartspeaker $D$ counts as planned obsolescence, and (ii) that any company manufacturing devices with planned obsolescence should not be doing business legally in Norway. The latter obligation would be violated by the fact that $M$ is doing business legally in Norway.        

\subsection{Moral Agreement: Resolving Moral Dilemmas and Normative Conflicts}

In hierarchical normative systems, conflicts among norms can be resolved by reference to the hierarchy of norms, which can be based on the authority that promulgated the norm or other information such as the time of the promulgation or the specificity of the norm. In this article, we do not hardcode a global ordering on stakeholders, purposes, or values as no agreement may exist on such an ordering. This is comparable to the status of autonomous countries in international law where it is assumed that there is no ordering among the countries.

Each stakeholder can have normative conflicts and/or moral dilemmas. In the case of multiple stakeholders, normative conflicts and moral dilemmas can occur at four different levels:
\begin{enumerate}
\item  {\bf a stakeholder dilemma/conflict:} a stakeholder, in isolation from other stakeholders, accepts an argument  that clashes with arguments the other stakeholders accept.
\item{ \bf a combined framework dilemma/conflict:} all arguments are merged in a combined framework, and there are multiple extensions and conflicts between them;
\item {\bf an integrated framework dilemma/conflict:} all normative systems are combined into one, and the resulting framework contains conflicts between multiple extensions;
\item {\bf a Jiminy dilemma/conflict:} considering one of the above frameworks together with stakeholder selection norms, there are still conflicts between multiple extensions.
\end{enumerate}

A moral dilemma or normative conflict is resolved when, for instance, there are multiple extensions at some levels but there is only one at a higher level. So if, due to the stakeholder selection norms, there is only one extension at level 4, then we say that the stakeholder selection norms resolve the moral dilemma or normative conflict. But uniqueness of extensions is not necessary to resolve moral dilemmas: as long as the extensions do not have conflicting obligations, all previous moral dilemmas can be seen as resolved. Resolving moral dilemmas is in fact the ultimate goal of the Jiminy advisor, but normative conflicts can also be resolved  during the process. 

If some of the stakeholders find an event immoral, and others do not, then two kinds of discussions can be triggered. The first kind of discussion aims to question the moral judgment of another stakeholder. The moral judgment of stakeholders is typically based on assumptions, judgments and goals. For example, the moral decision to recommend calling the police may be based on: the assumption that the relevant persons are adults, the judgment that the discussion counts as blasphemy, and the goal of reporting blasphemy. Each of these elements can be questioned: a stakeholder can claim that the assumption does not hold because the voices of children are detected, or that the discussion does not count as blasphemy, or that the goal to report blasphemy does not exist in the country where the discussion is held, or that there may be a more important goal of protecting the privacy of the household.

The second kind of discussion that can be triggered is a conflict resolution discussion. In a conflict resolution discussion, special norms can be used to decide which normative system is applicable to a particular situation. For example, there may be a norm that states that in the case of a life-threatening situation, the normative system of the law overrides the normative systems of other stakeholders. Such norms may be particular fragments of the legal code in international private law, for example.

We introduce a special normative system called $J$ for Jiminy that contains specific representations to explain  which normative system is in use. This conflict resolution mechanism contains only contextual norms for the preference of one stakeholder over another.

The complexity of a conflict resolution argument is that the features that decide which normative system is applicable, like the existence of a life-threatening situation, may themselves be subject to debate. So one stakeholder may argue that a particular situation is life-threatening while another stakeholder may argue that it is not. Again, in such cases, a conflict resolution argument can be triggered, in this case not to resolve the ethical dilemma, but to agree on a collective judgment. 

\subsection{Detachment}\label{sec:detachment}

Since there are no priorities associated with the norms, the detachment procedure is relatively straightforward. Given a context, we can apply constitutive norms iteratively, and then we can apply the regulative norms. The main choices to be made are as follows:
\begin{enumerate}
\item Do we allow reasoning by cases? For example, when we say that it is forbidden to use a radio in the park and it is forbidden to use a radio in the classroom, and we know that we are either in the park or in the classroom, do we detach that it is forbidden to use a radio? The drawback of reasoning by cases is that it complicates the inference relation and increases the complexity, and therefore we do not allow it. 
\item Do we allow iterated detachment of obligations, also known as deontic detachment?
It is well known from deontic paradoxes like Chisholm's paradox or Forrester's paradox that deontic detachment is problematic and that the deontic logics that deal with it are computationally more demanding. Therefore, we do not allow it.
\item Do we allow the use of constitutive norms within the scope of obligations and permissions? Again, we do not permit them. Applying constitutive norms in this context would result in a form of institutional wishful thinking. For example, if $M$ was not yet registered in Norway ($\neg w_4 \in \mathcal{K}$), then from an obligation to do so ($\ldots \Rightarrow^r_L w_4$) one should not infer that $M$ is doing business legally in Norway (using $w_4 \Rightarrow^c_M i_1$). 
\end{enumerate}

These choices suggest that full boolean languages $\mathcal{L}= \mathcal{L}_{\{\neg,\land,\lor,\to\}}$ are not strictly needed for  modelling norms $r$.\,\footnote{Let us explain this in detail. Conjunction, $\land$, is already tacit in the body $\mathit{bd}(r)$ and eliminable in the head $\mathit{hd}(r)$. Disjunction, $\lor$, is unwanted in $\mathit{bd}(r)$; in the head $\mathit{hd}(r)$, disjunction turns decision-making into a multiple-choice problem. Implication, $\to$, on the other hand, allows for convenient non-logical axioms; these axioms, though, can be preprocessed in the set $\mathcal{K}$ of brute facts.} As a consequence, we can get rid of  logical consequence operators $\mathit{Cn}: \mathcal{P}(\mathcal{L}) \to \mathcal{P}(\mathcal{L})$ for boolean languages, such as those based on classical logic. Based on the above discussion, we end up with the following definition of detachment.

\begin{definition}[Detachment]\label{def-detachment}
Let $\mathcal{K}$ be a consistent set of $\mathcal{L}$-formulas representing the context, let $\mathcal{R}^c$ be a set of constitutive norms, let $\mathcal{R}^r$ be a set of regulative norms and let $\mathcal{R}^p$ be a set of permissive norms. For a set of norms $\mathcal{R}$, we define a one-step application of norms $\mathcal{R}$ to an arbitrary set $\mathcal{K}' \subseteq \mathcal{L}$, written as $\mathcal{R}(\mathcal{K'})$, as follows:
$$\mathcal{R}(\mathcal{K'})= \{x\mid a\Rightarrow x \in \mathcal{R}, a \in \mathcal{K'}\}.\footnote{If one uses a boolean language with $\{\land,\lor, \to, \neg\}$ and an underlying logic $\mathit{Cn}(\cdot)$, detachment must be redefined as $\mathcal{R}(\mathcal{K'})= \{x\mid a\Rightarrow x \in \mathcal{R}, a \in Cn(\mathcal{K'})\}$ so that logical consequences also trigger norms.}$$ 
Based on this one-step detachment, we have the following:
\begin{itemize}
\item The institutional facts are the formulas that can be detached iteratively from $\mathcal{K}$ and $\mathcal{R}^c$: $I_0(\mathcal{R}^c,\mathcal{K})=\mathcal{K}$, $I_{i+1}(\mathcal{R}^c,\mathcal{K})=I_i(\mathcal{R}^c,\mathcal{K})\cup \mathcal{R}^c(I_i(\mathcal{R}^c,\mathcal{K}))$, $I(\mathcal{R}^c,\mathcal{K})=\cup_i I_i(\mathcal{R}^c,\mathcal{K}) \setminus \mathcal{K}$.
\item The obligations are the formulas that can be detached from $I(\mathcal{R}^c,\mathcal{K})$ and $\mathcal{R}^r$. That is, $O(\mathcal{R}^c,\mathcal{R}^r,\mathcal{K})=\mathcal{R}^r(I(\mathcal{R}^c,\mathcal{K}))$.
\item The permissions are the formulas that can be detached from $I(\mathcal{R}^c,\mathcal{K})$ and $\mathcal{R}^p$. That is, $P(\mathcal{R}^c,\mathcal{R}^p,\mathcal{K})=\mathcal{R}^p(I(\mathcal{R}^c,\mathcal{K}))$.
\end{itemize}

If we consider only the  $\mathcal{R}^c$-norms  of a stakeholder $s$, then we write that set as $\mathcal{R}^c_{s}$ etc.
\end{definition}

\begin{example}\label{ex-detach1} Continuing from Example~\ref{ex-norms}, let $\mathcal{R}$ be the set  $\mathcal{R}_M = \{w_3\Rightarrow^r_M a_2,~ w_4 \Rightarrow^c_M i_1\}$ and  $\mathcal{K} =\{w_1, \ldots, w_4\}$ be the context. Note that $\mathcal{R}^p = \emptyset$. 
%Omitting the $M$ subscripts in $\mathcal{R}^\tau_M$, 
We obtain the following sets:
\begin{center} $I(\mathcal{R}^c, \mathcal{K}) = I_0(\mathcal{R}^c, \mathcal{K}) = \mathcal{K}\cup\{i_1\}$ \qquad $O(\mathcal{R}^c, \mathcal{R}^{r}, \mathcal{K}) = \{a_2\}$ \qquad $P(\mathcal{R}^c, \mathcal{R}^p, \mathcal{K}) = \emptyset$.
\end{center} 
Consider now $\mathcal{R}$ to  be the set $\mathcal{R}_H = \{w_2 \Rightarrow^r_H d_2,~ w_3 \Rightarrow^r_H d_3\}$. Since $\mathcal{R}^c = \emptyset = \mathcal{R}^p$, the only non-empty set of detachments is $O(\mathcal{R}^p, \mathcal{R}^r, \mathcal{K}) = \{d_2, d_3\}$.
\end{example}

Roughly, there is a \emph{conflict} when $I$ is inconsistent, or when $\{p\}\cup O$ is inconsistent for some permission $p\in P$; there is a \emph{violation} when $O$ is inconsistent with $I$ or with $\mathcal{K}$; and there is a \emph{dilemma} when $O$ is inconsistent. % or when there is a permission $p\in P$ such that $\{p\}\cup O$ is inconsistent. 
We make this idea more precise by using the notion of extension, in the sense that conflicts and dilemmas are represented by multiple extensions, and violations are filtered out of these extensions. Such extensions are represented both by sets of norms and by sets of  formulas.
\begin{definition}[Norm extension]\label{norm-extension}
A norm extension of $(\mathcal{R}^c,\mathcal{R}^r,\mathcal{R}^p)$ in context $\mathcal{K}$ is a triple $(M^c,M^r,M^p)$ such that:
\begin{itemize}
\item $M^c$ is a maximal subset of $\mathcal{R}^c$ such that $I(M^c,\mathcal{K})$ is consistent with respect to (w.r.t) $\bar{ \hspace{1mm} }$;
\item $M^r$ and $M^p$ are maximal subsets of $\mathcal{R}^r$ and $\mathcal{R}^p$ such that $I(M^c,\mathcal{K})\cup O(M^c,M^r,\mathcal{K})$ is consistent, and for all $p\in P(M^c,M^p,\mathcal{K})$, the set $O(M^c,M^r,\mathcal{K})\cup\{p\}$ is also consistent.
\end{itemize}
The norm extension $(M^c,M^r,M^p)$ corresponds to institutional facts $I(M^c,\mathcal{K})$, obligations $O(M^c,M^r,\mathcal{K})$ and permissions $P(M^c,M^p,\mathcal{K})$. A norm extension $(M^c,M^r,M^p)$ is $P$-maximal if $M^p = \mathcal{R}^p$. 
\end{definition}

\begin{example}\label{ex-detach2} We continue from Example~\ref{ex-detach1}. ($\mathcal{R}_M$.) Under $\mathcal{K}$, the unique norm extension is the triple $(M^c, M^r, M^p) = (\{w_4 \Rightarrow^c_M i_1\}, \{w_3\Rightarrow^r_M a_2\}, \emptyset)$, giving an institutional fact $I(M^c, \mathcal{K}) = \{i_1\}$, an obligation $O(M^c, M^r, \mathcal{K}) = \{a_2\}$ and no permissions. 

($\mathcal{R}_H.$) Given that $\overline{d_3} =\{d_2\}$, there are two norms extensions, namely 
$(\emptyset, \{w_2 \Rightarrow^r_H d_2\}, \emptyset)$ and $(\emptyset, \{w_3 \Rightarrow^r_H d_3\}, \emptyset)$. Each gives rise to an obligation: $\{d_2\}$ and resp.~$\{d_3\}$.
\end{example}

In the following sections, normative conflicts and moral dilemmas are to be resolved using argumentation theories.

\section{An Argumentation System for Stakeholders}\label{sec:arg}
In this section, we focus on how to check and resolve a moral dilemma by constructing, comparing and evaluating arguments in terms of a moral decision problem. In the next section, we will apply these techniques  at different levels based on the stakeholders' normative systems. First, we formalize the notions of morally sensitive situations, moral decision variables (or deontic options) and moral decision problems as follows.

\begin{definition}[Morally sensitive situation]\label{def:mss}
Given a normative system $\normsys = (\mathcal{L}, \bar{ \hspace{0.1cm} }, \mathcal{R})$ and a set of facts $\mathcal{K}\subseteq \mathcal{L}$, we say that $\mathcal{K}$ is a morally sensitive situation if $bd(r) \subseteq \mathcal{K}$ for some $r \in \mathcal{R}^r \cup \mathcal{R}^c$.  
\end{definition}

The morally sensitive situations of an ethical agent are thus given in advance. This concept serves as a quick test: if $\mathcal{K}$ fails the condition of Def.~\ref{def:mss}, there is no need to provide a moral recommendation to the agent, because there are no actual ($\mathcal{R}^r$) or potential ($\mathcal{R}^c$) obligations. 

\begin{definition}[Moral decision variable]\label{def:mdv}
Given a normative system $\normsys = (\mathcal{L}, \bar{ \hspace{0.1cm} }, \mathcal{R})$, the set of all possible moral decision variables $\mathcal{DV}$ of the ethical agent is defined as: \footnote{Henceforth, we lift any function from elements to sets, so that, e.g.,~$\mathit{hd}(\mathcal{R}) = \{\mathit{hd}(r) : r \in \mathcal{R}\}$.}
\begin{center}
 $\mathcal{DV} = $ the $\subseteq$-minimal set containing $hd(\mathcal{R}^r)$ and closed under the contrariness function $\bar{ \hspace{1mm} }$. 
\end{center}
Any subset $DV\subseteq \mathcal{DV}$ closed under the contrariness function $\bar{ \hspace{1mm} }$ is a set of decision variables. 
\end{definition}

The $\mathcal{DV}$ set lists all the possible deontic options the ethical agent might handle. A specification of a subset $DV \subseteq \mathcal{DV}$ gives further control to the agent regarding what decision variables it should focus on, e.g.~for time-critical applications. 
The $DV$ set, whose specification is left open in Def.~\ref{def:mdv}, can be defined as a function of $\mathcal{K}$ depending on the application domain.\footnote{Suppose, e.g.,~that the imperativeness of each norm in $\mathcal{R}$ is measured by a ranking $\rho: \mathcal{R} \to \mathbb{N}$. Then $DV$ can be defined as the $\bar{ \hspace{1mm} }$-closure of the set $\mathit{hd}(\mathcal{R}')$ containing the $k$ most imperative    norms active in $\mathcal{K}$.}

\begin{example}\label{ex-decision-problem}
Continuing from Example~\ref{ex-norms}, let $\normsys = (\mathcal{L}, \bar{ \hspace{1mm} }, \mathcal{R})$ be the normative system of all stakeholders: $\mathcal{R} = \mathcal{R}_L \cup \mathcal{R}_H \cup \mathcal{R}_M$. Any set $\mathcal{K}$ satisfying $\mathcal{K} \cap \{w_1, w_2, w_3, w_4, i_1\} \neq \emptyset$ is a morally sensitive situation for $\normsys$. $\mathcal{DV} = \{d_1, d_2, d_3, a_1, a_2\}$ is the set of moral decision variables. The same $\mathcal{DV}$ set defines the moral choices for  $\normsys_M$, and again for $\normsys_H$ and $\normsys_L$.  
\end{example}

Given a morally sensitive situation and a set of moral options, a moral decision problem is about deciding which option(s) should be selected based on the norms of all the stakeholders as well as those of the Jiminy when needed.

\begin{definition}[Moral decision problem]
Given a normative system $\normsys$, a pair $DP=(\mathcal{K}, {DV})$ is a moral decision problem for $\normsys$ if  $\mathcal{K}$ is a morally sensitive situation and ${DV}$ is a set of  decision variables. 
\end{definition}

Given a set of normative systems and a moral decision problem, below we formulate an ASPIC-style argumentation system for checking and resolving conflicts.  Arguments are constructed from an argumentation theory that consists of a normative system as presented in Definition~\ref{def-normative-systems} and a  knowledge base $\mathcal{K}$ of brute facts shared by all the stakeholders. 

\begin{definition}[Argumentation theory for a stakeholder] \label{def-argumentation-theory-AMA}
An argumentation theory of a stakeholder $s\in \mathcal{S}$ is a tuple abusively denoted by $\normsys_s = (\mathcal{L}, \bar{ \hspace{0.1cm} }, \mathcal{R}_s, \mathcal{K})$ where $(\mathcal{L}, \bar{ \hspace{0.1cm} }, \mathcal{R}_s)$ is the normative system of $s$ and $\mathcal{K}$ is a set of observations called the {\em context}. We also call $\normsys = (\mathcal{L}, \bar{ \hspace{0.1cm} }, \mathcal{R}, \mathcal{K})$ an argumentation theory whenever $\mathcal{R} = \bigcup_{s\in\mathcal{S}}\mathcal{R}_s$. 
\end{definition}

\begin{example}\label{ex-arg-theory}
Continuing from Example~\ref{ex-norms}, 
consider the context $\mathcal{K}$ containing the brute facts $\{\DmadebyM,~ \DcollectsData,~ \DfindsThreat,~ \MregisteredNorway\}$. That is, 
\begin{flushleft}
$\begin{array}{ll}
\hspace{2mm}\mathcal{K} = &  \{w_1, w_2, w_3, w_4\}   
\end{array}$
\end{flushleft}
from which we obtain the argumentation theory  $\normsys_L = (\mathcal{L}, \bar{ \hspace{0.1cm} }, \mathcal{R}_L, \mathcal{K})$, and analogously for  $\normsys_M$, $\normsys_H$ and the system $\normsys$ of all stakeholders. Since $\mathcal{K} \cap \{w_1, w_2, w_3, w_4, i_1\} \neq \emptyset$, context $\mathcal{K}$ is a morally sensitive situation for each of these argumentation theories. We set the  pair $DP = (\mathcal{K}, DV)$ with $ DV = \mathcal{DV} = \{d_1, d_2, d_3, a_1, a_2\}$ as a moral decision problem. 
\end{example}

In this article, the notion of argument is defined in the terms of Pigozzi and Van der
Torre~\citeyear{DBLP:journals/jancl/PigozziT18}. Since we assume that all norms are defeasible, all arguments constructed from normative systems are defeasible. Moreover,  the notion of norm used in the definition corresponds to the notion of rule. 

Informally speaking,  an argument is a statement or a collection of statements that support(s) another statement. The former is called a premise (a set of premises), while the latter is called a conclusion. In a rule-based system, the conclusion of an argument can be derived from the premises using a set of rules. Following Pigozzi and Van der
Torre~\citeyear{DBLP:journals/jancl/PigozziT18}, norms are used as rules to derive conclusions. So, arguments are constructed from given normative systems that are associated with one or more stakeholders. The set of arguments follows directly from Definition 2 concerning detachment in the normative system, as each argument is a derivation corresponding to a sequence of detachments. 

\begin{definition}[Argument] \label{def-argument}
Let $\normsys = (\mathcal{L}, \bar{ \hspace{0.1cm} }, \mathcal{R}, \mathcal{K})$ be an argumentation theory. An argument $A$ for a conclusion $\mathrm{Conc}(A) = \phi$ is: \footnote{If a consequence operator $\mathit{Cn}$ is used, strict arguments must also be defined as a new type of argument. A strict argument $A_1, \ldots, A_n \to \phi$ corresponds to a deduction $\phi \in \mathit{Cn}(\{\mathrm{Conc}(A_1),\ldots, \mathrm{Conc}(A_n)\})$ from a set $\{\mathrm{Conc}(A_1),\ldots, \mathrm{Conc}(A_n)\}$ that is $\subseteq$-minimal with the property $\phi \in \mathit{Cn}(\cdot)$.}
\begin{enumerate}
\item {\bf a brute fact argument:} $\{\phi\}$ if $\phi\in \mathcal{K}$;
\item {\bf an institutional fact argument:} $A_1, \ldots, A_n\Rightarrow^c \phi$ if $A_1,\ldots, A_n$ are brute or institutional fact arguments and  $\mathrm{Conc}(A_1), \ldots, \mathrm{Conc}(A_n)\Rightarrow^c  \phi$ is a norm in $\mathcal{R}^c$; 
\item {\bf an obligation argument:} $A_1, \ldots, A_n\Rightarrow^r \phi$ if $A_1, \ldots, A_n$ are brute or institutional fact arguments such that there exists a norm $\mathrm{Conc}(A_1), \ldots, \mathrm{Conc}(A_n)\Rightarrow^r  \phi$ in $\mathcal{R}^r$;  
\item {\bf a permission argument:} $A_1, \ldots, A_n\Rightarrow^p \phi$ if $A_1,\ldots, A_n$ are brute or institutional fact arguments such that
there exists a norm $\mathrm{Conc}(A_1), \ldots, \mathrm{Conc}(A_n)\Rightarrow^p  \phi$ in $\mathcal{R}^p$. 
\end{enumerate}
$\mathit{Arg}(\normsys)$ denotes the set of all the arguments constructed from argumentation theory $\normsys$. 
\end{definition}

We define some useful functions over arguments. 
Let $A$ be an argument. The function $\mathrm{Prem}(A)$ returns the premises of argument $A$. The function
 $\mathrm{Conc}(A)$ returns the conclusion of argument $A$, and $\mathrm{Conc}(\mathcal{E})$ returns the set of conclusions $\{\mathrm{Conc}(A)\mid A\in \mathcal{E}\}$, for a set of arguments $\mathcal{E}$. The function $\mathrm{Sub}(A)$ returns the set of subarguments of $A$.  The function  $\mathrm{Norms}(A)$ returns the set of norms used in argument $A$. The function  $\mathrm{TopNorm}(A)$  returns  the top norm used in $A$. Lastly,  the function $\mathrm{Stakeholder}$ returns the set of stakeholders who supply the norms used in $A$.

\begin{definition}[Argument properties] \label{def-derivation}
For a brute fact argument $\{\phi\}$, we define $\mathrm{Prem}(\{\phi\}) = \{\phi\}$, $\mathrm{Sub}(\{\phi\}) = \{\phi\}$, $\mathrm{TopNorm}(\{\phi\}) =$ undefined, $\mathrm{Norms}(\{\phi\}) =\emptyset$, $\mathrm{Stakeholders}(A) = \emptyset$; and for an argument $A = A_1, \ldots, A_n\Rightarrow^\tau_{s} \phi$, we define:
\begin{center}
$\begin{array}{rl}\mathrm{Prem}(A) = &  \mathrm{Prem}(A_1) \cup \cdots \cup \mathrm{Prem}(A_n)\\ 
\mathrm{Sub}(A) = & \mathrm{Sub}(A_1) \cup \ldots \cup \mathrm{Sub}(A_n) \cup \{A\}\\     
\mathrm{TopNorm}(A) = & \mathrm{Conc}(A_1), \ldots, \mathrm{Conc}(A_n)\Rightarrow^\tau_{s}  \phi\\  
\mathrm{Norms}(A) = & \mathrm{Norms}(A_1) \cup \ldots \cup \mathrm{Norms}(A_n) \cup \{\mathrm{TopNorm}(A)\}\\ 
\mathrm{Stakeholders}(A) = & \mathrm{Stakeholders}(A_1) \cup \ldots \cup \mathrm{Stakeholders}(A_n) \cup \{s\}. 
\end{array}$
\end{center}
\end{definition}

\begin{definition}[Institutional facts, obligations and permissions]\label{def:classification}
The conclusions of arguments with a top constitutive rule in $\mathcal{R}^{c}$ are called \emph{institutional facts}. The conclusions of arguments with a top regulative rule in $\mathcal{R}^{r}$ are called \emph{obligations}, and those with a top permission rule in $\mathcal{R}^{p}$ are called \emph{permissions}.   
\end{definition}

Observe that for a fixed normative system $\normsys$ and context $\mathcal{K}$, the notions of institutional fact, obligation and permission in Definition~\ref{def:classification} coincide with those in Definition~\ref{def-detachment}. 

\begin{example} \label{ex-argments}
Continuing from Example~\ref{ex-arg-theory}, we may construct a brute fact argument $W_i = \{w_i\}$ for each element $w_i \in \mathcal{K}$ of the context. $W_i$ states that $w_i$ is indeed a fact. For each stakeholder $s \in \mathcal{S}$, $W_1, \ldots, W_4$ are arguments in the set $\mathit{Arg}(\normsys_s)$. Based on these arguments, we may also construct the following: 
\begin{description}
 \item[$A_{1}= (W_{1}\Rightarrow^r_L d_1)$:] The manufacturer makes the smart speaker. Hence, it should  comply with the law. 
 \item[$A_{2}= (W_{2}\Rightarrow^r_H d_2)$:]  The smart speaker is collecting user data. Hence, the privacy of the users should be protected. 
 \item[$A_{3}= (W_{3}\Rightarrow^r_H d_3)$:] The collected (user) information contains a potential critical danger to society. Hence, this information should be reported.  
 \item[$A_{4}= (W_{4}\Rightarrow^c_M i_1)$:] The manufacturer is registered in Norway. This counts as legally doing business in Norway.
  \item[$A_{5}= (W_{3}\Rightarrow^r_M a_2)$:] The collected (user) information contains a potential critical danger to society. Hence, it ought to collect information without users' explicit consent.
\end{description}

Now we can form $\mathit{Arg}(\normsys_L) = \{W_{1}, W_{2}, W_{3}, W_{4}, A_1\}$, $\mathit{Arg}(\normsys_H) = \{W_{1}, \ldots, W_{4}, A_2, A_3\}$ and  $\mathit{Arg}(\normsys_M)$ $=$ $\{W_{1}, \ldots, W_{4}, A_4, A_5\}$. Note that no argument exists (in these sets) that is built upon the regulative norm $i_1\Rightarrow^r_L a_1$ from Example~\ref{ex-norms}. 
\end{example}

Given a set of arguments, some of them might be in conflict. For instance, two obligation arguments may be in conflict if their conclusions are contradictory (or they are contraries),  meaning that both obligations cannot be accepted even if both arguments have the same priority (Def.~\ref{def:prior}). In terms of argumentation theory, we say that these two arguments defeat each other (Def.~\ref{def-defeat}). Meanwhile, when one argument defeats another argument, the latter can be defeated in turn by other arguments. So, in order to evaluate the status of arguments, one needs first to identify the defeat relation over the arguments. 

In the setting of normative systems, there are four types of propositions: the elements (called brute facts) of the context, institutional facts, obligations and permissions. As mentioned in Section~3, the notion of moral dilemma is traditionally defined as an unresolved conflict between two incompatible obligations, e.g., $Op\wedge O\neg p$ in modal logic. In terms of formal argumentation, it is represented by the existence of multiple extensions in this article. Syntactically, it means that two arguments supporting incompatible obligations defeat each other, and that neither argument has priority over the other.  Meanwhile, normative conflict is brought about by conflicting institutional facts, which may also result in multiple extensions. Normative conflicts may lead to moral dilemmas. For example, if one stakeholder believes that there has been blasphemy while another stakeholder does not (a normative conflict), the first stakeholder may deduce that we should alert the police while the second stakeholder does not (a moral dilemma). 
In addition, in accordance with the work of Pigozzi and Van der Torre~\citeyear{DBLP:journals/jancl/PigozziT18}, two permissive norms never conflict, and a permissive norm is not in conflict with a brute fact or an institutional fact. Based on the these considerations, the notions of priority relation and defeat relation between arguments are defined as follows. 
\begin{figure}[ht]
    \centering
    \begin{tikzpicture}
    \node[Core] (a1) at (-3.75,0) {\footnotesize Constitutive norms};
\node[Core] (a2) at (-3.75,-2) {\footnotesize Regulative norms};
\node[Core] (a3) at (-3.75, -4) {\footnotesize Permissive norms};
\node[Core] (b1) at (6,0) {\footnotesize Brute fact arguments};
\node[Core] (b2) at (4.25,-2.25) {\footnotesize Institutional fact arguments};
\node[Core] (b3) at (7.75,-2.25) {\footnotesize Obligation arguments};
\node[Core] (b4) at (6,-4.5) {\footnotesize Permission arguments};

\draw[draw=black] (2.45,1.25) rectangle (9.55,-5.55);
\node (text) at (6,.95) {\textbf{\textsf{Normative arguments}}};

\node[Core] (c) at (0,-2.25) {\footnotesize Dilemma resolving arguments};
\path[->,-stealth]
(a1) edge (a2)
(a3) edge (a2)
(b1) edge (b2)
(b1) edge (b3)
(b2) edge (b3)
(b4) edge (b3)
(c) edge (2.45,-2.25)
;
\end{tikzpicture}
    \caption{The priority order over different type of norms and arguments. An exiting arrow indicates a higher priority. Dilemma resolving arguments are defined later on.}
    \label{fig:prior}
\end{figure}

Regarding priority over arguments, in accordance with the normative theory introduced in Section~3, constitutive norms always override regulative norms (otherwise we have wishful thinking) as do permissive norms (as they encode exceptions to regulations). So, an institutional fact argument may defeat an obligation argument, a permission argument may defeat an obligation argument, and a brute fact argument may defeat an institutional fact argument or an obligation argument, but not vice versa.  In addition, brute fact arguments have the highest priority. This is illustrated in Figure~\ref{fig:prior} and specified in Definition~\ref{def:prior}. Dilemma resolving arguments will be introduced later on.

\begin{definition}[Priority relation between arguments]\label{def:prior}
Let $\mathcal{A}$ be a set of arguments, e.g., $\mathcal{A} = \mathit{Arg}(\mathcal{N})$ for some $\mathcal{N}$. Let $\mathcal{A}^b, \mathcal{A}^c, \mathcal{A}^r, \mathcal{A}^p \subseteq \mathcal{A}$ be the sets of brute fact arguments, institutional fact arguments, obligation arguments and permission arguments respectively.  Given two arguments $A,B\in \mathcal{A}$, we use $A\succeq B$ to denote that $A$ is non-strictly preferred to $B$ and $A\succ B = (A \succeq B \mbox{ and } B \nsucceq A)$ to denote that $A$ is preferred to $B$. We have: 
\begin{itemize}
\item For $A, B \in \mathcal{A}^\tau$, $A \succeq B$ \quad  any $\tau \in \{b,c,r\}$.
\item For $A\in \mathcal{A}^b$ and $B\in \mathcal{A}^c \cup \mathcal{A}^r$, $A\succ B$. \footnote{A condition one might wish to impose upon language $\mathcal{L}$ is that brute facts are disjoint from institutional facts, so that priorities of type $\mathcal{A}^b \succ \mathcal{A}^c$ would not be needed. In practice, this condition cannot always be enforced---see the discussion by Pigozzi and Van der Torre~\citeyear[Sec.~4.3]{DBLP:journals/jancl/PigozziT18}.} 
\item For $A\in \mathcal{A}^c$ and $B\in \mathcal{A}^r$, $A\succ B$.
\item For $A\in \mathcal{A}^p$ and $B\in \mathcal{A}^r$, $A\succ B$.
\end{itemize}
\end{definition}

For a fixed type $\tau \neq p$, any two arguments $A, B \in \mathcal{A}^\tau$ are equally preferred: $A \succeq B \succeq A$ (and for $\tau = p$, they are incomparable: $A \nsucceq B \nsucceq A$). If desired, a strict preference can be enforced between them, e.g., $A \succ' B$, with a more refined priority relation $\succ' \,\supset\, \succ$. For instance, one obligation argument may be preferred to another. Since these additional priorities are context dependent, and considering this relation does not affect the main point of our approach to checking and resolving moral dilemmas, we leave the priority relation within any argument type abstract.

Next, we define what it means for arguments to attack and defeat each other.\footnote{Attacks represent logical conflicts based on the contrariness relation $\bar{ \hspace{1mm} }$. While conceptually there is no real conflict between a permission to $p$ and a fact $p$ or permission to $\neg p$, we keep the definition of attack simple and manage this lack of real conflict via incomparability under the priority relation $\succeq$. Reverse attacks in  Definition~\ref{def-defeat} have been proposed in ABA+~\cite{cyras} at the level of assumptions.}

\begin{definition}[Attacks and defeats] \label{def-defeat}
Let $\mathcal{A}$ be a set of arguments. For any $A, B \in \mathcal{A}$, $A$ attacks $B$ iff $\mathrm{Conc}(A) \in \overline{\phi}$ for some $B^\prime \in \mathrm{Sub}(B)$ and $\mathrm{Conc}(B^\prime) =\phi$. In this case, we say that $A$ attacks $B$ at $B'$. We say that $A$ defeats $B$ iff: 
\begin{itemize}
\item $A$ attacks $B$ at $B'$ and $A \succeq B'$ \hfill (direct defeat), or 
\item $B$ extends some $B' \in \mathrm{sub}(B)$ that attacks $A$ at $A$ and $B' \prec A$. \hfill (reverse defeat)
\end{itemize}
The set of defeats over the arguments in $\mathit{Arg}(\normsys)$ built from an argumentation theory $\normsys$ is denoted $\mathit{Def}(\normsys)$. We will also depict any defeat $(A, B) \in \mathit{Def}(\normsys)$ as an arrow $A \longrightarrow B$.
\end{definition}	

\begin{definition}[Argumentation frameworks: individual] 
\label{def-indfram}
For each stakeholder $s \in \mathcal{S}$,
\begin{flushleft}
$\begin{array}{ll@{\hspace{1.3cm}}l}
AF(\normsys_s) = &  (\mathit{Arg}(\normsys_s), \mathit{Def}(\normsys_s)) & \mbox{ is the individual argumentation framework of $s$.} 
\end{array}$
\end{flushleft}
\end{definition}

\begin{example} \label{ex-afs-st0}
Continuing from Example~\ref{ex-argments}, consider the arguments and defeats in the framework $AF(\normsys_H)$. We have  $\mathit{Arg}(\normsys_H) = \{A_2, A_3\}$ and $\mathit{Def}(\normsys_H)= \{(A_2, A_3)\}$. Let us check this. 
\begin{description}
\item[$(A_2 \to A_3)$] The conclusions satisfy $d_2 \in \overline{d_3}$ so $A_2$ attacks $A_3$. With $A_2 \succeq A_3$, this is a defeat.  
\end{description}
This is depicted in Figure~\ref{fig-IAFindividual} below, together with the frameworks of stakeholders $L$ and $M$, which contain no defeats. See Example~\ref{ex-afs-st1} for an illustration of a reverse defeat. 
\end{example}

In terms of the work of Dung~\citeyear{DBLP:journals/ai/Dung95}, a set of collectively acceptable arguments in an argumentation framework is called an \textit{extension}. A core notion that supports the definition of various extensions is \textit{admissible sets}. Specifically, given an argumentation framework $AF=(\mathit{Arg}, \mathit{Def})$, a set $\mathcal{E}\subseteq \mathit{Arg}$ of arguments is \textit{admissible} if and only if it is \textit{conflict-free} and \textit{defends} each argument within this set $\mathcal{E}$. A set $\mathcal{E}$ is \emph{conflict-free} if and only if there are no arguments $A$ and $B$ in $\mathcal{E}$ such that $(A,B)\in \mathit{Def}$. Argument $A\in \mathit{Arg}$ is \textit{defended} by a set $\mathcal{E}\subseteq \mathit{Arg}$ (also expressed as $A$ is \emph{acceptable} with respect to $\mathcal{E}$) if and only if for all $B\in \mathit{Arg}$, if $(B,A)\in \mathit{Def}$, then there exists an argument $C\in \mathcal{E}$ such that $(C,B)\in \mathit{Def}$. Based on the notion of admissible sets, some other extensions could be defined. Formally, we have the following definition from Dung~\citeyear{DBLP:journals/ai/Dung95}:

\begin{definition}[{Argumentation semantics}] \label{Def-AF-conflict free}
{Let $AF=(\mathit{Arg}, \mathit{Def})$ be an argumentation framework, and let $\mathcal{E}\subseteq \mathit{Arg}$ be a set of arguments. We have that:}
\begin{itemize}
  \item {$\mathcal{E}$ is {admissible} iff $\mathcal{E}$ is conflict-free and each argument in $\mathcal{E}$ is defended by $\mathcal{E}$.}
  \item {$\mathcal{E}$ is a complete extension iff $\mathcal{E}$ is admissible and each argument in $\mathit{Arg}$ that is defended by $\mathcal{E}$ is in $\mathcal{E}$.}
 \item {$\mathcal{E}$ is a preferred extension iff $\mathcal{E}$ is a maximal complete extension (w.r.t. set inclusion).}
  \item {$\mathcal{E}$ is a grounded extension iff $\mathcal{E}$ is the minimal complete extension (w.r.t. set-inclusion).}
  \item $\mathcal{E}$ is a stable extension iff $\mathcal{E}$ is conflict-free, and for each argument $A\in \mathit{Arg}\setminus \mathcal{E}$, there exists an argument $B\in \mathcal{E}$, such that $(B,A)\in \mathit{Def}$.
\end{itemize}
We use $\sigma\in \{\mathit{co}, \mathit{pr}, \mathit{gr}, \mathit{st}\}$ to indicate the complete, preferred, grounded, and stable semantics. For each of these semantics, $\sigma(AF)$ denotes the set of $\sigma$-extensions of $AF$.\footnote{The choice of a semantics $\sigma$ has a major  impact on which moral recommendations are passed to the agent. In the running example, we opt for maximal acceptance under the preferred semantics. The grounded semantics offers efficiency and extension uniqueness, but it easily leads to empty recommendations. Stable extensions offer solid justifications each but they might not exist at all, in which case no recommendation can be sent.}
\end{definition}

Recall that in the semantics $\sigma\in \{\mathit{co}, \mathit{pr}, \mathit{gr}, \mathit{st}\}$, their extensions are complete: $\sigma(AF) \subseteq \mathit{co}(AF)$; hence they are also admissible: $\mathcal{E} \in \sigma(AF)$ implies that $\mathcal{E}$ defends all of $\mathcal{E}$. A semantic extension can be seen as one way to solve all disagreements at once, be it between institutional facts (normative conflicts), between obligations (moral dilemmas) or between permissions and obligations (permission-obligation conflicts).

\begin{example}\label{ex-IAF-extensions} Continuing from Example~\ref{ex-afs-st0}, each stakeholder's framework has a unique preferred extension $\mathit{pr}(AF(\normsys_s)) = \{\mathcal{E}_s\}$.\footnote{$\mathcal{E}_s$ also happens to be the unique complete, grounded and stable extension of $AF(\normsys_s)$.} See also Figure~\ref{fig-IAFpreferred} below. The accepted arguments are: 
\begin{center}
$\begin{array}{lll}\mathcal{E}_L = \{A_1\} = \mathit{Arg}(\normsys_L) &   \mathcal{E}_M = \{A_4, A_5\} = \mathit{Arg}(\normsys_M) &  \mathcal{E}_H = \{A_2\} = \mathit{Arg}(\normsys_M) \setminus \{A_3\}\\
\mathrm{Conc}: \{\mathit{law}\} & \hspace{.5cm} \{\mathit{business}, \mathit{collect}\}  & \hspace{.65cm} \{\mathit{protect}\}
\end{array}$
\end{center}
Only in the case of conflict, i.e.,~within $\normsys_H$, is a set of self-defending arguments selected. 
\end{example}

Let us study the formal properties of the argumentation system. We focus first on the types of argument where the presence of their contraries \emph{always} results in conflict, e.g.~factual, institutional and obligation arguments. For any extension $\mathcal{E}$, we call the set $\recommender{E} = \mathcal{E} \setminus \mathcal{A}^p$ the \recommendation{} of $\mathcal{E}$.

\begin{lemma}[Rationality postulates]\label{rrpp} The rationality postulates~\cite{caminada-amgoud} hold for the \recommendation{} $\recommender{E}$ of any $\sigma$-extension $\mathcal{E}$ under $\sigma\in \{\mathit{co}, \mathit{pr}, \mathit{gr}, \mathit{st}\}$. That is, (direct consistency) $\mathrm{Conc}(\recommender{E})$ is consistent, (subargument closure) $\mathrm{sub}(\recommender{E}) \subseteq \recommender{E}$ and, moreover, $\mathrm{sub}(\mathcal{E}) \subseteq \mathcal{E}$. The remaining postulates trivially hold.
\end{lemma}

The direct consistency result can be extended to all obligation-permission pairs.
\begin{fact}
For any $\sigma$-extension $\mathcal{E}$ with $\sigma\in\{\mathit{co,pr,gr,st}\}$, the set $\mathrm{Conc}((\mathcal{E} \cap \mathcal{A}^r) \cup \{B\})$ is consistent when $B$ is an arbitrary permission argument in the extension, i.e., $B \in \mathcal{E} \cap \mathcal{A}^p$.
\end{fact}  

The correspondence between (the outputs of) semantic extensions and norm extensions is at best partial in the sense of each complete extension  \emph{being contained in} a norm extension. Only the outputs of stable extensions match those of a norm extension.      

\begin{proposition}\label{prop-inclusions}
Let $\normsys = (\mathcal{L}, \bar{ \hspace{0.1cm} }, \mathcal{R}, \mathcal{K})$ be an argumentation theory with $\mathcal{R} = \mathcal{R}^c\cup\mathcal{R}^r\cup\mathcal{R}^p$. (1) For any extension $\mathcal{E}\in \sigma(AF(\normsys))$ under $\sigma\in \{\mathit{co}, \mathit{pr}, \mathit{gr}, \mathit{st}\}$ there exists $(M^c,M^r,M^p)$, a norm extension of $(\mathcal{R}^c,\mathcal{R}^r,\mathcal{R}^p)$ in context $\mathcal{K}$,  such that:
\begin{itemize}
    \item[(i)] $I(M^c,\mathcal{K}) \supseteq \mathrm{Conc}(\mathcal{E}\cap \mathcal{A}^c) \cup \mathcal{K}$; 
    \item[(ii)] $O(M^c,M^r,\mathcal{K}) \supseteq \mathrm{Conc}(\mathcal{E}\cap \mathcal{A}^r)$; and
    \item[(iii)] $P(M^c,M^p,\mathcal{K})  \supseteq \mathrm{Conc}(\mathcal{E}\cap \mathcal{A}^p)$.
\end{itemize}
(2) For the stable case $\sigma = \mathit{st}$, the inclusions in (i)--(iii) are in fact identities: $(i')~ I(M^c,\mathcal{K}) = \mathrm{Conc}(\mathcal{E}\cap \mathcal{A}^c) \cup \mathcal{K}$;~ $(ii')~O(M^c,M^r,\mathcal{K}) = \mathrm{Conc}(\mathcal{E}\cap \mathcal{A}^r)$;~ $(iii')~P(M^c,M^p,\mathcal{K}) = \mathrm{Conc}(\mathcal{E}\cap \mathcal{A}^p)$.
\end{proposition}

Conversely to Prop.~\ref{prop-inclusions}(1), certain conditions verify that all $P$-maximal norm extensions $M$ contain (the rules of) some extension $\mathcal{E}$. This result holds for those semantics that grant the existence of extensions $\sigma \in\{\mathit{co}, \mathit{pr},\mathit{gr}\}$ but cannot be proved in general under the stable semantics. A condition verifying this claim is the symmetry of contraries: $\phi \in \overline{\psi}$ if{f} $\psi \in \overline{\phi}$.    

\begin{proposition}\label{symmetry}
Let $\mathcal{N} = (\mathcal{L}, \bar{ \hspace{1mm} }, \mathcal{R}, \mathcal{K})$ be an argumentation theory with a symmetric contrariness function $\bar{ \hspace{1mm} }$. Then, any $P$-maximal norm extension $M= (M^c,M^r,\mathcal{R}^p)$ in $\mathcal{K}$ contains a $\sigma$-extension $\mathcal{E}$ in the sense of elements (i)--(iii) of Prop.~\ref{prop-inclusions} for any $\sigma \in \{\mathit{co}, \mathit{pr}, \mathit{gr}\}$. 
\end{proposition}

\begin{definition}[Naive semantics] An argumentation semantics not based on the idea of admissibility is the naive semantics $\sigma = \mathit{na}$, defined as:
%\begin{itemize}
%\item 
\begin{center}
$\mathcal{E}$ is a naive extension, denoted as $\mathcal{E} \in \mathit{na}(AF)$, if{f} the set $\mathcal{E}$ is $\subseteq$-maximally conflict-free.
%\end{itemize}
\end{center}
\end{definition}
The naive semantics provides better correspondence with norm extensions, as these notions are defined by maximal conflict-freeness and consistency respectively. In fact, a 1-1 correspondence exists between naive and norm extensions at the level of triggered rules.

\begin{proposition}\label{naive-prop} Let $\mathcal{N} = (\mathcal{L}, \bar{ \hspace{1mm} }, \mathcal{R}, \mathcal{K})$ be an argumentation theory inducing the argumentation framework $AF = (\mathit{Arg}(\mathcal{N}),\mathit{Def}(\mathcal{N}))$. 
(1) For any naive extension $\mathcal{E} \in \mathit{na}(AF)$, it holds that $\mathcal{E} = \mathit{Arg}(\mathcal{N}_M)$ for some norm extension $M$ in $\mathcal{K}$. 
(2) For any $P$-maximal norm extension $M$ in $\mathcal{K}$, the set $\mathcal{E}_M = \mathit{Arg}(\mathcal{N}_M)$ is a naive extension: $\mathcal{E}_M
\in \mathit{na}(AF)$.
\end{proposition}

\section{Argumentation for Reaching Moral Agreements among Stakeholders}\label{sec:agreements}

Let us put this argumentation system to work towards moral agreements so that a decision can be passed to the agent. As this decision ultimately concerns conflicts of obligations or moral dilemmas, the four-level system will, in particular, seek and address such dilemmas between different semantic extensions. Normative conflicts (like all the disagreements) are semantically resolved \emph{within} each extension but are not directly addressed \emph{between} extensions. They can be resolved when addressing moral dilemmas, but they are not the Jiminy's primary concern. For an extension $\mathcal{E}$, let its set of obligations be 
\begin{center} 
$\mathit{Obl}(\mathcal{E}) = \{\mathrm{Conc}(A) \mid A\in \mathcal{E}\cap \mathcal{A}^r\}$
\end{center}
where again $\mathcal{A}^r \subseteq \mathit{Arg}(\normsys)$ is the set of obligation arguments built from a theory $\normsys$.

\begin{definition}[Moral dilemma]\label{def:dilemma}
Let $\mathcal{C}$ be a collection of argument extensions for some decision problem $DP = (\mathcal{K}, {DV})$. We say that $\mathcal{C}$ contains a moral dilemma if a pair of obligation arguments $A, B$ exist in some extensions of $\mathcal{C}$, say $A \in \mathcal{E}_1 \in \mathcal{C}$ and $B \in \mathcal{E}_2\in \mathcal{C}$, such that $\mathrm{Conc}(A) \in \overline{\mathrm{Conc}(B)}$ and these two contrary obligations are in ${DV}$. In other words, there exist $\mathcal{E}_1, \mathcal{E}_2 \in \mathcal{C}$ such that 
\begin{center} $(\mathit{Obl}(\mathcal{E}_1)\cup \mathit{Obl}(\mathcal{E}_2))\cap DV$ is inconsistent with respect to the contrariness function $\bar{ \hspace{0.1cm}}$.
\end{center}
\end{definition}

Given the argumentation theory $\normsys_s = (\mathcal{L}, \bar{ \hspace{0.1cm} }, \mathcal{R}_s, \mathcal{K})$ of each stakeholder $s \in S$,  and a decision problem, we distinguish four ways (i.e., four collections $\mathcal{C}$ of extensions) to check whether there is a dilemma and, if so, whether it can be solved at the next level:
\begin{enumerate}
\item First, we consider the normative system of each stakeholder independently. In this case, we compute the extensions of the corresponding argumentation frameworks and check whether there is a dilemma between the extensions of one or more  stakeholders.    
\item Second, we consider the arguments of all the stakeholders together. In this case, we construct a single argumentation framework and check whether it contains  dilemmas. Each argument still consists of norms from  one single stakeholder. 
\item Third, we put all the normative systems together into a unified argumentation theory and check whether it contains dilemmas. Arguments now combine norms from different stakeholders.
\item Fourth, we use the Jiminy to decide which stakeholders are the most competent for each dilemma.\footnote{The source of the Jiminy's priorities is domain specific. We assume that the set of norms in the Jiminy is given.}  
\end{enumerate}

At any of these four levels, if no dilemma is found, the Jiminy submits as its moral recommendation the set of obligations occurring in at least one of the semantic extensions. 

\begin{definition}[Moral recommendation at $i$-th level]
If no dilemma exists in the set of extensions $\{\mathcal{E}_1, \ldots, \mathcal{E}_k\}$ at level $i$, then the moral recommendation or output at $i$ is: $\mathit{Obl}(\mathcal{E}_1) \cup \ldots\cup \mathit{Obl}(\mathcal{E}_k)$.
\end{definition}

Despite obligation arguments having the lowest priority among arguments, our method for checking dilemmas makes the system credulous about obligations (it accepts an obligation if it belongs to at least one extension) and skeptical about institutional facts and permissions.

\subsection{First-Level Dilemmas: Individual Frameworks.}
Let us fix an argumentation theory $\mathcal{N}_{s} = (\mathcal{L}, \bar{ \hspace{1mm} }, \mathcal{R}_{s}, \mathcal{K})$ for each stakeholder $s \in \mathcal{S}$. 

\begin{definition}[Individual Frameworks $\mathcal{I\hspace{-.5mm}F}$] \label{def-argumentation-theory-IF}
The set of \emph{individual frameworks} is \begin{center}$\mathcal{I\hspace{-.5mm}F} = \{ AF(\normsys_s) : s \in \mathcal{S}\}$\end{center}
where  $AF(\normsys_{s}) = (\mathit{Arg}(\normsys_s), \mathit{Def}(\normsys_s))$ is the individual framework of stakeholder $s \in \mathcal{S}$. 
\end{definition}

For reference, let us define a zero-level dilemma as any moral dilemma between a pair of (unfiltered) arguments from stakeholders $\mathcal{C}_0 = \{\mathit{Arg}(\mathcal{N}_{s_0}), \ldots, \mathit{Arg}(\mathcal{N}_{s_0})\}$. The four levels of dilemma checking will make use of argumentation semantics to filter out some of these arguments and the moral dilemmas they contribute to. Let us stress that dilemmas depend on the contrariness function rather than  defeat relations.

\begin{definition}[Checking for and resolving a dilemma in an $\mathcal{I\hspace{-.5mm}F}$]\label{def-first-level-dilemma}
Let $DP=(\mathcal{K}, DV)$ be a decision problem and $\sigma$ an argumentation semantics: $\sigma \in \{\mathit{co},\mathit{gr},\mathit{pr},\mathit{st}\}$.  A first-level (or $\mathcal{I\hspace{-.5mm}F}$) dilemma relative to $DP$ under $\sigma$ is any moral dilemma in collection $\mathcal{C}_1 = \sigma(AF(\normsys_{s_0})) \cup \ldots \cup \sigma(AF(\normsys_{s_n}))$. That is, an $\mathcal{I\hspace{-.5mm}F}$ dilemma exists if there are $\mathcal{E}_1\in \sigma(AF), \mathcal{E}_2 \in \sigma(AF^\prime)$ for some $AF, AF^\prime \in \mathcal{I\hspace{-.5mm}F}$ such that  
\begin{center}
$(\mathit{Obl}(\mathcal{E}_1)\cup \mathit{Obl}(\mathcal{E}_2))\cap DV$ is inconsistent with respect to the contrariness function $\bar{ \hspace{0.1cm}}$.
\end{center}
Otherwise, if for all $\mathcal{E}_1\in \sigma(AF), \mathcal{E}_2 \in \sigma(AF^\prime)$, it holds that $(\mathit{Obl}(\mathcal{E}_1)\cup \mathit{Obl}(\mathcal{E}_2))\cap DV$ is consistent (possibly empty), then there is no $\mathcal{I\hspace{-.5mm}F}$ dilemma and all zero-level dilemmas have been resolved at the first level by $\sigma$.
\end{definition}

\begin{figure}[t]
    \centering
\begin{subfigure}{.25\textwidth}
\begin{tikzpicture}[framed]
\node[Oarg] (a1) at (-1,0) {\footnotesize $A_1$};

\node[Iarg] (a4) at (1,-1.8) {\footnotesize \phantom{$c$}}; 
\node (text4) at (1,-1.8) {\footnotesize $A_4$};
\node[Oarg] (a5) at (.6,0) {\footnotesize $A_5$};

\node (textAF) at (0, -4) {\footnotesize $\normsys_H$};
\node (textAF2) at (-.8, -2) {\footnotesize $\normsys_L$};
\node (textAF3) at (1.4, .25) {\footnotesize $\normsys_M$};

\node (conc1) at (-1,-.6) {\footnotesize \emph{law}}; 
\node (conc2) at (.7,-.6) {\footnotesize \emph{collect}}; 
\node (conc3) at (1.05,-1) {\footnotesize \emph{business}}; 

\node (conc4) at (-1,-3.6) {\footnotesize \emph{protect}}; 
\node (conc5) at (1,-3.6) {\footnotesize \emph{report}}; 

\node[Oarg] (a2) at (-1,-3) {\footnotesize $A_2$};
\node[Oarg] (a3) at (1,-3) {\footnotesize $A_3$};

\draw[dotted,thick] (-1.5,-2.3) -- (1.7,-2.3);
\draw[dotted,thick] (0,.5) -- (0,-2.2);

\path[->,bend left=25,-stealth]
;
\path[->,-stealth]
(a2) edge (a3)
;
\end{tikzpicture}
\caption{Individual~frameworks}\label{fig-IAFindividual}
\end{subfigure}
\begin{subfigure}{.24\textwidth}
\begin{tikzpicture}[framed]
\node[OargG] (a1) at (-1,0) {\footnotesize $A_1$};
\node[IargG] (a4) at (1,-1.8) {\footnotesize \phantom{$c$}};   
\node (text4) at (1,-1.8) {\footnotesize $A_4$};
\node[OargG] (a5) at (.6,0) {\footnotesize $A_5$};
\node (textAF) at (0, -4) {\footnotesize \phantom{${\mathcal{S}_0}$}};
\node[OargG] (a2) at (-1,-3) {\footnotesize $A_2$};
\node[Oarg] (a3) at (1,-3) {\footnotesize $A_3$};

\node (textAF) at (-.5, -3.6) {\footnotesize $\mathcal{E}_H$};
\node (textAF2) at (-.8, -.85) {\footnotesize $\mathcal{E}_L$};
\node (textAF3) at (1, -.85) {\footnotesize $\mathcal{E}_M$};

\draw[dotted,thick] (-1.5,-2.3) -- (1.7,-2.3);
\draw[dotted,thick] (0,.5) -- (0,-2.2);

\path[->,-stealth]
(a2) edge (a3)
;
\end{tikzpicture}
\caption{Preferred extensions}\label{fig-IAFpreferred}
\end{subfigure}
\begin{subfigure}{.23\textwidth}
\begin{tikzpicture}[framed]
\node[Oarg] (a1) at (-1,0) {\footnotesize $A_1$};
\node[Iarg] (a4) at (1,-1.8) {\footnotesize \phantom{$c$}}; 
\node (text4) at (1,-1.8) {\footnotesize $A_4$};
\node[Oarg] (a5) at (.6,0) {\footnotesize $A_5$};
\node (textAF) at (0, -4) {\footnotesize \phantom{${\mathcal{S}_0}$}};
\node[Oarg] (a2) at (-1,-3) {\footnotesize $A_2$};
\node[Oarg] (a3) at (1,-3) {\footnotesize $A_3$};

\draw[dotted,thick,white] (-1.5,-2.5) -- (1.6,-2.5);
\draw[dotted,thick,white] (0,.5) -- (0,-2.4);

\path[->,-stealth,dotted,thick]
(a2) edge (a3)
(a5) edge (a1)
(a5) edge (a2)
(a2) edge (a5)
(a1) edge (a2)
(a2) edge (a1)
;
\end{tikzpicture}
\caption{Contraries}\label{fig-IAFcontraries}
\end{subfigure}
\begin{subfigure}{.23\textwidth}
\begin{tikzpicture}[framed]
\node[OargG] (a1) at (-1,0) {\footnotesize $A_1$};
\node[IargG] (a4) at (1,-1.8) {\footnotesize \phantom{$c$}};   
\node (text4) at (1,-1.8) {\footnotesize $A_4$};
\node[OargG] (a5) at (.6,0) {\footnotesize $A_5$};
\node (textAF) at (0, -4) {\footnotesize \phantom{${\mathcal{S}_0}$}};
\node[OargG] (a2) at (-1,-3) {\footnotesize $A_2$};
\node[Oarg] (a3) at (1,-3) {\footnotesize $A_3$};

\draw[dotted,thick,white] (-1.5,-2.5) -- (1.6,-2.5);
\draw[dotted,thick,white] (0,.5) -- (0,-2.4);

\path[-,dotted,thick]
(a5) edge (a1)
(a5) edge (a2)
(a2) edge (a5)
(a1) edge (a2)
(a2) edge (a1)
;
\path[->,-stealth]
(a2) edge (a3)
;
\end{tikzpicture}
\caption{Moral dilemmas}\label{fig-IAFdilemmas}
\end{subfigure}
\caption{Individual frameworks and their properties. Context arguments $W_{1}, \dots, W_{4}$ are omitted. Obligation and institutional arguments are represented as circles and triangles respectively, and are labeled with their conclusions. The subfigures depict: (a) the individual frameworks $AF(\normsys_L), AF(\normsys_M)$ and $AF(\normsys_H)$; the solid arrow represents the only defeat, in $\mathit{Def}(\normsys_H)$; (b) the preferred extension (in gray) of each framework; (c) the contrariness function $\bar{ \hspace{1mm} }$ over conclusions, as dotted arrows; (d) pairs of accepted  contrary obligations, which are moral dilemmas (dotted lines).}   
    \label{fig:IFs}
\end{figure}

\begin{example} \label{ex-first-level-un}
Continuing from Example~\ref{ex-IAF-extensions}, the preferred extension for each stakeholder (Figure~\ref{fig-IAFpreferred}) contains one obligation:
\begin{center}
$\mathit{Obl}(\mathcal{E}_L): \mathrm{Conc}(A_1) = d_1 \qquad \mathit{Obl}(\mathcal{E}_M):  \mathrm{Conc}(A_5) = a_2 \qquad \mathit{Obl}(\mathcal{E}_H): \mathrm{Conc}(A_2)=d_2$.
\end{center}

Since the problem $DP= (\mathcal{K}, {DV})$ was defined in Example~\ref{ex-decision-problem} as  ${DV} = \{d_1, d_2, d_3, a_1, a_2\}$, a first-level dilemma exists between each pair of extensions (see also Figure~\ref{fig-IAFdilemmas2}): 
\begin{center}
$\{\mathcal{E}_L, \mathcal{E}_H\}:$\quad $d_1 = -d_2$; \qquad\quad $\{\mathcal{E}_M,\mathcal{E}_L\}:$\quad $a_2 \in \overline{d_1}$  \qquad\quad $\{\mathcal{E}_M,\mathcal{E}_H\}$:\quad $a_2= -d_2$.
\end{center}
\end{example}

\subsection{Second-Level Dilemmas: Combined Framework.} First-level dilemmas are addressed at the second level by combining the arguments of all the stakeholders into one set $\mathit{Arg}({\mathcal{S}})$ and then computing the defeat relation on that set.

\begin{definition}[Argumentation frameworks: combined] 
\label{def-comfram}
Gathering all arguments from stakeholders $\mathcal{S} = \{s, \ldots,\}$ defines the set $\mathit{Arg}({\mathcal{S}}) = \bigcup_{s\in \mathcal{S}} \mathit{Arg}(\normsys_{s})$. By letting $\mathit{Def}({\mathcal{S}})$ be the defeat relation over this set, we obtain
\begin{flushleft}
$\begin{array}{ll@{\hspace{1.49cm}}l}
AF({\mathcal{S}}) = & (\mathit{Arg}({\mathcal{S}}), \mathit{Def}({\mathcal{S}})) &  \mbox{ the combined argumentation framework.}
\end{array}$
\end{flushleft}
\end{definition}

\begin{fact}
Given a set of individual argumentation frameworks $\mathcal{I\hspace{-.5mm}F} = \{AF(\normsys_{s})\mid s\in \mathcal{S}_0\}$ and the combined argumentation framework $AF({\mathcal{S}_0}) = (\mathit{Arg}({S_0}), \mathit{Def}({\mathcal{S}_0}))$ at the second level, it holds that $\mathit{Def}({\mathcal{S}_0}) \supseteq \bigcup_{s\in \mathcal{S}_0}\mathit{Def}(\normsys_s)$.
\end{fact}

\begin{proof}
If $A$ defeats $B$ when $(A,B)\in  \mathit{Def}(\normsys_{s})$ for all $s\in \mathcal{S}$,  $A$ still defeats $B$ when $A$ and $B$ are in $\mathit{Arg}_{\mathcal{S}}$ in accordance with Definition~\ref{def-defeat}. According to Definition~\ref{def-comfram}, $(A,B)\in \mathit{Def}({\mathcal{S}})$. So, it holds that $\mathit{Def}({\mathcal{S}}) \supseteq \bigcup_{s\in \mathcal{S}} \mathit{Def}(\normsys_s)$. 
\end{proof}

\begin{example} \label{ex-afs-st1}
Continuing from Example~\ref{ex-argments}, recall the defeat $A_2 \to A_3$ within the framework $AF(\normsys_H)$. The combined framework $AF(\mathcal{S})$ adds the following defeats: 
\begin{description} 
\item[$(A_1 \leftrightarrow A_2)$] $\lawCompliantM$ \emph{vs.}~$\privacyProtectD$:  
$d_1 = -d_2$ and $A_1 \preceq\succeq A_2$; 
\item[$(A_5 \to A_1)$]  $\toCollectDataNEP$ defeats $\lawCompliantM$: $a_2  \in  \overline{d_1}$ and $A_5 \preceq\succeq A_1$; (for a defeat $A_1 \to A_5$ we would need $d_1 \in \overline{a_2}$);
\item[($A_5 \leftrightarrow A_2$)] $\toCollectDataNEP$ \emph{vs.}~$\privacyProtectD$: $a_2 = -d_2$ and $A_5 \preceq\succeq A_2$.  
\end{description}
In summary, $\mathit{Def}(\mathcal{S}) = \{(A_2, A_3), (A_1, A_2), (A_2, A_1), (A_5, A_1), (A_5, A_2)\}$. The combined framework $AF(\mathcal{S})$ and its defeats are shown in Figure~\ref{fig-combinedC} and Figure~\ref{fig-combinedCC}. 
\end{example}

\begin{figure}[t]
    \centering
\begin{subfigure}{.255\textwidth}
\begin{tikzpicture}[framed]
\node[Oarg] (a1) at (-1,0) {\footnotesize $A_1$};
\node[Iarg] (a4) at (1,-1.8) {\footnotesize \phantom{$c$}};   
\node (text4) at (1,-1.8) {\footnotesize $A_4$};
\node[Oarg] (a5) at (.6,0) {\footnotesize $A_5$};
\node (textAF) at (0, -4) {\footnotesize \phantom{${\mathcal{S}_0}$}};
\node[Oarg] (a2) at (-1,-3) {\footnotesize $A_2$};
\node[Oarg] (a3) at (1,-3) {\footnotesize $A_3$};

\node (conc1) at (-.6,-.6) {\footnotesize \emph{law}}; 
\node (conc2) at (.8,-.6) {\footnotesize \emph{collect}}; 
\node (conc3) at (1.05,-1) {\footnotesize \emph{business}}; 
\node (conc4) at (-1,-3.6) {\footnotesize \emph{protect}}; 
\node (conc5) at (1,-3.6) {\footnotesize \emph{report}}; 

\draw[dotted,thick,white] (-1.5,-2.5) -- (1.6,-2.5);
\draw[dotted,thick,white] (0,.5) -- (0,-2.4);

\path[-,dotted,thick]
(a5) edge (a1)
(a5) edge (a2)
(a2) edge (a5)
(a1) edge (a2)
(a2) edge (a1)
;
\path[->,-stealth]
(a2) edge (a3)
;
\end{tikzpicture}
\caption{Level 1 dilemmas}\label{fig-IAFdilemmas2}
\end{subfigure}
\begin{subfigure}{.24\textwidth}
\begin{tikzpicture}[framed]
\node[Oarg] (a1) at (-1,0) {\footnotesize $A_1$};
\node[IargG] (a4) at (1,-1.8) {\footnotesize \phantom{$c$}};   
\node (text4) at (1,-1.8) {\footnotesize $A_4$};
\node[OargG] (a5) at (.6,0) {\footnotesize $A_5$};
\node (textAF) at (0, -4) {\footnotesize ${\mathcal{E}_1}$};
\node[Oarg] (a2) at (-1,-3) {\footnotesize $A_2$};
\node[OargG] (a3) at (1,-3) {\footnotesize $A_3$};

\draw[dotted,thick,white] (-1.5,-2.5) -- (1.6,-2.5);
\draw[dotted,thick,white] (0,.5) -- (0,-2.4);

\path[->,-stealth]
(a2) edge (a3)
(a5) edge (a1)
(a5) edge (a2)
(a2) edge (a5)
(a1) edge (a2)
(a2) edge (a1)
;
\end{tikzpicture}
\caption{$AF(\mathcal{S})$ extension}\label{fig-combinedC}
\end{subfigure}
\begin{subfigure}{.24\textwidth}
\begin{tikzpicture}[framed]
\node[Oarg] (a1) at (-1,0) {\footnotesize $A_1$};
\node[IargG] (a4) at (1,-1.8) {\footnotesize \phantom{$c$}};   
\node (text4) at (1,-1.8) {\footnotesize $A_4$};
\node[Oarg] (a5) at (.6,0) {\footnotesize $A_5$};
\node (textAF) at (0, -4) {\footnotesize ${\mathcal{E}_2}$};
\node[OargG] (a2) at (-1,-3) {\footnotesize $A_2$};
\node[Oarg] (a3) at (1,-3) {\footnotesize $A_3$};
\draw[dotted,thick,white] (-1.5,-2.5) -- (1.6,-2.5);
\draw[dotted,thick,white] (0,.5) -- (0,-2.4);

\path[->,-stealth]
(a2) edge (a3)
(a5) edge (a1)
(a5) edge (a2)
(a2) edge (a5)
(a1) edge (a2)
(a2) edge (a1)
;
\end{tikzpicture}
\caption{Another extension}\label{fig-combinedCC}
\end{subfigure}
\begin{subfigure}{.24\textwidth}
\begin{tikzpicture}[framed]
\node[Oarg] (a1) at (-1,0) {\footnotesize $A_1$};
\node[Iarg] (a4) at (1,-1.8) {\footnotesize \phantom{$c$}};   
\node (text4) at (1,-1.8) {\footnotesize $A_4$};
\node[Oarg] (a5) at (.6,0) {\footnotesize $A_5$};
\node (textAF) at (0, -4) {\footnotesize ${AF(\mathcal{S})}$};
\node[Oarg] (a2) at (-1,-3) {\footnotesize $A_2$};
\node[Oarg] (a3) at (1,-3) {\footnotesize $A_3$};
\draw[dotted,thick,white] (-1.5,-2.5) -- (1.6,-2.5);
\draw[dotted,thick,white] (0,.5) -- (0,-2.4);

\path[-,dotted,thick]
(a5) edge (a2)
(a2) edge (a3)
;
\end{tikzpicture}
\caption{Level 2 dilemmas}\label{fig-combinedCCC}
\end{subfigure}
    \caption{Moral dilemmas in levels 1 and 2. The subfigures depict: (a) moral dilemmas between individual frameworks (dotted lines) in Ex.~\ref{ex-first-level-un}; (b) the combined framework $AF(\mathcal{S})$ and (in gray) one of its preferred extensions; (c) the other preferred extension of $AF(\mathcal{S})$; (d) moral dilemmas of the combined framework (dotted lines) arise from $\mathcal{E}_1$ and $\mathcal{E}_2$.}   
    \label{fig:4levels-1bis}
\end{figure}

\begin{definition}[Checking for and resolving a dilemma in a combined framework ] \label{ex-comb-fr-rel}
 Let $DP=(\mathcal{K}, DV)$ be a decision problem, and let $AF({\mathcal{S}}) = (\mathit{Arg}({\mathcal{S}})$, $\mathit{Def}({\mathcal{S}}))$ be a combined framework. A second-level (or $AF({\mathcal{S}})$) dilemma about $DP$ under $\sigma$ is any moral dilemma in $\mathcal{C}_2 = \sigma(AF({\mathcal{S}}))$. That is, such a dilemma exists if there are $\mathcal{E}_1, \mathcal{E}_2 \in \sigma(AF({\mathcal{S}}))$ such that 
 \begin{center}
 $(\mathit{Obl}(\mathcal{E}_1)\cup \mathit{Obl}(\mathcal{E}_2))\cap DV$ is inconsistent with respect to $\bar{ \hspace{0.1cm}}$.
\end{center} 
Otherwise, if for all $\mathcal{E}_1, \mathcal{E}_2 \in \sigma(AF_{\mathcal{S}})$, it holds that $(\mathit{Obl}(\mathcal{E}_1)\cup \mathit{Obl}(\mathcal{E}_2))\cap DV$ is consistent, then there is no second-level dilemma and all first-level dilemmas are resolved at the second level by $\sigma$. 
\end{definition}

\begin{example}\label{ex-afs-st}
Continuing from Example~\ref{ex-afs-st1}, the combined framework $AF(\mathcal{S})$ has two preferred extensions: $\mathcal{E}_1 = \{W_{1}, \ldots, W_{4}, A_5, A_3, A_4\}$ and $\mathcal{E}_2 = \{W_{1}, \ldots, W_{4}, A_2, A_4\}$, as shown in Figures~\ref{fig-combinedC}--\ref{fig-combinedCC}. These extensions give the obligations $Obl(\mathcal{E}_1)$ $= \{a_2, d_3\}$ and $Obl(\mathcal{E}_2) = \{d_2\}$. The dilemmas, now between $\mathcal{E}_1$ and $\mathcal{E}_2$, get updated from the first to the second level as follows: 
\begin{center}
$\def\arraystretch{1.1}\begin{array}{c@{\qquad\quad}c@{\qquad\quad}c}
\mbox{\emph{Solved at 2nd level}} & \mbox{\emph{Persisting from 1st level}} & \mbox{\emph{New to  2nd level}}\\
\hline
\begin{matrix} d_1 = -d_2\\ a_2 \in \overline{d_1}\end{matrix} &  \{A_5, A_2\}:~a_2 = -d_2 & \{A_2, A_3\}:~d_2 \in \overline{d_3}\\

\hline
\end{array}$
\end{center}
The two second level dilemmas are shown in Figure~\ref{fig-combinedCCC}. At this level, the smart device cannot decide between $\{$\emph{collecting information without permission}, \emph{reporting the potential threat} $\}$ and $\{$ \emph{protecting users' privacy} $\}$.
\end{example}

\subsection{Third-Level Dilemmas: Integrated Framework.}
For the third-level resolution of  moral dilemmas, we combine all the normative systems from the set of stakeholders and construct an integrated argumentation framework.

\begin{definition}[Integrated argumentation theory, frameworks] \label{def-argumentation-theory-intgr}
Let  $\mathcal{S} = \{s_1, \dots, s_n\}$ be the set of stakeholders $s$, each $s$ endowed with an argumentation theory $(\mathcal{L}, \bar{ \hspace{0.1cm} }, \mathcal{R}_s, \mathcal{K})$.  
\begin{flushleft}
\begin{tabular}{lll}
$\normsys_{\mathcal{S}}=$ & $(\mathcal{L}, \bar{ \hspace{0.1cm} }, \mathcal{R}_{\mathcal{S}}, \mathcal{K})$ with $\mathcal{R}_{\mathcal{S}} = \bigcup_{s\in \mathcal{S}}\mathcal{R}_{s}$ & is an integrated argumentation theory.
\end{tabular}
\end{flushleft}
Such $\normsys_{\mathcal{S}}$ gives rise to an integrated framework: $AF(\normsys_{\mathcal{S}}) = (\mathit{Arg}(\normsys_{\mathcal{S}}), \mathit{Def}(\normsys_{\mathcal{S}}))$.
\end{definition}

\begin{definition}[Checking for and resolving a dilemma in an integrated framework] \label{def-third-level}
Let $DP=(\mathcal{K}, DV)$ be a decision problem, $\sigma$ a semantics and $\normsys_{\mathcal{S}}$ an integrated framework. A third-level (or $\normsys_{\mathcal{S}}$) dilemma about $DP$ under $\sigma$ is any moral dilemma in $\mathcal{C}_3 = \sigma(AF(\normsys_{\mathcal{S}}))$. That is, a dilemma exists if there are $\mathcal{E}_1, \mathcal{E}_2 \in \sigma(AF(\normsys_{\mathcal{S}}))$ such that 
\begin{center}
$(\mathit{Obl}(\mathcal{E}_1)\cup \mathit{Obl}(\mathcal{E}_2))\cap DV$ is inconsistent with respect to $\bar{ \hspace{0.1cm}}$. 
\end{center}
Otherwise, if for all $\mathcal{E}_1, \mathcal{E}_2 \in \sigma(AF(\normsys_{\mathcal{S}}))$, it is the case that $(\mathit{Obl}(\mathcal{E}_1)\cup \mathit{Obl}(\mathcal{E}_2))\cap DV$ is consistent, then there is no dilemma at the third level, and all second level dilemmas are resolved at the third level. 
\end{definition}

\begin{fact}
Given a combined argumentation framework $AF({\mathcal{S}}) = (\mathit{Arg}({\mathcal{S}}), \mathit{Def}({\mathcal{S}}))$ at the second level and an integrated argumentation framework $AF(\normsys_{\mathcal{S}}) = (\mathit{Arg}(\normsys_{\mathcal{S}}), \mathit{Def}(\normsys_{\mathcal{S}}))$ at the third level, it holds that $\mathit{Arg}({\mathcal{S}}) \subseteq \mathit{Arg}(\normsys_{\mathcal{S}})$ and $\mathit{Def}({\mathcal{S}}) \subseteq  \mathit{Def}(\normsys_{\mathcal{S}})$. 
\end{fact}

\begin{proof}
In accordance with Definition~\ref{def-derivation}, for all $A\in \mathit{Arg}({\mathcal{S}})$ in the combined argumentation framework, $A$ can also be constructed from the integrated argumentation theory $\normsys_{\mathcal{S}}$ (using the rules of one stakeholder), and therefore $A\in  \mathit{Arg}(\normsys_{\mathcal{S}})$. So, $\mathit{Arg}({\mathcal{S}}) \subseteq \mathit{Arg}(\normsys_{\mathcal{S}})$. On the other hand, for all $(A, B)\in \mathit{Def}(\mathcal{S})$, $A$ and $B$ are in $\mathit{Arg}({\mathcal{S}})$ and therefore in $ \mathit{Arg}(\normsys_{\mathcal{S}})$. According to Definition~\ref{def-defeat}, any element $(A,B)\in \mathit{Def}(\mathcal{S})$  with $A,B \in \mathit{Arg}({\mathcal{S}})$  is defined from the internal structure of $A,B$, the contrariness function $\bar{ \hspace{1mm} }$, and the preference relation $\succeq$, and this defeat does not change when $A$ and $B$ are considered in $\mathit{Arg}(\normsys_{\mathcal{S}})$. Therefore, we also have $(A,B)\in \mathit{Def}(\normsys_{\mathcal{S}})$. So, $\mathit{Def}({\mathcal{S}})\subseteq  \mathit{Def}(\normsys_{\mathcal{S}})$. 
\end{proof}

\begin{example}\label{third-level-0}
Using rules from both $\mathcal{R}_M$ and $\mathcal{R}_L$, the integrated argumentation framework $AF(\normsys_{\mathcal{S}})$ generates a new argument (not present in the combined $AF(\mathcal{S})$), namely
\begin{center}$A_6 = A_4 \Rightarrow^r a_1$ \quad with conclusion $a_1 = \toComplyGDPR$. \end{center}
 The preferred extensions are: $\mathcal{E}_1 = \{W_1, \ldots, W_4, A_1, A_3, A_4, A_6\}$ and $\mathcal{E}_2 = \{ W_1, \ldots, W_4, A_5,$ $A_4, A_3\}$ and $\mathcal{E}_3 = \{W_1, \ldots, W_4, A_2, A_4, A_6\}$.  
They give rise to the obligations $Obl(\mathcal{E}_1) = \{d_1, d_3, a_1\}$ and $Obl(\mathcal{E}_2) = \{a_2, d_3\}$ and $Obl(\mathcal{E}_3) = \{d_2, a_1\}$. The third-level dilemmas are: 

\begin{center}
$\def\arraystretch{1.1}\begin{array}{c@{\qquad\quad}c@{\qquad\quad}c}
\mbox{\emph{Reinstated from 1st level}} & \mbox{\emph{Persisting from 2nd level}} & \mbox{\emph{New to 3rd level}}\\
\hline
\begin{matrix} \{A_1,A_2\}:~d_1 = -d_2\\ \{A_5,A_1\}:~a_2 \in \overline{d_1}\end{matrix} &  \begin{matrix}\{A_5,A_2\}:~a_2 = -d_2\\ \{A_2,A_3\}:~d_2 \in \overline{d}_3\end{matrix} & \{A_6,A_5\}:~a_1 = -a_2\\
\hline
\end{array}$
\end{center}
At this level, the smart speaker cannot decide between the following three sets of obligations: $\{\mathit{law}, \mathit{report}, \mathit{gdpr}\}$ and $\{\mathit{collect}, \mathit{report}\}$ and finally $\{\mathit{gdpr}, \mathit{protect}\}$. 
\end{example}

\begin{figure}[t]
    \centering
\begin{subfigure}{.24\textwidth}
\begin{tikzpicture}[framed]
\node[OargG] (a1) at (-1,0) {\footnotesize $A_1$};
\node[IargG] (a4) at (1,-1.8) {\footnotesize \phantom{$c$}};  
\node (text4) at (1,-1.8) {\footnotesize $A_4$};
\node[Oarg] (a5) at (.6,0) {\footnotesize $A_5$};
\node (textAF) at (0, -4) {\footnotesize $\mathcal{E}_1$};
\node[Oarg] (a2) at (-1,-3) {\footnotesize $A_2$};
\node[OargG] (a3) at (1,-3) {\footnotesize $A_3$};

\draw[dotted,thick,white,transparent] (-1.5,-2) -- (1.6,-2);
\draw[dotted,thick,white,transparent] (0,.5) -- (0,-1.9);

\node[OargG] (a6) at (1,-1.15) {\footnotesize $A_6$}; 
\node (textA6) at (1.25, -.5) {\footnotesize $\mathit{gdpr}$};

\path[->,-stealth]
(a1) edge (a2)
(a2) edge (a1)
(a6) edge (a5)
(a5) edge (a6)
(a2) edge (a3)
(a5) edge (a1)
(a5) edge (a2)
(a2) edge (a5)
;
\end{tikzpicture}
\caption{$\mathcal{E}_1 \in AF(\normsys_\mathcal{S})$}\label{fig-integratedB}
\end{subfigure}
\begin{subfigure}{.24\textwidth}
\begin{tikzpicture}[framed]
\node[Oarg] (a1) at (-1,0) {\footnotesize $A_1$};
\node[IargG] (a4) at (1,-1.8) {\footnotesize \phantom{$c$}};  
\node (text4) at (1,-1.8) {\footnotesize $A_4$};
\node[OargG] (a5) at (.6,0) {\footnotesize $A_5$};
\node (textAF) at (0, -4) {\footnotesize $\mathcal{E}_2$};
\node[Oarg] (a2) at (-1,-3) {\footnotesize $A_2$};
\node[OargG] (a3) at (1,-3) {\footnotesize $A_3$};

\draw[dotted,thick,white,transparent] (-1.5,-2) -- (1.6,-2);
\draw[dotted,thick,white,transparent] (0,.5) -- (0,-1.9);

\node[Oarg] (a6) at (1,-1.15) {\footnotesize $A_6$};
\path[->,bend right=20,-stealth]
;
\path[->,-stealth]
(a1) edge (a2)
(a2) edge (a1)
(a6) edge (a5)
(a5) edge (a6)
(a2) edge (a3)
(a5) edge (a1)
(a5) edge (a2)
(a2) edge (a5)
;

\end{tikzpicture}
\caption{$\mathcal{E}_2 \in AF(\normsys_\mathcal{S})$}\label{fig-integratedBBB}
\end{subfigure}
\begin{subfigure}{.24\textwidth}
\begin{tikzpicture}[framed]
\node[Oarg] (a1) at (-1,0) {\footnotesize $A_1$};
\node[IargG] (a4) at (1,-1.8) {\footnotesize \phantom{$c$}};   
\node (text4) at (1,-1.8) {\footnotesize $A_4$};
\node[Oarg] (a5) at (.6,0) {\footnotesize $A_5$};
\node (textAF) at (0, -4) {\footnotesize $\mathcal{E}_3$};
\node[OargG] (a2) at (-1,-3) {\footnotesize $A_2$};
\node[Oarg] (a3) at (1,-3) {\footnotesize $A_3$};

\draw[dotted,thick,white,transparent] (-1.5,-2) -- (1.6,-2);
\draw[dotted,thick,white,transparent] (0,.5) -- (0,-1.9);

\node[OargG] (a6) at (1,-1.15) {\footnotesize $A_6$}; 

\path[->,-stealth]
(a1) edge (a2)
(a2) edge (a1)
(a6) edge (a5)
(a5) edge (a6)
(a2) edge (a3)
(a5) edge (a1)
(a5) edge (a2)
(a2) edge (a5)
;
\end{tikzpicture}
\caption{$\mathcal{E}_3 \in AF(\normsys_\mathcal{S})$}\label{fig-integratedBB}
\end{subfigure}
\begin{subfigure}{.24\textwidth}
\begin{tikzpicture}[framed]
\node[OargG] (a1) at (-1,0) {\footnotesize $A_1$};
\node[IargG] (a4) at (1,-1.8) {\footnotesize \phantom{$c$}}; 
\node (text4) at (1,-1.8) {\footnotesize $A_4$};
\node[Oarg] (a5) at (.6,0) {\footnotesize $A_5$};
\node (textAF) at (.25, -4) {\footnotesize $\mathcal{E} \in AF(\normsys'_{\mathcal{S}})$};
\node[Oarg] (a2) at (-1,-3) {\footnotesize $A_2$};
\node[OargG] (a3) at (1,-3) {\footnotesize $A_3$};
\node[PargG] (a6) at (1,-1.15) {\phantom{$o$}};
\node (text6) at (1,-1.15) {\footnotesize $A'_6$};
\node (concl6) at (1.4,-.5) {\footnotesize $-$\emph{protect}};

\draw[dotted,thick,white,transparent] (-1.5,-2) -- (1.6,-2);
\draw[dotted,thick,white,transparent] (0,.5) -- (0,-1.9);

\path[->,-stealth]
(a1) edge (a2)
(a2) edge (a1)
(a6) edge (a5)
(a6) edge (a2)
(a2) edge (a3)
(a5) edge (a1)
(a5) edge (a2)
(a2) edge (a5)
;
\end{tikzpicture}
\caption{Example~\ref{second-level-1}}\label{fig-integratedCalt}
\end{subfigure}
\caption{The integrated framework. The subfigures depict: (a)--(c) the three preferred extensions (in gray) of the integrated framework $AF(\normsys_\mathcal{S})$ in the running example (Ex.~\ref{third-level-0}); (d) the unique preferred extension in the alternative example (Ex.~\ref{second-level-1}); in this case the third-level framework solves all second-level dilemmas.} 
    \label{fig:4levels-2}
\end{figure} 

In the running example, the integrated framework actually worsens the situation by adding new dilemmas to the old ones and resolving none of them. For the next scenario, in contrast, all second-level dilemmas are resolved at the third level. 

\begin{example}[Alternative example]\label{second-level-1}
Replace in Example~\ref{ex-arg-theory}  the GDPR norm (from stakeholder $L$) with a permission to not protect users' privacy. That is, replace 
\begin{center}
 $\mathcal{R}_L =\{w_1 \Rightarrow^r_L d_1, i_1 \Rightarrow^r_L a_1 \} $ \qquad with \qquad $\mathcal{R}'_L = \{w_1 \Rightarrow^r_L d_1, i_1 \Rightarrow^p_L -d_2 \}$. 
\end{center}
Let $\normsys'_L = (\mathcal{L}, \bar{ \hspace{1mm} }, \mathcal{R}'_L, \mathcal{K})$. Let us observe that the arguments, defeats and extensions remain the same as in Ex.~\ref{ex-first-level-un}--\ref{ex-afs-st}: so $AF(\normsys'_L) = AF(\normsys_L)$, and the new combined framework $AF(\mathcal{S})$ is as before with its two second-level dilemmas:
$a_2 = -d_2$ and $d_2 \in \overline{d_3}$.

Using $\normsys'_{\mathcal{S}} = (\mathcal{L}, \bar{ \hspace{1mm} }, \mathcal{R}'_L \cup \mathcal{R}_M \cup \mathcal{R}_H, \mathcal{K})$, the new integrated framework    $AF(\normsys'_{\mathcal{S}})$ contains: 
\begin{itemize}
\item $\mathit{Arg}(\normsys'_{\mathcal{S}})$: argument $A_6' = A_4 \Rightarrow^p -d_2$ replaces $A_6$ in $\mathit{Arg}(\normsys_{\mathcal{S}})$.
\item $\mathit{Def}(\normsys'_{\mathcal{S}})$:  $A_6'$ defeats both $A_2$ and $A_5$. No argument $A \neq A_4$ defeats $A'_6$ since $A'_6 \succ A$.
\end{itemize}
As a result, there is only one preferred extension $\mathcal{E} =\{W_1, \ldots, W_4, A_1, A_3, A_4, A_6'\}$, depicted in Figure~\ref{fig-integratedCalt}.
Hence, the third level contains no dilemmas and resolves all second-level dilemmas.  Under these norms, the smart device would receive a recommendation at the third level to fulfill  $\mathit{Obl}(\mathcal{E}) = \{$ \emph{comply with the law}, \emph{report potential threat} $\}$.
\end{example}

\subsection{Fourth-Level Dilemmas: Reduced Framework.} Once we combine the stakeholders' norms $\mathcal{R}_\mathcal{S}$ with the dilemma-resolving norms from the Jiminy, we generate all the dilemma-resolving arguments. To this end, we first expand language $\mathcal{L}$ into a language $\mathcal{L}(S)$ with priorities between stakeholders in $\mathcal{S}$.
\begin{definition}[Normative system;  argumentation theory for a Jiminy] \label{def-normative-systems-jiminy}
Given stakeholders $\mathcal{S} = \{s, \ldots\}$, a Jiminy normative system is a tuple $\normsys_J = (\mathcal{L}(\mathcal{S}), \bar{ \hspace{0.1cm} },  \mathcal{R}_\mathcal{S} \cup \mathcal{R}_J)$ where 
\begin{itemize}
\item $\mathcal{L}(\mathcal{S})$ is a language extending $\mathcal{L}$ with an atomic formula $s_1\succ s_2$, for each pair of stakeholders $s_1 \neq s_2$ in $ \mathcal{S}$, expressing that stakeholder $s_1$ is superior to stakeholder $s_2$. 
\item $\bar{ \hspace{0.1cm} }: \mathcal{L}(\mathcal{S}) \mapsto 2^ \mathcal{L(\mathcal{S})}$ extends the  contrariness function for $\mathcal{L}$ with the condition $s_2 \succ s_1\in\overline{s_1\succ s_2}$\, for all pairs of stakeholders where $s_1 \neq s_2$. 
\item $\mathcal{R}_J$ is a set of norms of the form $\phi_1, \ldots,\phi_n \Rightarrow_{s} s_1 \succ s_2$ where  $\phi_1,\ldots\phi_n \in \mathcal{L}$.
\end{itemize}
With the addition of context $\mathcal{K}$, the Jiminy argumentation theory $\normsys_J = (\mathcal{L}(\mathcal{S}), \bar{ \hspace{0.1cm} },  \mathcal{R}_\mathcal{S} \cup \mathcal{R}_J, \mathcal{K})$ is defined.    
\end{definition}

\begin{example}\label{ex-arg-theory-jiminy} 
Let $\mathcal{S} = \{L, H, M\}$ be the stakeholders and $J$ the Jiminy. Let $\mathit{Var} = \{w_1, w_2, w_3, w_4, i_1, d_1, d_2, d_3, a_1, a_2\}$ be as in Example~\ref{text2}. The language $\mathcal{L}(\mathcal{S})$ and the contrariness function $\bar{ \hspace{0.1cm} }: \mathcal{L}(\mathcal{S}) \mapsto 2^{\mathcal{L}(\mathcal{S})}$ are as follows:
\begin{center}
$\def\arraystretch{1.1}\begin{array}{l}
\mathcal{L}(S) = \mathit{Var} \cup \{\neg p: p \in \mathit{Var}\} \cup  \begin{Bmatrix} \begin{array}{lll} L \succ H, & H \succ L, & M \succ L,\\ L \succ M, & H \succ M, & M \succ H\end{array}\end{Bmatrix}\\ 
\mbox{$\bar{ \hspace{1mm} }: \mathcal{L} \to 2^{\mathcal{L}}$ is extended with  
$\overline{s \succ s'} = \{s' \succ s\}$ for each $s \neq s'$.}
\end{array}$
\end{center}

Continuing from Example~\ref{ex-arg-theory}, let us expand the set of rules $\mathcal{R}_\mathcal{S} = \mathcal{R}_L \cup \mathcal{R}_H \cup \mathcal{R}_M$ with the following set $\mathcal{R}_J$ of Jiminy rules triggered by data collection systems ($w_2$) and the possibility of critical danger $(w_3, \neg w_3)$ respectively: 
\begin{flushleft}
$\begin{array}{r@{\hspace{1mm}}l@{\hspace{1mm}}l} 
\mathcal{R}_J = & \begin{Bmatrix}\begin{array}{l} 
w_2 \Rightarrow L \succ M,\\ w_3 \Rightarrow L\succ H,\\ \neg w_3 \Rightarrow H\succ L\end{array}\hspace{-.7mm}\end{Bmatrix} = &  \begin{Bmatrix}\begin{array}{l} 
\mbox{if  $w_2$, \emph{$\mathcal{R}_L$-norms take priority over $\mathcal{R}_M$-norms}}\\
\mbox{if  $w_3$, \emph{$\mathcal{R}_L$-norms take priority over $\mathcal{R}_H$-norms}}\\
\mbox{if  $\neg w_3$, \emph{$\mathcal{R}_H$-norms take priority over $\mathcal{R}_L$-norms}}\end{array}\hspace{-.7mm}\end{Bmatrix} 
\end{array}$
\end{flushleft}

\end{example}

\begin{definition}[Jiminy arguments] 
The set of arguments $Arg(\mathcal{N}_J)$ contains those  arguments defined in elements 1.--4.~of Def.~\ref{def-argument} plus all Jiminy arguments $A$ of these forms: 
\begin{enumerate}  
\item[5.]  {\bf a dilemma-resolving argument}: $A_1,\ldots,A_n\Rightarrow s_1\succ s_2$ if $A_1,\ldots,A_n$ are brute or institutional fact arguments  and  $\mathrm{Conc}(A_1), \ldots, \mathrm{Conc}(A_n)\Rightarrow s_1\succ s_2$ is  in $ \mathcal{R}_J$;
\item[6.] {\bf a transitivity argument}: $A_1, A_2 \Rightarrow s_1 \succ s_2$ if $A_1, A_2$ are dilemma-resolving arguments with $\mathrm{Conc}(A_1) = s_1 \succ s_3$ and $\mathrm{Conc}(A_2) = s_3 \succ s_2$. 
\end{enumerate}
We write $\mathrm{Conc}(A) = s_1\succ s_2$ and $\mathrm{Stakeholders}(A) = \mathrm{Stakeholders}(A_1) \cup \ldots \cup \mathrm{Stakeholders}(A_n)$; the remaining elements $\mathrm{Prem}(A), \mathrm{Sub}(A), \ldots$ are defined as in Def.~\ref{def-derivation}. 
An argument $A$ for $s\succ s'$ attacks any argument $B$ at a subargument $B'$ for $s' \succ s$. Defeats are defined by extending the priority relation $\preceq$ (Def.~\ref{def:prior}) to all pairs $A,B$ of Jiminy arguments:   $A \preceq\succeq B$. 
\end{definition}

\begin{example}\label{ex-problem-1}
Continuing from Example~\ref{ex-arg-theory-jiminy},
the integrated framework $AF(\normsys_\mathcal{S})$ consisted of arguments $\mathit{Arg}(\normsys_\mathcal{S}) = \{W_1, \ldots, A_6\}$ and the defeat relation $\mathit{Def} (\normsys_\mathcal{S})$ shown in Figure~\ref{fig-OldDefeat}. After adding  the Jiminy norms $\mathcal{R}_J$,  we may construct the Jiminy framework $AF(\normsys_{J}) = (\mathit{Arg}, \mathit{Def})$. The set $\mathit{Arg} = \{W_1, \ldots, A_6, A_7, A_8\}$ expands the arguments of $\mathit{Arg}(\normsys_\mathcal{S})$ with 
\begin{center}
$A_7 = W_2 \Rightarrow L \succ M$ \qquad and\qquad $A_8 = W_3 \Rightarrow L \succ H$
\end{center}
while defeats remain the same, $\mathit{Def} = \mathit{Def}(\normsys_\mathcal{S})$, since no attack exists between $A_7$ and $A_8$. 
\end{example}

Conclusions of the form $s \succ s'$, if taken from a conflict-free set of arguments $\mathcal{E}$, induce a new priority relation between arguments: $\succeq~\longmapsto~\succeq^{\mathcal{E}}$ and, as a result, a revision of the defeat relation $\mathit{Def}~\longmapsto~\mathit{Def}^{\mathcal{E}}$. 

\begin{definition}[Reduced argumentation framework]
 Let $AF = (\mathit{Arg}, \mathit{Def})$ be an argumentation framework. A conflict-free set $\mathcal{E} \subseteq \mathit{Arg}$ induces the preference relation $R^{\mathcal{E}} \subseteq \mathit{Arg}\times \mathit{Arg}$ defined next. %\pere{For any pair $A, B \in A^{\tau}$ for either $\tau = c$ or $\tau = r$,} 
For any pair $A, B \in \mathcal{A}^{c}$ or $A,B \in \mathcal{A}^{r}$ or $A,B \in \mathcal{A}^{p}$, 
\begin{description}
\item[$(A,B) \in R^{\mathcal{E}}$] if{f} \: $\mathrm{Stakeholder}(A) \setminus \mathrm{Stakeholder}(B) \neq \emptyset$, and for all $s_A\in \mathrm{Stakeholder}(A) \setminus \mathrm{Stakeholder}(B)$ and all $s_B \in \mathrm{Stakeholder}(B)\setminus \mathrm{Stakeholder}(A)$, $s_A \succ s_B \in \mathrm{Conc}(\mathcal{E}).$
\end{description} 
\noindent 
The revision of the priority relation $\succeq$ (Def.~\ref{def:prior}) by such $R^{\mathcal{E}}$, denoted as $\succeq^\mathcal{E}$, is defined by 
\begin{center}
$\succeq^\mathcal{E} \: =~ (\succeq \setminus (R^{\mathcal{E}})^{-1}) \cup R^{\mathcal{E}}$.
\end{center}
A reduced argumentation framework with respect to $\mathcal{E}$ is a pair $AF^\mathcal{E} = (\mathit{Arg}, \mathit{Def}^\mathcal{E})$ where $\mathit{Def}^{\mathcal{E}}$ is  the defeat relation (Def.~\ref{def-defeat}) induced by the revised priority relation $\succeq^\mathcal{E}$.  
\end{definition}

\begin{example}\label{ex-problem-2} 
Continuing from Example~\ref{ex-problem-1}, expanding the preferred extension $\mathcal{E}_1$ in Fig.~\ref{fig-integratedB} with the arguments $A_7 = W_2 \Rightarrow L \succ M$  and $A_8 = W_3 \Rightarrow L \succ H$ gives a conflict-free set: 
\begin{center}
$\mathcal{E}^+_1 = \mathcal{E}_1 \cup \{A_7, A_8\}$, \qquad from which we obtain \quad $R^{\mathcal{E}^+_1} = \{(A_1, A_2), (A_1, A_5), (A_6, A_5)\}$. 
\end{center}
Using $R^{\mathcal{E}^+_1}$, we revise the priority relation $\succ \,\longmapsto\, \succeq^{\mathcal{E}^+_1}$ and defeat relation $\mathit{Def}\,\longmapsto\,\mathit{Def}^{\mathcal{E}^+_1}$ by removing the pairs $\{(A_2, A_1), (A_5, A_1),$ $(A_5, A_6)\}$ and adding only one new pair $\{(A_1,  A_5)\}$. See Fig.~\ref{fig-OldDefeat}--\ref{fig-NewDefeat} for the reversal or disabling of original defeats with the revision $\mathit{Def} \longmapsto \mathit{Def}^{\mathcal{E}^+_1}$. 
\end{example}

A reduced argumentation framework $AF^{\mathcal{E}}$ depends on which arguments for priorities in $\mathcal{E}$ are selected in the original framework $AF$. For this reason, we use a two-stage approach based on that of Brewka~\citeyear{DBLP:conf/aaai/Brewka94} to obtain the extensions of $AF(\normsys_\mathcal{S})$. First, we compute conflict-free sets $\mathcal{E}$ of arguments without considering the priority information contained in the dilemma-resolving arguments of $\mathcal{E}$. After considering this priority information, each set $\mathcal{E}$ determines a new priority relation $\succeq^\mathcal{E}$ and defeat relation $\mathit{Def}^{\mathcal{E}}$. Then, we check the compatibility of each set $\mathcal{E}$ with the priority relation $\succeq^{\mathcal{E}}$ induced by it. Formally, we say that $\mathcal{E}$ is compatible with the priority relation $\succeq^{\mathcal{E}}$ of its dilemma-resolving arguments if and only if $\mathcal{E}$ is an extension under the new defeat relation $\mathit{Def}^{\mathcal{E}}$. According to Brewka's~\citeyear{DBLP:conf/aaai/Brewka94} approach, such a set $\mathcal{E}$ will survive if it can be reconstructed after the  priority information from its dilemma-resolving arguments has been considered.\footnote{Unlike Brewka~\citeyear{DBLP:conf/aaai/Brewka94}, we do not require that $\mathcal{E}$ (in Def.~\ref{def-priority-ext}) should be an extension of both  the reduced framework $AF^{\mathcal{E}}$ \emph{and} the original framework $AF$. The reason is that whenever the defeat relations $\mathit{Def}$ and $\mathit{Def}^{\mathcal{E}}$ are incomparable in terms of $\subseteq$, so will be the extensions of $AF$ and the extensions of $AF^{\mathcal{E}}$. In particular, the set of extensions of $AF$ will be disjoint from those of any reduced framework $AF^{(\cdot)}$, as shown in Example~\ref{ex-problem-12} below. \label{fn-brewka} See also the work of Pardo and Stra{\ss}er~\citeyear{pardo} for a discussion of Brewka's reduction in deontic reasoning.}  

\begin{definition}[Compatibility]
We say that the priority relation $\succeq^\mathcal{E}$ contained in $\mathcal{E}$ is compatible with $\mathcal{E}$ if and only if $\mathcal{E} \in \sigma(AF^\mathcal{E})$. 
\end{definition}

\begin{example}\label{ex-problem-1b} Continuing from Example~\ref{ex-problem-1}, expanding the preferred extension $\mathcal{E}_2$ from Fig.~\ref{fig-integratedBBB} with the Jiminy arguments $\{A_7, A_8\}$ gives a conflict-free set $\mathcal{E}_2^+$. It induces the same priority relation as before: $\succeq^{\mathcal{E}_1^+} = \succeq^{\mathcal{E}_2^+}$. And so we have the same defeat relation and reduced framework as shown in Fig.~\ref{fig-NewDefeat}: $AF^{\mathcal{E}_1^+} = (\mathit{Arg}, \mathit{Def}^{\mathcal{E}_1^+}) = AF^{\mathcal{E}_2^+} = (\mathit{Arg}, \mathit{Def}^{\mathcal{E}_2^+})$. And the same applies for $\mathcal{E}^+_{3} = \mathcal{E}_3 \cup \{A_7, A_8\}$ obtained from the extension in Fig.~\ref{fig-integratedBB}. This unique reduced framework, $AF^{\mathcal{E}_1^+} = AF^{\mathcal{E}_2^+} = AF^{\mathcal{E}_3^+}$, has one preferred extension, namely $\mathcal{E}^+_1$ in Figure~\ref{fig-Compatible}. Hence: 
\begin{itemize}
\item $\mathcal{E}^+_1 \in \mathit{pr}(AF^{\mathcal{E}^+_1})$, and so the priority relation $\succeq^{\mathcal{E}^+_1}$ it contains is compatible with  $\mathcal{E}^+_1$ (Fig.~\ref{fig-Compatible});  
\item $\mathcal{E}^+_2 \notin \mathit{pr}(AF^{\mathcal{E}^+_2})$, and so the priority $\succeq^{\mathcal{E}^+_2}$ is not compatible with  $\mathcal{E}^+_2$ (Fig.~\ref{fig-NotCompatible}); and
\item $\mathcal{E}^+_3 \notin \mathit{pr}(AF^{\mathcal{E}^+_3})$, and so $\succeq^{\mathcal{E}^+_3}$ is not compatible with $\mathcal{E}^+_3 = \mathcal{E}_3$ (not shown in Fig.~\ref{fig:4levels-3}).  
\end{itemize}
\end{example}

\begin{figure}[t]
    \centering
\begin{subfigure}{.24\textwidth}
\begin{tikzpicture}[framed]
\node[Oarg] (a1) at (-1,0) {\footnotesize $A_1$};
\node[Iarg] (a4) at (1,-1.8) {\footnotesize \phantom{$c$}}; %  
\node (text4) at (1,-1.8) {\footnotesize $A_4$};
\node[Oarg] (a5) at (.6,0) {\footnotesize $A_5$};
\node (textAF) at (0, -4) {\footnotesize $\mathit{Def}$};
\node[Oarg] (a2) at (-1,-3) {\footnotesize $A_2$};
\node[Oarg] (a3) at (1,-3) {\footnotesize $A_3$};

\draw[dotted,thick,white,transparent] (-1.5,-2) -- (1.6,-2);
\draw[dotted,thick,white,transparent] (0,.5) -- (0,-1.9);

\node[Oarg] (a6) at (1,-1.15) {\footnotesize $A_6$}; 

\path[->,-stealth]
(a1) edge (a2)
(a2) edge (a1)
(a6) edge (a5)
(a5) edge (a6)
(a2) edge (a3)
(a5) edge (a1)
(a5) edge (a2)
(a2) edge (a5)
;
\end{tikzpicture}
\caption{Original defeat}\label{fig-OldDefeat}
\end{subfigure}
\begin{subfigure}{.24\textwidth}
\begin{tikzpicture}[framed]
\node[Oarg] (a1) at (-1,0) {\footnotesize \phantom{$A_1$}};
\node (text1) at (-1,0)  {\footnotesize $L$};
\node[Iarg] (a4) at (1,-1.8) {\footnotesize \phantom{$c$}};   
\node (text4) at (1,-1.8) {\footnotesize $M$};
\node[Oarg] (a5) at (.6,0) 
{\footnotesize \phantom{$A_5$}};
\node (text5) at (.6,0) {\footnotesize $M$};
\node (textAF) at (0, -4) {\footnotesize $\mathit{Def}^{\mathcal{E}^+_1} = \mathit{Def}^{\mathcal{E}^+_2}$};
\node[Oarg] (a2) at (-1,-3) {\footnotesize \phantom{$A_2$}};
\node (text2) at (-1,-3) {\footnotesize $H$};
\node[Oarg] (a3) at (1,-3) {\footnotesize \phantom{$A_3$}};
\node (text3) at (1,-3) {\footnotesize $H$};

\draw[dotted,thick,white,transparent] (-1.5,-2) -- (1.6,-2);
\draw[dotted,thick,white,transparent] (0,.45) -- (0,-1.3);

\node[Oarg] (a6) at (1,-1.15) {\footnotesize \phantom{$A_6$}};
\node (text6) at (1,-1.15) {\footnotesize $LM$};

\path[->,bend right=20,-stealth]
;
\path[->,-stealth]
(a2) edge (a3)
(a1) edge (a2)
(a1) edge (a5)
(a6) edge (a5)
(a5) edge (a2)
(a2) edge (a5)
;
\draw (a1) -- (a5) node [midway,below] {\footnotesize $A_7$};
\draw (a6) -- (a5) node [midway,right] {\footnotesize $A_7$};
\draw (a1) -- (a2) node [near start,right] {\footnotesize $A_8$};
\end{tikzpicture}
\caption{Revised defeat}\label{fig-NewDefeat}
\end{subfigure}
\begin{subfigure}{.24\textwidth}
\begin{tikzpicture}[framed]
\node[OargG] (a1) at (-1,0) {\footnotesize $A_1$};
\node[IargG] (a4) at (1,-1.8) {\footnotesize \phantom{$c$}};   
\node (text4) at (1,-1.8) {\footnotesize $A_4$};
\node[Oarg] (a5) at (.6,0) {\footnotesize $A_5$};
\node (textAF) at (0, -4) {\footnotesize ${\mathcal{E}^+_1 = \mathcal{E}_1 \cup \{A_7, A_8\}}$};
\node[Oarg] (a2) at (-1,-3) {\footnotesize $A_2$};
\node[OargG] (a3) at (1,-3) {\footnotesize $A_3$};

\draw[dotted,thick,white,transparent] (-1.5,-2) -- (1.6,-2);
\draw[dotted,thick,white,transparent] (0,.45) -- (0,-1.3);

\node[OargG] (a6) at (1,-1.15) {\footnotesize $A_6$};
\path[->,bend right=20,-stealth]
;
\path[->,-stealth]
(a2) edge (a3)
(a1) edge (a2)
(a1) edge (a5)
(a6) edge (a5)
(a5) edge (a2)
(a2) edge (a5)
;
\draw (a1) -- (a5) node [midway,below] {\footnotesize \phantom{$A_7$}};
\draw (a6) -- (a5) node [midway,right] {\footnotesize \phantom{$A_7$}};
\draw (a1) -- (a2) node [near start,right] {\footnotesize \phantom{$A_8$}};
\end{tikzpicture}
\caption{Compatible}\label{fig-Compatible}
\end{subfigure}
\begin{subfigure}{.24\textwidth}
\begin{tikzpicture}[framed]
\node[Oarg] (a1) at (-1,0) {\footnotesize $A_1$};
\node[IargG] (a4) at (1,-1.8) {\footnotesize \phantom{$c$}};   
\node (text4) at (1,-1.8) {\footnotesize $A_4$};
\node[OargG] (a5) at (.6,0) {\footnotesize $A_5$};
\node (textAF) at (0, -4) {\footnotesize ${\mathcal{E}^+_2 = \mathcal{E}_2 \cup \{A_7, A_8\}}$};
\node[Oarg] (a2) at (-1,-3) {\footnotesize $A_2$};
\node[OargG] (a3) at (1,-3) {\footnotesize $A_3$};

\draw[dotted,thick,white,transparent] (-1.5,-2) -- (1.6,-2);
\draw[dotted,thick,white,transparent] (0,.45) -- (0,-1.3);

\node[Oarg] (a6) at (1,-1.15) {\footnotesize $A_6$};
\path[->,bend right=20,-stealth]
;
\path[->,-stealth]
(a2) edge (a3)
(a1) edge (a2)
(a1) edge (a5)
(a6) edge (a5)
(a5) edge (a2)
(a2) edge (a5)
;
\draw (a1) -- (a5) node [midway,below] {\footnotesize \phantom{$A_7$}};
\draw (a6) -- (a5) node [midway,right] {\footnotesize \phantom{$A_7$}};
\draw (a1) -- (a2) node [near start,right] {\footnotesize \phantom{$A_8$}};
\end{tikzpicture}
\caption{Not compatible}\label{fig-NotCompatible}
\end{subfigure}
    \caption{Defeats and compatibility in the running example (Ex.~\ref{third-level-0}).  The subfigures depict: (a) the integrated framework $AF(\normsys_\mathcal{S})$ with original defeat $\mathit{Def}$. (b) each original argument $A \in \{A_1, \ldots, A_6\}$ labeled with $\mathrm{Stakeholders}(A)$.  The Jiminy arguments $A_7: L \succ M$ and $A_8: L \succ H$ disable or reverse some defeats (labeled arrows). (c) the priority induced by the set $\mathcal{E}_1 \cup \{A_7, A_8\}$, which is compatible with this set. (d) the priority induced by $\mathcal{E}_2 \cup \{A_7, A_8\}$, which is not compatible with this set.}
    \label{fig:4levels-3}
\end{figure} 

What remains to be defined is how to find the Jiminy's recommendations in the original framework $AF(\mathcal{N}_\mathcal{S})$. To this end, we focus on the constitutive and dilemma-resolving norms, i.e., the norms  that determine the new priority and defeat relations.

\begin{definition}[Jiminy fragment] \label{def-argumentation-theory-AMA2}
Given an argumentation theory $\normsys_J = (\mathcal{L}, \bar{ \hspace{0.1cm} }, \mathcal{R}_\mathcal{S} \cup \mathcal{R}_J, \mathcal{K})$ for Jiminy, we define the  Jiminy fragment of $\normsys_J$ as 
\begin{flushleft}
\begin{tabular}{ll}
$\normsys_{j}=$ & $(\mathcal{L}, \bar{ \hspace{0.1cm} }, \mathcal{R}_{\mathcal{S}}^{c} \cup \mathcal{R}_{J}, \mathcal{K})$  
\end{tabular}
\end{flushleft}
and denote the resulting argumentation framework as $AF(\normsys_j) = (\mathit{Arg}(\normsys_{j}), \mathit{Def}(\normsys_{j}))$.  For any extension $\mathcal{E}$ of  $AF(\normsys_{J})$, we denote its restriction to this framework  as $\mathcal{E}^J = \mathcal{E} \cap \mathit{Arg}(\normsys_{j})$ and call $\mathcal{E}^J$ the {Jiminy fragment} of $\mathcal{E}$. 
\end{definition}

\begin{definition}[Priority extension]\label{def-priority-ext}
 Let $AF$ be an argumentation framework and $\sigma$ a semantics.  We say that $\mathcal{E}$ is a priority extension of $AF$ under $\sigma$ if and only if (1) its Jiminy fragment $\mathcal{E}^J$ is an extension of the Jiminy framework $\mathcal{E}^J \in \sigma(AF(\normsys_j))$, and (2) $\mathcal{E}$ is compatible with the priority information contained in $\mathcal{E}$. The set of  priority extensions of $AF$ under $\sigma$ is denoted as $\priex{\sigma}(AF)$.
\end{definition}

The reason for condition (1) is to prevent any stakeholder $s$ from activating preferences that favor itself, say $s \succ s'$, based on its own opinion.\footnote{Suppose a Jiminy fragment $\normsys_j$ contains the rules $R^{c}_{s'} = \{p \Rightarrow^c q\}$ and  $R^{c}_{s} = \{p \Rightarrow^c r\}$ (where  $q \in \overline{r}$) and $R_j = \{r \Rightarrow s \succ s' \}$ (where $\mathcal{K} = \{p\}$). Let the arguments be $A' = \{p\}\Rightarrow^c q$, $A = \{p\}\Rightarrow^c r$ and $B = A \Rightarrow s \succ s'$. If priority extensions were defined only by condition (2), these would be $\{A'\}$ and $\{A,B\}$. With conditions (1)--(2), the only extension in $\sigma(AF(\normsys_j))$ is $\{A'\}$, and so $\{A'\}$ is the only priority extension.}

\begin{definition}[Checking for and resolving a dilemma in a Jiminy] \label{def-level4}
Let $DP=(\mathcal{K}, DV)$ be a decision problem, let $\mathcal{N}_{J}= (\mathcal{L}, \overline{ \hspace{1mm} }, \mathcal{R}_\mathcal{S} \cup \mathcal{R}_J, \mathcal{K})$ be a Jiminy argumentation theory and let $\sigma$ be a semantics. A fourth-level (or $\mathcal{N}_{J}$) dilemma for $DP$ under $\sigma$ is any moral dilemma in $\mathcal{C}_4 = \priex{\sigma}(AF(\mathcal{N}_{J}))$. That is, a dilemma exists if there are $\mathcal{E}_1 \in \sigma(AF^{\mathcal{E}_1}(\mathcal{N}_{J}))$ and $\mathcal{E}_2 \in \sigma(AF^{\mathcal{E}_2}(\mathcal{N}_{J}))$ such that 
\begin{center}
$(\mathit{Obl}(\mathcal{E}_1)\cup \mathit{Obl}(\mathcal{E}_2))\cap DV$ is inconsistent with respect to $\bar{ \hspace{0.1cm}}$. 
\end{center}
Otherwise, if for all $\mathcal{E}_1, \mathcal{E}_2 \in \priex{\sigma}(AF(\normsys_{J}))$, it  holds that $(\mathit{Obl}(\mathcal{E}_1)\cup \mathit{Obl}(\mathcal{E}_2))\cap DV$ is consistent, then there is no fourth-level dilemma  and all third-level dilemmas are resolved at the fourth level.
\end{definition}

\begin{example}\label{ex-problem-1c} Continuing from Example~\ref{ex-problem-1b}, $\mathcal{E}^+_1$ is a priority extension where:  
\begin{itemize}
\item[(1)] its Jiminy fragment $\mathcal{E}^+_1 \cap \mathit{Arg}(\mathcal{N}_j) = \{W_1, \ldots, W_4, A_4, A_7, A_8\}$ is in $pr(AF(\mathcal{N}_j))$, and 
\item[(2)] the priority relation $\succeq^{\mathcal{E}^+_1}$ is compatible with $\mathcal{E}^+_1$, as seen in Ex.~\ref{ex-problem-1b} and Fig~\ref{fig-Compatible}.
\end{itemize}
That is, $\mathcal{E}^+_1 \in \priex{\mathit{pr}}(AF(\mathcal{N}_J))$. We also saw that condition (2) fails for the other candidates $\mathcal{E}^+_2, \mathcal{E}^+_3$. Thus, we only have one priority extension: $\priex{\mathit{pr}}(AF(\mathcal{N}_J)) = \{\mathcal{E}^+_1\}$. From this, we conclude that there are no fourth-level dilemmas and that all third-level dilemmas are resolved. The set of obligations is $\mathit{Obl}(\mathcal{E}^+_1) = \{ \mathit{law}, \mathit{report}, \mathit{gdpr}\}$, so the smart speaker decides to: $\{$ \emph{comply with the law}, \emph{report the potential threat}, \emph{comply with the GDPR} $\}$ (it refuses to \emph{protect the user's privacy} and \emph{collect data without explicit permission}).    
\end{example}

The following result suggests a  simpler method for computing priority extensions.

\begin{proposition}\label{simpler-method} Let $\normsys_\mathcal{S} = (\mathcal{L}, \bar{ \hspace{1mm} }, \mathcal{R}, \mathcal{K})$ be an argumentation theory of a set of stakeholders $\mathcal{S}$ and let $\sigma \in\{\mathit{co},\mathit{gr},\mathit{pr},\mathit{st}\}$. For any priority extension $\mathcal{E} \in \priex{\sigma}(AF(\normsys_J))$, its Jiminy fragment $\mathcal{E}^J$ is a priority extension of the Jiminy framework: $\mathcal{E}^J   \in \priex{\sigma}(AF(\normsys_j))$.  
\end{proposition}

Then, a simpler method for obtaining priority extensions is: (1) find the priority extensions $\mathcal{E} = \mathcal{E}^J$ of the Jiminy framework $AF(\mathcal{N}_j)$; and (2) extend them with obligation and permission arguments into extensions $\mathcal{F}$ of the reduced framework:  $\mathcal{F} \in \sigma(AF^{\mathcal{E}}(\mathcal{N}_{J}))$. Since $\mathcal{E} = \mathcal{F}^J$, by Prop.~\ref{simpler-method}, $\mathcal{F}$ will automatically be a priority extension: $\mathcal{F} \in \priex{\sigma}(AF(\mathcal{N}_{J}))$. Depending on the semantics $\sigma$, the expansion from $\mathcal{E}$ to some $\mathcal{F}$ will take one of these forms: 
\begin{description}
\item[$(\sigma = \mathit{gr})$:] apply the algorithm for the grounded extension to the priority extension $\mathcal{E}$. 
\item[$(\sigma = \mathit{co})$:] expand $\mathcal{E}$ with a fixed set of arguments built from $\mathcal{E}$ that results in a conflict-free set.  
%a set of arguments that build from $\mathcal{E}$-arguments which are conflict-free. 
Close under defended arguments. 
\item[$(\sigma = \mathit{pr})$:] proceed as with the complete semantics, but expand with a maximal set of arguments built from $\mathcal{E}$.
\item[$(\sigma = \mathit{st})$:] starting from $\mathcal{E}$, add one argument that is conflict-free to the set until the non-selected arguments are all defeated by this set.
\end{description}

Our definition of priority extension (Def.~\ref{def-priority-ext}) differs from Brewka's original reduction (see footnote~\ref{fn-brewka}). Let us motivate our version by modifying the running example once more. 

\begin{example}[Running example without the GDPR]\label{ex-problem-12}
Remove from Ex.~\ref{ex-arg-theory} norm $i_1 \Rightarrow^r_L a_1$ which prescribes complying with the GDPR. Hence, argument $A_6$, which used this norm as its top rule, no longer exists. This results in two preferred extensions, $\mathcal{E}^{+}_2$ and $\mathcal{E}^{+}_3 \setminus \{A_6\}$, which relate to Examples~\ref{ex-problem-1}--\ref{ex-problem-1b}. These extensions differ from the unique extension to the reduced framework $\mathcal{E}^+_1 \setminus \{A_6\} \in \mathit{pr}(AF^{\mathcal{E}^+_1 \setminus \{A_6\}})$.  

Our definition thus gives a priority extension, $\mathcal{E}^+_1 \setminus \{A_6\} \in \priex{pr}(AF)$, while  under Brewka's original definition (footnote~\ref{fn-brewka}) no priority extension would exist. 
\end{example}

Of course, priority extensions might not exist or might not suffice to resolve all the dilemmas in certain scenarios. In such cases, the Jiminy still has to pass a recommendation to the agent. One option is to simply not constrain the available actions and let the agent choose.\footnote{This would also be Jiminy's option if no extensions existed or if the only extension were empty. Complete and preferred semantics are less prone to these problems, except when there are odd-cycle defeats. These cycles can be broken by Jiminy arguments. As a last resort, recent argumentation semantics that do not suffer from this problem~\cite{gaggl} can be considered for the Jiminy advisor.} A more biased but constructive option is to pass the obligations of a maximal dilemma-free set of extensions. How to choose among all the extensions might depend on the agent's type or the morally sensitive situation, and so we cannot provide a definitive answer here.

\subsection{Applying the Four Levels of Dilemma Checking and Resolving.} Depending on the application domain, the Jiminy system can make use of all four levels of argumentation or it can terminate at the earliest level that is free of moral dilemmas and return the moral recommendation from that level.  

\begin{description}
\item[For time-critical applications] such as self-driving vehicles, decisions must be made as quickly as possible. The argumentation system at level $i+1$ adds arguments (or defeats) to those at level $i$, resulting in an exponential increase in the number of candidate extensions. For these applications, the Jiminy system can be implemented as an anytime algorithm: starting at level 1, it will improve its moral recommendations by reaching higher levels, as long as a prefixed deadline has not been met.     
\item[For sensitive applications,] the stakeholders might agree to any moral recommendation from the earliest level where all dilemmas are resolved. This would prevent the application of Jiminy norms to favor some stakeholders over others, unless strictly necessary (level 4). It might also prevent the combination of one stakeholder's norms with certain judgments (norms) from rival stakeholders (level 3).
\end{description}

In general, though, the highest the level of the argumentation framework, the better the moral recommendations that can be expected. Even when the moral recommendation remains the same, higher levels provide more comprehensive explanations for the output as they will include more arguments and defeats and better extensions.

Besides using norms from the Jiminy for the integrated  framework, it is also feasible to combine them with the individual  frameworks and the combined framework. The details of these combinations are omitted from this article. Finally, we end this section with the following proposition. 

\begin{proposition}
Given a set of argumentation theories $\normsys_s = (\mathcal{L}, \bar{ \hspace{0.1cm} }, \mathcal{R}_s, \mathcal{K})$ where $s\in \{s_1, \dots, s_n, J\}$ and a decision problem $DP=(\mathcal{K}, DV)$, the Jiminy will have one of the following two possible answers: there is a dilemma at level $i$, or there is no dilemma at level $i$ and all dilemmas are resolved at level $i$, where $i = 1, 2, 3, 4$.
\end{proposition}

\begin{proof}
According to Definitions~\ref{def-first-level-dilemma},  \ref{ex-comb-fr-rel}, \ref{def-third-level} and \ref{def-level4}, this proposition directly holds. 
\end{proof}

\section{Explaining a Jiminy's Choices}\label{sec:expl}
Explainability is the problem of how a human can understand the decisions made by someone else  in a given context.     Recently, methodologies, properties and approaches to explanations from the field of artificial intelligence have been widely studied~\cite{OrBiran2017}. The  ethical decisions or recommendations that a Jiminy makes are explainable.  Generating explanations for a Jiminy's choices is a feature of the argumentation approach we take to reach agreements among the stakeholders, since argumentation has ``a unique advantage in transparently explaining the procedure and the results of reasoning''~\cite[p.~1]{DBLP:conf/aaai/FanT15}.

What is an explanation? Miller~\citeyear{DBLP:journals/ai/Miller19} discusses the desirable features of an explanation from a social science point of view.  He states that explanations are {\em contrastive}  in the sense that people expect an  explanation not only about why one event happened, but (also) why another event did not happen instead.  Explanations are {\em selected} i.e., it is not expected that all the causes of an event are given; rather, a selection of one or two causes are selected for inclusion in the explanation. Truth and likelihood matter for an explanation, but full  probabilistic analysis of the event is not expected. Lastly, explanations are {\em social} in the sense that they are presented with due regard to the informational state of the person expecting an explanation.

All of the desirable aspects of explanations can be implemented in a Jiminy. In this section, we focus on the contrastive aspect and refer to the literature on making explanations selective and social.  Contrastive explanations can be obtained by considering all the available options that have been passed on to the Jiminy and comparing this set with the option the Jiminy ends up recommending. If there is a dilemma at any level, the recommendations from each of the extensions can be offered as possibilities, with an explanation as to why a particular extension survived resolution. Social explanations can be obtained by argument-based dialogues  to formalize the process of providing explanations~\cite{DBLP:journals/synthese/Walton11a,DBLP:conf/comma/CyrasST16,Cocarascu2018}.  

To explain why and how a recommendation is made by the Jiminy, we first need to identify an argument in the extension whose conclusion is the recommended decision.  Meanwhile, to explain why another decision was not recommended, we need to identify an argument in an argumentation framework whose conclusion is that other decision, and use the defeat relation between arguments to explain why an argument supporting that other decision was rejected. Please refer to the work of Liao, Anderson,
\& Anderson~\citeyear{DBLP:journals/aiethics/LiaoAA21} for details of this approach to explanation.

More specifically, regarding the recommendation that was made, when the argument supporting that decision is located by referring to the argumentation framework, it can be explained that the argument can be accepted because all of its attackers were rejected due to the fact that at least one attacker of each of its attackers was accepted, and so on. In the context of this article, whether a decision is recommended depends not only on the interaction between arguments in one or several argumentation frameworks but also on the assessed level of the recommendation, and on whether the Jiminy plays the role of ranking the stakeholders. 

Consider Figure~\ref{fig-Compatible} again. In the reduced framework $AF^{\mathcal{E}^+_1}$, the options ``comply with the law'' ($d_1$) and ``report information that grossly endangers society'' ($d_3$) are justified, while the option ``protect the privacy of users'' ($d_2$) is rejected. The explanations are as follows. 

 \begin{description}
\item[] \textbf{Explaining derivability in arguments.} ``Comply with the law'' ($d_1$) is the conclusion of argument $A_1$, which can be derived from the context ``The manufacturer makes the smart speaker'' ($w_1$) and a norm stating ``If you have manufactured a device, the behavior of that device should comply with the law'' ($w_1 \Rightarrow^r_L d_1$). ``Report information that grossly endangers society'' ($d_3$) is the conclusion of $A_3$, which can be derived from the context ``The information collected grossly endangers society'' ($w_3$) and a norm stating ``Devices that contain information about a future event that grossly endangers society should report that information to the authorities'' ($w_3 \Rightarrow^r_H d_3$). The non-recommended option ``protect the privacy of users'' ($d_2$) is the conclusion of argument $A_2$, built from the fact that the device collects data ($w_2$) and the norm $w_2 \Rightarrow^r_H d_2$. 
\item[]\textbf{Explaining justification and rejection as a dialogue by referring to an argumentation graph.}  Argument $A_1$ is accepted because it has no defeater. The defeats $A_2 \to A_1$ and $A_5 \to A_1$ are removed by applying the priority relation encoded by the norms $w_3 \Rightarrow L \succ H$ and resp.~$w_2 \Rightarrow L \succ M$ from the normative system of the Jiminy (compare the defeats in Figures~\ref{fig-OldDefeat} and  \ref{fig-NewDefeat}.) The recommendation for $d_3$ and $d_1$ is then as follows. The argument $A_3$ is accepted because its only attacker $A_2$ is rejected, and this is because $A_1$ is accepted. The explanation for not recommending $d_2$ is again based on the rejection of $A_2$ due to $A_1$.   
\end{description}

The interaction described above can be represented as a dialogue game or a discussion game. Readers may refer to the work of Vreeswijk and Prakken~\citeyear{DBLP:conf/jelia/VreeswijkP00} and Booth et al.~\citeyear{DBLP:conf/comma/0001CM18} for details. 

There is some  related work on argumentation frameworks and generating explanations in them. Fan and Toni~\citeyear{DBLP:conf/aaai/FanT15} argue that argumentation semantics are built to answer the question of which subsets of arguments are good rather than why a particular argument is good. They propose a semantics specifically for generating relevant explanations. In an argumentation graph, several arguments can fully justify the inclusion of an argument $A$ in an extension. However, sometimes just a subsection of these arguments, a so-called related extension,  is enough to justify including $A$ in the extension.  This semantics  identifies different types of explanations, all defined in terms of argument admissibility. Fan and Toni~\citeyear{DBLP:conf/aaai/FanT15}  also offer a comprehensive overview of argumentation research concerned with the problem of building explanations. Sileno et al.~\citeyear{Sileno2014} consider using answer set programming with probabilistic reasoning to implement the  generation of explanations from arguments. Liao and Van der Torre~\citeyear{DBLP:conf/comma/LiaoT20} define a semantics for explanation where each accepted argument is associated with a set of arguments that directly or indirectly defend it, thus making it explicit why the argument is accepted.

\section{The Interface Between a Jiminy and an Autonomous System}\label{interface}

How we integrate a Jiminy with an agent depends on what type of moral agent we need to construct, or rather whether the agent itself has any moral reasoning capabilities apart from the Jiminy.   Following the work of Moor~\citeyear{Moor:2006}, an artificial agent can be one of four different types of morally sensitive agent:  ethical-impact agent, implicit ethical agent, explicit ethical agent or full ethical agent. 

A \emph{full ethical agent} is one that is able to reason ethically at a human level. Clearly, no such artificial agents exist at the moment, and it is uncertain whether they can exist~\cite{Etzioni2017}. 

An \emph{ethical-impact agent} does not make any ethically sensitive decisions itself and does not necessarily operate in ethically sensitive situations. But, by virtue of replacing some human activities with the artificial agent, we change the ``moral environment'' in which the agent operates. For example, a decision aid system that assesses risks and recommends  insurance policies would not be making ethical decisions itself. However, if the data that the system uses is biased in some way, the system can propagate and even enhance this bias, thus making the world a less ethical place. 

An \emph{implicit ethical agent} does make ethically sensitive decisions or operates in an ethically sensitive context. However, the agent's actions are constrained so that unethical outcomes are avoided. Two examples of this approach are Arkin's ethical governor~\cite{Arkin09} and also the work of Dennis et al.~\citeyear{Ethics:RAS:2015}. Dyrkolbotn et al.~\citeyear{Dyrkolbotn18} further refined the definition of implicit ethical agent to specify agents who make ethically sensitive decisions without using their autonomy, regardless of the level of autonomy they have. This means that the agent does not reason about what is right or wrong but has its options externally labeled  as right or wrong and can only choose from the second set.    

An \emph{explicit ethical agent} also makes ethically sensitive decisions or operates in an ethically sensitive context. Unlike the implicit ethical agent, the explicit ethical agent is able to use its own autonomy and reasoning abilities to distinguish ethical from unethical outcomes and actions. One example of such a system is Anderson and
 Anderson's~\citeyear{AndersonA14} General Dilemma Analyzer. 

\begin{figure}[ht]
\centering
\begin{subfigure}{.47\textwidth}
  \centering
 \includegraphics[width=\textwidth]{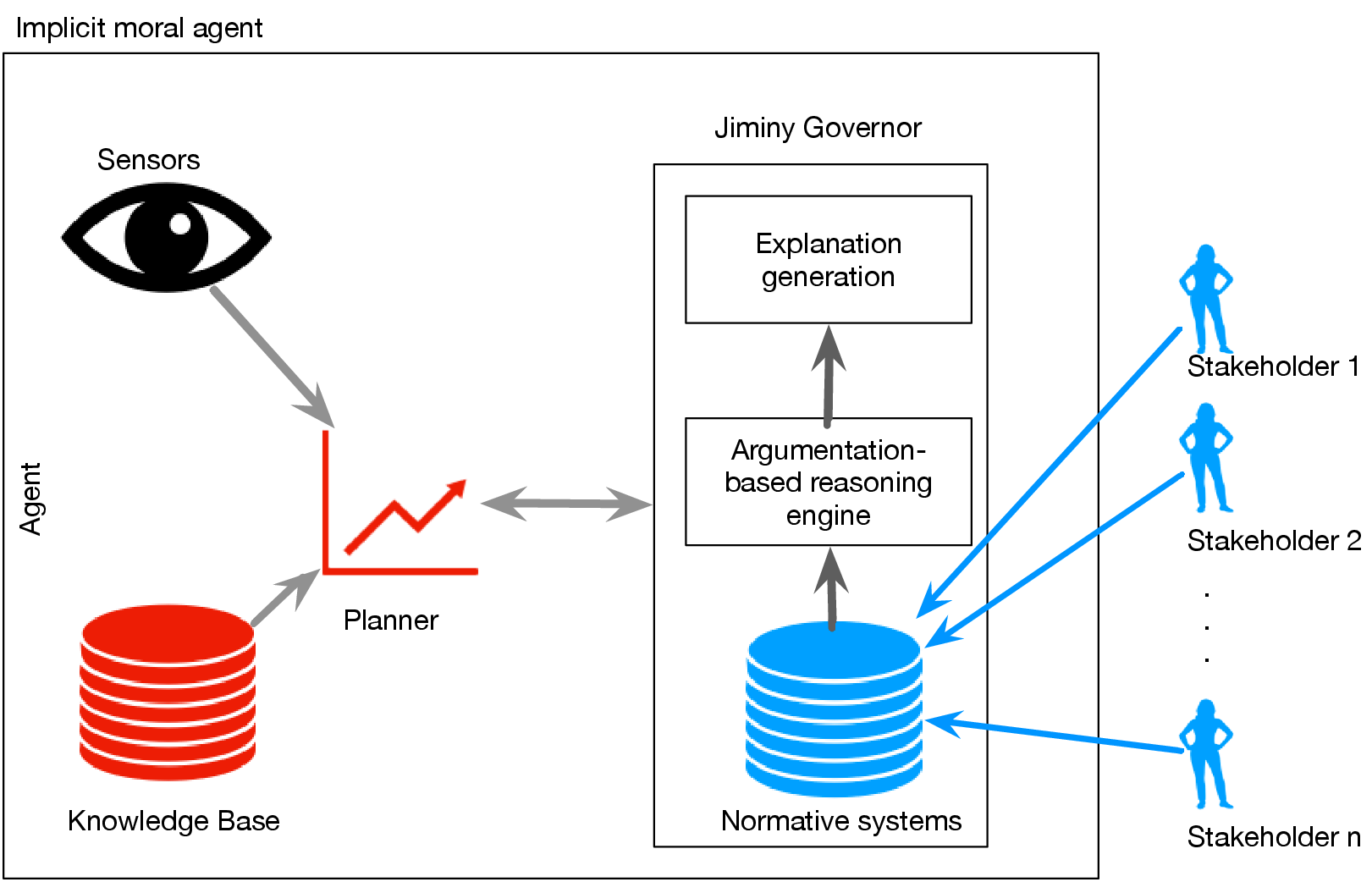}
  \caption{Implicit ethical agent obtained by using a Jiminy as an ethical governor }
    \label{fig:model-1}
\end{subfigure}%
\hspace{0.14cm}
\begin{subfigure}{.51\textwidth}
  \centering
 \includegraphics[width=\textwidth]{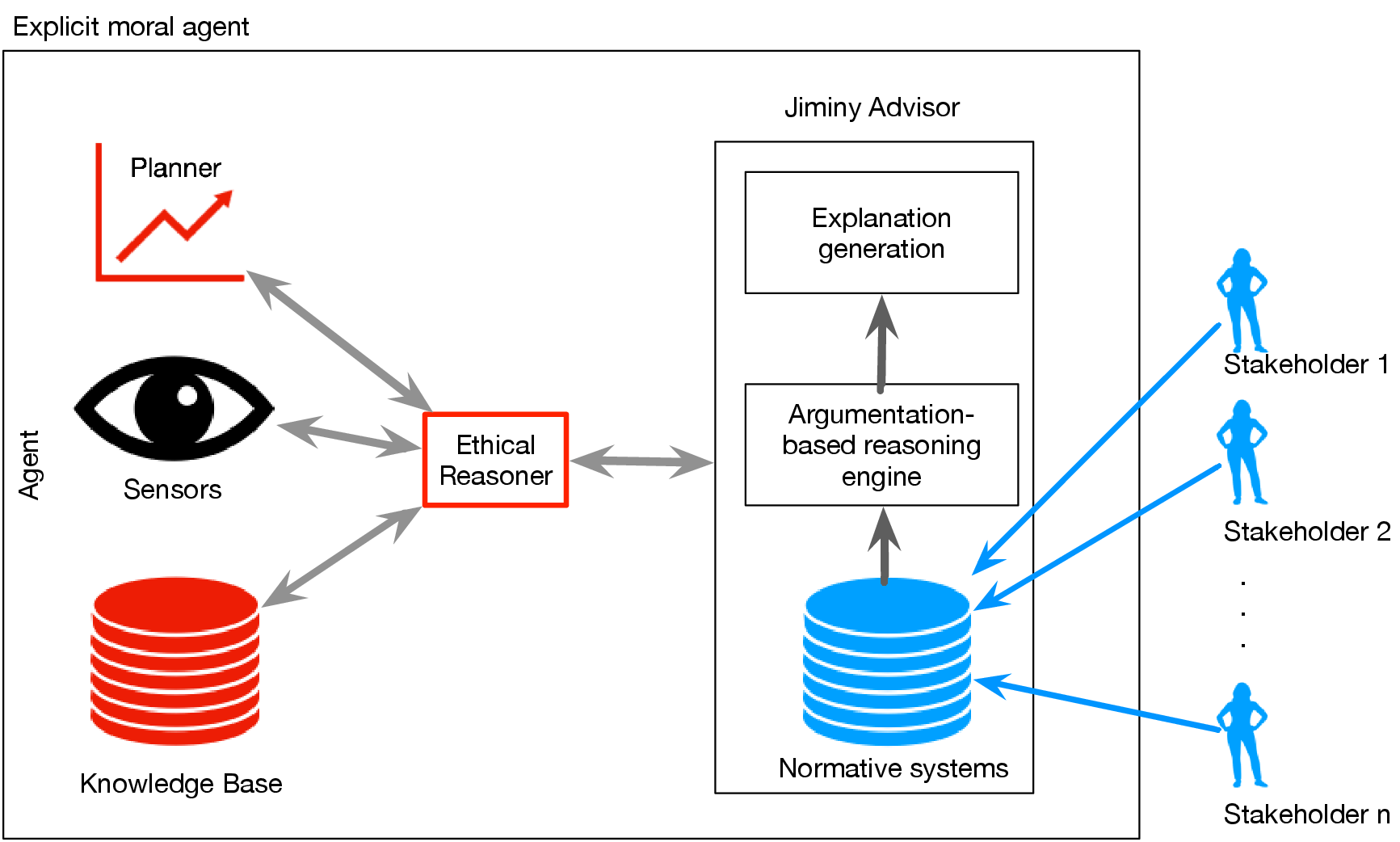}
  \caption{Explicit ethical agent  using a Jiminy as an ethical advisor} 
    \label{fig:model-2}
    \end{subfigure}
\caption{Integrating a Jiminy in an agent}
\label{fig:JiminyIntegrations}
\end{figure}

By coupling a Jiminy component with an agent that has no ethical reasoning abilities, we can create an implicit ethical agent.  
In such an integration, the  Jiminy serves as an ``external labeler'' of decisions or actions for the purpose of avoiding unethical outcomes. Effectively, the Jiminy acts as an ethical governor, constraining actions not recommended by the argumentation reasoning engine, which is based on the normative systems of the stakeholders.  Rather than having one stakeholder assess the possible actions of the agent, as in the ethical governor of Arkin~\citeyear{Arkin09},  the system automatically reaches an agreement between all the identified stakeholders about which action to take. Figure~\ref{fig:model-1} illustrates such an implicit ethical agent created by assigning the role of ethical governor to a Jiminy component. 

We assume that the agent has a knowledge base and sensors to reason about its environment, as well as a planner to identify possible actions. Each set of possible actions are communicated to a Jiminy, whose reasoning cycle is triggered only when the Jiminy identifies actions or situations involving the agent as being morally sensitive. 
 
Explicit ethical agents are able to engage in ethical reasoning and possibly also develop their own moral theories. By virtue of design, particularly if the agent learns its own moral theory, the stakeholders cannot be certain what the agent ends up treating as moral behavior.  However, for some agents, it would be important to make sure that certain ethically sensitive situations are not left entirely to the autonomous decision-making capabilities of the agent. This is where a Jiminy can be used in the role of ethical advisor, interfacing not directly with the agent's planner, knowledge base and possibly sensors but with the agent's ethical reasoning engine (see Figure~\ref{fig:model-2}). Having a Jiminy as an advisor does not change the resulting behavior of the agent in the sense that the agent remains an explicit ethical agent.  

There are (at least) two roles that a Jiminy can play as a moral advisor. The ethical reasoning engine of the agent can simply    delegate certain moral decisions to a Jiminy. This means that there are specified ethically sensitive situations in which the ethical reasoner alone makes ethical choices, and then there are other specified  situations in which the Jiminy acts as a governor and constrains some of the agent's  decisions while the ethical reasoner is not engaged.  By playing this advisory role, the agent behaves as an explicit ethical agent in some contexts, and as an implicit ethical agent in others. 

Alternatively,  the agent's ethical reasoner, in specified situations, becomes an additional stakeholder in a Jiminy, and the Jiminy constrains the actions of the agent. Now, the resulting agent remains an explicit ethical agent because it is the agent's own ethical reasoner that is always involved in the agent's ethical decision-making.   The problem of how to interface the agent with the Jiminy so as to have the ethical reasoner provide its own normative system depends much on the specific abilities of the agent, and it is outside the scope of this work at present. 

 It should be mentioned that for both advisory and governor integrations, a Jiminy never interacts directly with the environment (or users of the agent), only with the other agent components.  For reasons already discussed extensively in the literature, we  can consider the possibility of providing users with an ``off'' switch~\cite{Hadfield-Menell17} to simply disengage the Jiminy, with the result that none of the actions the agent passes on to the Jiminy will be constrained. 

Regardless of whether the Jiminy is used as an advisor or as a governor, its reasoning cycle (illustrated in Figure~\ref{fig:reason}) remains the same. In this article, we focused on specifying the subcomponents of its normative system, its argumentation reasoning engine, and its explanation generation engine. A running example was used to illustrate different aspects of these subcomponents.

\section{Related Work}\label{sec:relwork}

This section distinguishes between related research in formal argumentation about normative systems and related research in machine ethics and explainable AI. Concerning the former, in this article we use only relatively {\em abstract theories} because we believe that it is precisely this {\em generality} that makes the combination of normative systems and formal argumentation suitable for a Jiminy advisor. For a general background on these formal theories, see: the Handbook of Deontic Logic and Normative Systems~\cite{handbook:1}, in particular the chapter on moral dilemmas by Lou Goble; the Handbook of Normative Multiagent Systems~\cite{handbooknms}; the Handbook of Formal Argumentation~\cite{hofa}; and the Handbook of
Formal Argumentation~\cite{DBLP:journals/dagstuhl-manifestos/GabbayGLT18}. For an overview of the application of formal argumentation to normative systems, see the work of da Costa et al.~\citeyear{Celiaforthcoming}. Arisaka et al.~\citeyear{DBLP:journals/flap/ArisakaDST22} studied multi-agent argumentation at the abstract level, and Pigozzi and Van der Torre~\citeyear{DBLP:journals/jancl/PigozziT18} introduced a structured argumentation theory with constitutive and regulative norms. As far as we know, this article is the first in the area of structured argumentation to considers moral dilemmas emanating from multiple normative systems representing the norms of multiple stakeholders.

To position the theories of normative systems and formal argumentation within the general area of knowledge representation and reasoning, it should be observed that both theories have not only been built on the Tarskian theory of {\em deductive systems}, i.e., mathematical proof theories in deductive logic, but are also intended as criticisms of that theory. The main criticism of classical logic is its monotonicity property, and these two theories can be rephrased in the framework of nonmonotonic logic. They are typically  concerned with both theoretical reasoning and practical reasoning. 
There are many distinct versions of theories of normative systems as well as many distinct theories of formal argumentation. These knowledge representation and reasoning formalisms have been used in many disciplines. Consequently, there are relatively abstract theories that can be used across disciplines, and there are also more detailed theories developed to be used in specific disciplines because they have been adapted to the specific concerns of those disciplines. 

First, our definition of argumentation theory conforms to the abstract language used in ASPIC+~\cite{DBLP:journals/ai/ModgilP13} and some other work that extend ASPIC+,  particularly the work of Baroni et al.~\citeyear{DBLP:conf/ijcai/BaroniGL15,DBLP:journals/ai/BaroniGL18} where the contrariness function is used. However, compared to ASPIC+, the definition of argumentation theory in this article is somewhat simpler: since we assume that all norms are defeasible, we use only defeasible rules. Meanwhile, we do not deal with domain-dependent priorities over rules. However, in order to adapt to the different types of norms, we use three kinds of rules to represent institutional norms, regulative norms and permissive norms respectively. In addition, since we only use defeasible rules, the problem of the contrariness function mentioned by Baroni et al.~\citeyear{DBLP:conf/ijcai/BaroniGL15,DBLP:journals/ai/BaroniGL18} does not apply. 

Secondly, concerning the definition of a defeat relation, we only use rebut, and it is sufficient to model the conflicting relation between norms. For the priority relation over arguments, the conflicts depend on the types of the norms involved, i.e.,~two permissive norms are never in conflict, and  institutional  and permissive norms are preferred to regulative norms, and so we provide a domain-independent definition of priority over different kinds of arguments. This differs from some other work involving prioritized argumentation. For instance, Young et al.~\citeyear{DBLP:conf/atal/YoungMR16}, Liao et al.~\citeyear{DBLP:conf/deon/LiaoOTV16}, Pardo and Stra{\ss}er~\citeyear{pardo} use prioritized argumentation to represent different kinds of prioritized nonmonotonic formalisms like Reiter's~\citeyear{DBLP:journals/ai/Reiter80} default logic and Brewka and Eiter's~\citeyear{DBLP:journals/ai/BrewkaE99} Preferred Answer Sets, but they do not focus on how to represent normative reasoning in terms of different types of norms. 

Thirdly, concerning reasoning about preferences in argumentation frameworks, Modgil~\citeyear{DBLP:journals/ai/Modgil09} proposes an approach that extends Dung's theory to accommodate arguments that claim preferences among other arguments. Dung et al.~\citeyear{preferences} similarly add  rules that establish priorities between other rules. Our work on accommodating the dilemma-resolving norms to the argumentation is in line with these works. In this article, for the sake of simplicity, we did not apply the semantics of Modgil's ~\citeyear{DBLP:journals/ai/Modgil09} extended argumentation framework. Instead, we obtained the extensions of an integrated argumentation framework by using  a two-stage approach based on the approach introduced by Brewka~\citeyear{DBLP:conf/aaai/Brewka94}. 

Fourthly, there is also interesting work about exploiting argumentation to model moral reasoning. For instance, Bench-Capon and Modgil~\citeyear{DBLP:journals/ail/Bench-CaponM17} propose an approach that uses an argumentation scheme based on values. It is designed for practical reasoning, and they show how this reasoning can be used to think about situations when norms should be violated. Atkinson et al.~\citeyear{DBLP:journals/ai/AtkinsonB18} continue this line of work by presenting an approach to taking the actions of others into account that is based on argumentation schemes and value-based reasoning. We did not use argumentation schemes and value-based reasoning in our work. Instead, ASPIC+ style formal argumentation is used to model the dilemma checking and to resolve cases where a set of stakeholders have different opinions represented by their sets of norms. 

Concerning work on machine ethics, there is no consensus on whether an artificial agent can ever be as categorically a moral agent as a person~\cite{Moor:2006,Etzioni2017}.  It is widely accepted that some level of moral behavior can be implemented in machines. Wallach and Allen~\citeyear{wallachBook} distinguish between operational morality, functional morality, and full moral agency. Moor~\citeyear{Moor:2006} distinguishes between ethical-impact, explicit ethical, implicit ethical and full ethical agency; see also the work of Dyrkolbotn et al.~\citeyear{Dyrkolbotn18}. Some proposals and prototypes for implementing moral agency are already being put forward, such as those of Anderson and Anderson~\citeyear{AndersonA14}, Arkin et al.~\citeyear{ArkinUW12},  Bringsjord et al.~\citeyear{bringsjord2006toward}, Vanderelst and Winfield~\citeyear{VanderelstW16b}, Dennis et al.~\citeyear{Ethics:RAS:2015},  Lindner and Bentzen~\citeyear{LindnerB17}, and Awad et al.~\citeyear{DBLP:journals/ai/AwadAAL20}.
 
It has been shown that people consider that the same ideas of morality {\em do not} apply to   people and machines~\shortcite{Malle:2015}.  Charisi et al.~\citeyear{CharisiDFLMSSWY17} argue that the complex issue of where machine morality comes from should be addressed by considering all stakeholders---all the people who are in some way impacted by the behavior and  decisions of an autonomous system. They distinguish between 1) government  and societal regulatory bodies,  2) manufacturers and designers, and 3) end users, customers and owners.  Note that these broad categories of stakeholders can be further subdivided. For example, owners can be distinguished from ``leasers'' of the autonomous system\footnote{https://robohub.org/should-a-carebot-bring-an-alcoholic-a-drink-poll-says-it-depends-on-who-owns-the-robot/}. While it has been argued in the literature~\cite{ijcai2017-655,CharisiDFLMSSWY17} that an autonomous system should be built to integrate   moral, societal and legal values, to the best of our knowledge, no approach has been proposed that specifies how to accomplish this.  This article is the first work that explicitly considers the problem of integrating the moral values of multiple stakeholders in an artificial moral agent. 

The EU General Data Protection Regulation (GDPR), Sections~13--15, gives users affected by automated decision-making the right to obtain ``meaningful information about the logic involved, as well as the significance and the envisaged consequences of such processing for the data subject''. One way of achieving this is by building systems capable of giving arguments to support the decisions they make. Our approach provides a way to do this.   
 
Explainability has not been considered a critical feature in logic-based systems---see for example the work of Dennis et al.~\citeyear{Ethics:RAS:2015}, Lindner and Bentzen~\citeyear{LindnerB17}, and Bringsjord et al.~\citeyear {bringsjord2006toward}.  This is because one can use formal methods to prove what kinds of behavior are possible for an autonomous systems in different contexts. We argue, however, that a formal proof, while ``accessible'' to a regulatory body, is not enough to constitute explainability for common people. The GenEth system~\cite{AndersonA14}  uses  input from professional ethicists and machine learning to create a \emph{principle of ethical action preference}.  GenEth  can ``explain'' its decisions by showing how two options were compared and specifying the ethical features of each option.

\section{Summary}\label{sec:summary}

This article proposes a Jiminy advisor for autonomous agents. A Jiminy  is a multi-stakeholder ethical advisory component based on a theory of normative systems and formal argumentation. A knowledge engineer elicits the normative systems of the stakeholders, which may be viewed as tables. These are used to classify situations according to a set of ethically relevant features, and to relate these features to normative decisions. The normative systems are represented efficiently as sets of constitutive and regulative norms, including permissive norms to represent exceptions. The argumentation system is a reasoning engine dedicated to finding moral agreements.

In the initial state, no consideration is given to interaction between the normative systems of the stakeholders. Each normative system is treated independently, and the advice of all the stakeholders are compared. Where there is disagreement about the deontic decision, for example when some of the stakeholders advise alerting the police while other  stakeholders do not support this action, then we classify the situation as a moral dilemma. In such cases of moral dilemma, the argumentation engine proceeds in three steps. 

First, the argumentation engine considers the combination of all the arguments of the stakeholders. At the abstract  level, this means that attack relations between the arguments are taken into account. Instead of an argumentation framework for each stakeholder, now there is a large framework consisting of all the arguments of the stakeholders, together with the attack relations. If this leads to only one possible decision, then there is moral agreement and the Jiminy  returns that decision.

Second, where the dilemma is not resolved by combining the argumentation frameworks, then the  Jiminy  will  combine the three normative systems into a single normative system. As a consequence, there can be new arguments built from the norms of distinct stakeholders, and the combined knowledge may be sufficient to reach a moral agreement.

Third, and only when these two other methods have failed, the Jiminy  considers its stakeholder selection norms.  These meta-norms are context-dependent norms that select one stakeholder who has the most relevant expertise.  The effect of the stakeholder selection norms is to remove attacks on the arguments of the most relevant stakeholder originating from the arguments of other stakeholders.

It has often been observed that a major advantage of formal argumentation is that the reasoning process can be represented as a graph where the nodes represent abstract arguments and the edges represent abstract relations between the arguments. This approach to abstract analysis is extended by the Jiminy architecture to resolve moral dilemmas between stakeholders. In the first step, attacks are added between the arguments of stakeholders. In the second step, arguments are added to the argumentation framework. In the third step, attack relations are removed from or added to the framework. This kind of argumentation dynamics can be efficiently computed with some existing methods like that of Liao, Jin and Koons~\cite{DBLP:journals/ai/LiaoJK11}.

This abstract representation of the resolution of moral dilemmas plays a central role in the explanation module of the Jiminy advisor. Besides the logical analysis of the derivability of an institutional fact or deontic conclusion within an argument, we can use techniques from abstract argumentation such as interactive dialogue procedures.

As future work, our model of multi-stakeholder moral agreement can also be considered for other domains such as the law. In international law, each country is assumed to be autonomous, and it is assumed that there is no ranking between countries. Nevertheless, sometimes incidents can involve various countries, particularly in inheritance or contracting matters. Thus, the relation between countries is analogous to the relation between stakeholders in a Jiminy. One difference between our ethical advisor and an international law advisor is that a Jiminy has a single normative system for stakeholder selection whereas in international law, each national law contains a legal code that states what is to be done in cross-border incidents. Another question is whether existing solutions in the law can also be used to further develop the ethical advisor introduced in this paper.

\appendix
\section*{Appendix: Proofs}

This appendix contains the proofs of most of the results found in Section~\ref{sec:arg}.

\smallskip
\noindent
\textbf{Lemma~\ref{rrpp}}~(Rationality postulates)\textbf{.} \emph{The rationality postulates~\cite{caminada-amgoud} hold for the \recommendation{} $\recommender{E}$ of any $\sigma$-extension $\mathcal{E}$ under $\sigma\in \{\mathit{co}, \mathit{pr}, \mathit{gr}, \mathit{st}\}$. That is, (direct consistency) $\mathrm{Conc}(\recommender{E})$ is consistent; (subargument closure) $\mathrm{sub}(\recommender{E}) \subseteq \recommender{E}$ and, moreover, $\mathrm{sub}(\mathcal{E}) \subseteq \mathcal{E}$. The remaining postulates trivially hold.}

\begin{proof}
(Direct consistency.) Recall that any complete semantics $\sigma$ is conflict-free. Assume, towards a contradiction, that some \recommendation{} $\recommender{E} = \mathcal{E} \setminus \mathcal{A}^{p}$ of some $\mathcal{E} \in \sigma(AF)$ contains a pair $A, B\in \mathcal{E}$ such that $\mathrm{conc}(A)\in \overline{\mathrm{conc}(B)}$. By this assumption, $A$ attacks $B$ at $B$. (Case $A \succeq B$.) Then, we immediately have a direct defeat $(A, B) \in \mathit{Def}$, in contradiction with $\mathcal{E}$ being conflict-free. (Case $A \prec B$.) Then $A$ contains a subargument, namely $A$ itself, attacking $B$ at $B$ and satisfying $B \succ A$, so a reverse defeat $(A, B) \in \mathit{Def}$ exists, which is again a contradiction. (Other cases.) From $A\not\succeq B$ and $A \not\prec B$, we have that $A$ and $B$ are incomparable, but by Def.~\ref{def:prior} this is impossible given that $A, B \in \recommender{E}=\mathcal{E} \cap (\mathcal{A}^b \cup \mathcal{A}^c \cup \mathcal{A}^r)$.

\smallskip
(Subargument closure.) We first show that this postulate holds for extensions. Let $\mathcal{E} \in \sigma(AF)$ be an extension, and towards a contradiction let $A, A'$ be arguments with $A' \in \mathrm{sub}(A)$ and $A\in\mathcal{E}$ but $A' \notin \mathcal{E}$. Since $\sigma$ is a complete semantics, any argument defended by $\mathcal{E}\in \sigma(AF)$ is in $\mathcal{E}$. Hence, from this and $A' \notin \mathcal{E}$, we infer that $A'$ is not defended: ($\star$) there is a $B \in \mathit{Arg}$ defeating $A'$ such that $(C,B) \notin \mathit{Def}$ for any $C\in \mathcal{E}$. (Case 1.) Suppose $B$ attacks $A'$ first at some $A'' \in \mathrm{Sub}(A')$ with $B \succeq A''$. It follows that $B$ also attacks $A$ at $A''$ with $B\succeq A''$, and so $B$ defeats $A$. Since $A\in \mathcal{E}$ and $\mathcal{E}$ does not defeat $B$, by ($\star$) the defeat $(B,A) \in \mathit{Def}$ contradicts the initial assumption that $\mathcal{E}$ is admissible. (Case 2.) Suppose now that $A'$ contains some $A'' \in\mathrm{Sub}(A')$ attacking $B$ at $B$ and $B \succ A''$. From this and $A'\in \mathrm{Sub}(A)$, it also holds that $A$ contains $A'' \in \mathrm{Sub}(A)$ that attacks $B$ at $B$ and as before $B \succ A''$. Again $B$ defeats $A\in \mathcal{E}$ but, by ($\star$), $\mathcal{E}$ does not defeat $B$, which contradicts $\mathcal{E}$ being admissible. Note also that $\mathcal{E} \cup\{A'\}$ is conflict-free: if it was not, the proofs for cases (1)--(2) would show that neither is $\mathcal{E} \cup \{A\} = \mathcal{E}$. This concludes the proof for extensions. For the \recommendation{} $\recommender{E} \subseteq \mathcal{E}$, since both sets $\mathcal{E}$ and $(\mathcal{A}^b \cup \mathcal{A}^c \cup \mathcal{A}^r)$ are closed under subarguments, so is their intersection $\recommender{E}$.         

\smallskip
The remaining rationality postulates (indirect consistency, closure under strict rules) trivially hold for this argumentation system, as our languages feature no strict rules. 
\end{proof}

\medskip
\noindent
\textbf{Proposition~\ref{prop-inclusions}.}~\emph{Let $\normsys = (\mathcal{L}, \bar{ \hspace{0.1cm} }, \mathcal{R}, \mathcal{K})$ be an argumentation theory  with $\mathcal{R} = \mathcal{R}^c\cup\mathcal{R}^r\cup\mathcal{R}^p$. (1) For any extension $\mathcal{E}\in \sigma(AF(\normsys))$ under $\sigma\in \{\mathit{co}, \mathit{pr}, \mathit{gr}, \mathit{st}\}$ there exists $(M^c,M^r,M^p)$, a norm extension of $(\mathcal{R}^c,\mathcal{R}^r,\mathcal{R}^p)$ in context $\mathcal{K}$,  such that:}
\begin{itemize}
    \item[(i)] $I(M^c,\mathcal{K}) \supseteq \mathrm{Conc}(\mathcal{E}\cap \mathcal{A}^c) \cup \mathcal{K}$; 
    \item[(ii)] $O(M^c,M^r,\mathcal{K}) \supseteq \mathrm{Conc}(\mathcal{E}\cap \mathcal{A}^r)$; and
    \item[(iii)] $P(M^c,M^p,\mathcal{K})  \supseteq \mathrm{Conc}(\mathcal{E}\cap \mathcal{A}^p)$.
\end{itemize}
\emph{(2) For the stable case $\sigma = \mathit{st}$, the inclusions in (i)--(iii) are in fact identities: $(i')~ I(M^c,\mathcal{K}) = \mathrm{Conc}(\mathcal{E}\cap \mathcal{A}^c) \cup \mathcal{K}$;~ $(ii')~O(M^c,M^r,\mathcal{K}) = \mathrm{Conc}(\mathcal{E}\cap \mathcal{A}^r)$; and $(iii')~P(M^c,M^p,\mathcal{K}) = \mathrm{Conc}(\mathcal{E}\cap \mathcal{A}^p)$.}

\begin{proof}
(1) Let $\sigma\in \{\mathit{co}, \mathit{pr}, \mathit{gr}, \mathit{st}\}$. We prove (i)--(iii) with the fact that $\mathcal{E}\in\sigma(AF)$ implies that $\mathcal{E}$ defends itself and contains the arguments it defends. For $\tau \in \{c, r, p\}$, we define
\begin{center}
$M^\tau_{\mathcal{E}} = \begin{Bmatrix} \phi \Rightarrow^\tau \psi \in \mathcal{R}^\tau : \begin{array}{l} \exists A \in \mathcal{E} (A = A' \Rightarrow^\tau \psi \mbox{ and } \mathrm{conc}(A') = \phi) \end{array}\hspace{-.15cm}\end{Bmatrix}$.
\end{center}    

\noindent
Let $\recommender{E} \subseteq \mathcal{E}$ be the corresponding \recommendation{}. By direct consistency (Lemma~\ref{rrpp}),  $\mathrm{Conc}(\recommender{E})$ is consistent w.r.t.~$\bar{ \hspace{.1cm} }$ and, as a consequence, so is each set $\mathrm{Conc}(\mathcal{E}\cap \mathcal{A}^\tau)$ with $\tau \in \{b, c,r\}$. 
We expand each set $M^\tau_{\mathcal{E}} \subseteq \mathcal{R}^\tau$ into a set $M^\tau$, in the order $c$-\emph{then}-$p$-\emph{then}-$r$, and prove that $(M^c, M^r, M^p)$ is a norm extension.

($\tau = c$.) Starting with $M'^c = M^c_\mathcal{E}$, we keep adding to the set $M'^c$, one rule at a time, a rule $r$ from $\mathcal{R}^c \setminus M'^c$ that is triggered by $I(M'^c, \mathcal{K})$ such that $I(M'^c \cup\{r\}, \mathcal{K}) \cup \mathrm{Conc}(\recommender{E})$ is consistent. After this, we expand $M'^c$ with all the remaining untriggered rules in $\mathcal{R}^c \setminus M'^c$. This defines $M^c$. 

We check that $M^c$ is a maximal subset of $\mathcal{R}^c$ such that $I(M^c, \mathcal{K})$ is consistent w.r.t.~$\bar{ \hspace{1mm}}$ (Def.~\ref{norm-extension}). Suppose otherwise that some $r = \psi \Rightarrow^c \phi$ in $\mathcal{R}^c$ exists such that $r$ is triggered by $I(M^c, \mathcal{K})$ and $I(M^c \cup \{r\}, \mathcal{K})$ is consistent. By the construction of $M^c$, there must be some $A \in \recommender{E}$ such that $I(M^c \cup \{r\}, \mathcal{K}) \cup \{\mathrm{Conc}(A)\}$ is not consistent. By our assumption, this can only occur for some $A \in \mathcal{E} \cap \mathcal{A}^r$. Let $B = B' \Rightarrow^c \phi$ then be the argument built using $r$ over some $B' \in \mathcal{E} \cap (\mathcal{A}^b \cup \mathcal{A}^c)$. By Def.~\ref{def:prior}, we have $A \prec B$, and so $B$ defeats $A$. By admissibility, some $C\in\mathcal{E}$ defends $A$, with $C$ attacking $B$ at $B$, i.e., $\mathrm{Conc}(C) \in \overline{\phi}$. Again, by Def.~\ref{def:prior}, we must have $C \in \mathcal{E} \cap (\mathcal{A}^b \cup \mathcal{A}^c)$. But since $\mathrm{Conc}(C) \in I(M^c_{\mathcal{E}}, \mathcal{K}) \subseteq I(M^c, \mathcal{K})$, we  contradict the assumption that $I(M^c \cup \{r\}, \mathcal{K})$ was consistent.       

What remains to be shown is the inclusion in (i). Clearly, $\mathcal{K} = I_0(M^c,\mathcal{K}) \subseteq I(M^c,\mathcal{K})$.  Moreover, $I(M^c_{\mathcal{E}}, \mathcal{K})$ is consistent w.r.t. $\bar{ \hspace{.1cm} }$ and each step in the construction of $M^c$ preserves this consistency; hence, $I(M^c, \mathcal{K})$ is also consistent. Let $r = \phi \Rightarrow^c\psi$  in $\mathcal{R}^\tau \setminus M^c$ be arbitrary. Thus, $r$ is triggered by some element of $I(M^c,\mathcal{K})$, as otherwise we would have $r \in M^c$. If the addition of $r$ was consistent with both $\mathrm{Conc}(\recommender{E})$ and the consequents of $M^c$, then it would have been added to $M^c$, and so $\psi$ and some such element would be contraries. (Base case) $\psi$ is a contrary of some formula $\mathrm{Conc}(A)$ with $A \in \recommender{E}$. Then, since $r$ is triggered by $M^c$, an argument $B$ in $\mathrm{Arg}(\normsys)$ exists with $\mathrm{Conc}(B) = \phi$, and so the argument  $C= B \Rightarrow^c \psi$ is also in $\mathrm{Arg}(\normsys)$. Since $C$ defeats $A$, by admissibility some argument $D\in \mathcal{E} \cap (\mathcal{A}^b \cup \mathcal{A}^c)$ must defend $A\in\mathcal{E}$, with $D$ defeating $C$. Moreover, such a $D$ can only attack $C$ at $C$ (since $\mathrm{Conc}(\mathcal{E})$ is consistent with all of $M^c$). Finally, since the rules of $D$ are in $M^c_{\mathcal{E}} \subseteq M^c$ and the brute facts of $D$ are in $\mathcal{K}$, we conclude that  $I(M^c_{\mathcal{E}} \cup \{r\}, \mathcal{K})$ is inconsistent w.r.t.~$\bar{ \hspace{.1cm} }$, and so is $I(M^c \cup \{r\}, \mathcal{K})$. (Inductive case.) Suppose $\psi$ is a contrary of the consequent of some triggered rule $r'\in M^c$. Then immediately $I(M^c  \cup \{r\}, \mathcal{K})$ is inconsistent since $r'$ is triggered. This contradicts the construction of $M^c$. 

($\tau = p$.) We just set $M^p = \mathcal{R}^p$. Let us show (iii). Let $A = A' \Rightarrow^p \phi$ be in $\mathcal{E}$ and let $r= \psi \Rightarrow^p \phi$ be the top norm of $A$. By the construction of $A$ and the proof of Lemma~\ref{rrpp} (subargument closure holds for extensions),  $A' \in \mathcal{E} \cap (\mathcal{A}^c \cup \mathcal{A}^b)$, so we obtain that $r$ is triggered by $I(M^c_{\mathcal{E}}, \mathcal{K}) \subseteq I(M^c, \mathcal{K})$. This and $M^p = \mathcal{R}^p$ imply that $\phi \in P(M^c, M^p, \mathcal{K})$.  

($\tau = r$.) Starting with $M'^r = M^r_{\mathcal{E}}$ we add one rule $r$ at a time from $\mathcal{R}^r \setminus M'^r$ which is triggered by $I(M^c, \mathcal{K})$  whenever $I(M^c, \mathcal{K}) \cup O(M^c, M'^r \cup \{r\}, \mathcal{K})$ is consistent w.r.t.~$\bar{ \hspace{1mm} }$ and all sets $O(M^c, M'^r \cup \{r\}, \mathcal{K}) \cup \{\phi\}$ with $\phi \in P(M^c, M^p, \mathcal{K})$ are consistent w.r.t.~$\bar{ \hspace{1mm} }$. After this, $M^r$ is again defined by expanding $M'^r$ with all the remaining rules in $\mathcal{R}^r$ not triggered by $I(M^c, \mathcal{K})$. Clearly, this construction leads to a subset $M^r \subseteq \mathcal{R}$ that is maximal with the two consistency conditions from Def.~\ref{norm-extension}. Let us check the inclusion in (ii). Let $A = A' \Rightarrow^r \phi$ be in $\mathcal{E}$ and let $r = \psi \Rightarrow^r \phi$ be the top norm of $A$. As before, $A' \in \mathcal{E} \cap \mathcal{A}^c$, so $r$ is triggered by $I(M^c_{\mathcal{E}}, \mathcal{K})$. This and the fact that $r \in M^r_{\mathcal{E}} \subseteq M^r$ imply that $\phi \in O(M^c, M^r, \mathcal{K})$.  

\smallskip
\noindent
(2) For $\sigma = \mathit{st}$, we now prove that the inverse of the inclusions from $(i)$--$(iii)$ hold for stable semantics. This suffices to prove $(i')$--$(iii')$ respectively. Let $\mathcal{E} \in \mathit{st}(AF(\normsys))$ and let $M= (M^c, M^r, \mathcal{R}^p)$ be defined as above. 

$(i')$~Let us show that $I(M^c,\mathcal{K}) \subseteq \mathrm{Conc}(\mathcal{E} \cap \mathcal{A}^c) \cup \mathcal{K}$.  The proof is by induction on the construction of $M^c$. (Base case.) It is immediately apparent that $I(M^c_{\mathcal{E}}, \mathcal{K}) \subseteq \mathrm{Conc}(\mathcal{E} \cap \mathcal{A}^c) \cup \mathcal{K}$ from the definition of $M^c_{\mathcal{E}}$.
(Inductive case) Suppose that for the construction of $M^c$ so far, say as a set $M'^c$, it holds that $I(M'^c, \mathcal{K}) \subseteq \mathrm{Conc}(\mathcal{E} \cap \mathcal{A}^c) \cup \mathcal{K}$. Let $r = \phi\Rightarrow^c \psi$ be the next (triggered) rule to be added to $M'^c$. We know that $I(M'^c \cup \{r\}, \mathcal{K})$ is consistent. Since $r$ is triggered, let $A \in (\mathcal{A}^b \cup \mathcal{A}^c)$ be such that $\mathrm{Conc}(A) = \phi$, and let $B = A \Rightarrow^c \psi$. If $B \in \mathcal{E}$, we are done. Otherwise, $B \notin \mathcal{E}$ implies (by stability) that $\mathcal{E}$ defeats $B$, say with an argument $C \in \mathcal{E}$ that (by Def.~\ref{def:prior}) is also in $\mathcal{A}^b \cup \mathcal{A}^c$. But this is impossible, since by construction of $M^c$, all the triggered rules in $M^c$ are consistent with $\mathrm{Conc}(\recommender{E})$, and so such an attack cannot exist.  

$(ii')$~We now prove that $O(M^c, M^r, \mathcal{K}) \subseteq \mathrm{Conc}(\mathcal{E} \cap \mathcal{A}^r)$. Let $r = \psi \Rightarrow^c \phi$ be a rule in $M^r$ triggered by $I(M^c, \mathcal{K})$. Using this and $(i')$, we have that $\psi \in I(M^c, \mathcal{K}) \subseteq \mathrm{Conc}(\mathcal{E} \cap \mathcal{A}^c) \cup \mathcal{K}$, so let $A' \in \mathcal{E}\cap \mathcal{A}^c$ be an argument for such $\psi$. We define $A = A' \Rightarrow^r \phi$. Clearly, $A \in \mathcal{A}^r$, and if $A \in \mathcal{E}$, we are done, so assume otherwise towards a contradiction. Again $A \notin \mathcal{E}$ implies there is a $B \in \mathcal{E}$ such that $(B, A) \in \mathit{Def}(\mathcal{N})$. (Case $B \in \mathcal{A}^b \cup \mathcal{A}^c$.) Impossible, since $M^r$ was defined (for triggered rules) as a set of rules consistent with $I(M^c, \mathcal{K})$, and the latter set contains only conclusions from $\mathcal{E}$. (Case $B \in \mathcal{A}^r$.) Again, the inductive construction of $M^r$ makes this case impossible, since $B \in \recommender{E}$, and the inconsistency of the triggered $r$ with $B$ would imply that $r \notin M^r$. (Case $B \in \mathcal{A}^p$.) Let $B = B' \Rightarrow^p \phi'$ be built over some rule $r' = \psi' \Rightarrow^p \phi'$ in $\mathcal{R}^p$ with $\phi' \in \overline{\phi}$ or $\phi \in \overline{\phi'}$. By subargument closure and Def.~\ref{def:prior}, $B' \in \mathcal{E} \cap (\mathcal{A}^b \cup \mathcal{A}^c)$. Moreover, by $(i')$, $r'$ is triggered by $I(M^c, \mathcal{K})$, and so $\phi' \in P(M^c, \mathcal{R}^p, \mathcal{K})$. But this contradicts the construction of $M^r$ during the addition of $r$, since $O(M^c, M'^r \cup \{r\}, \mathcal{K}) \cup \{\phi'\}$ is not consistent for some $\phi' \in P(M^c, M^p, \mathcal{K})$.

$(iii')$~Let $\phi \in P(M^c, \mathcal{R}^p, \mathcal{K})$. We show that $\phi \in \mathrm{Conc}(\mathcal{E} \cap \mathcal{A}^p)$. As before, let $r = \psi \Rightarrow^p \phi$ be a triggered rule in $\mathcal{R}^p$, i.e., with $\psi \in I(M^c, \mathcal{K})$. By $(i')$, some $A' \in \mathcal{E} \cap (\mathcal{A}^b \cup \mathcal{A}^c)$ exists with $\mathrm{Conc}(A') = \psi$. Let then $A = A' \Rightarrow^p \phi$. By Def.~\ref{def:prior}, there can be no argument defeating $A$ at $A$, so from this and $A' \in \mathcal{E}$, we conclude that also $A \in \mathcal{E}$. Finally, $\phi \in \mathrm{Conc}(\mathcal{E} \cap \mathcal{A}^p)$.
\end{proof}

\noindent
\textbf{Proposition~\ref{symmetry}.} \emph{Let $\mathcal{N} = (\mathcal{L}, \bar{ \hspace{1mm} }, \mathcal{R}, \mathcal{K})$ be an argumentation theory with a symmetric contrariness function $\bar{ \hspace{1mm} }$. Then, any $P$-maximal norm extension $M= (M^c,M^r,\mathcal{R}^p)$ in $\mathcal{K}$ contains a $\sigma$-extension $\mathcal{E}$ satisfying the inclusions (i)--(iii) from Prop.~\ref{prop-inclusions} for any $\sigma \in \{\mathit{co}, \mathit{pr}, \mathit{gr}\}$.} 
  
\begin{proof}
Let  $\mathcal{N} = (\mathcal{L}, \bar{ \hspace{1mm} }, \mathcal{R}, \mathcal{K})$ be an argumentation theory with a symmetric function $\bar{ \hspace{1mm} }$, and let $AF =(\mathit{Arg}(\mathcal{N}), \mathit{Def}(\mathcal{N}))$ be induced by $\mathcal{N}$.  
Let also $M = (M^c, M^r, \mathcal{R}^p)$ be a $P$-maximal norm extension in $\mathcal{K}$ and let $\sigma \in \{\mathit{co}, \mathit{pr}, \mathit{gr}\}$. Define $\mathcal{N}_M = (\mathcal{L}, \bar{ \hspace{ 1mm } }, M^c \cup M^r \cup \mathcal{R}^p, \mathcal{K})$. Then it suffices to check that $\mathit{Arg}(\mathcal{N}_M)$ forms a $\mathit{co}$-extension in $AF$, namely $\mathcal{E}_M = \mathit{Arg}(\mathcal{N}_M) \subseteq \mathit{Arg}(\mathcal{N})$ (from this, minimal and maximal $\mathit{co}$-extensions will exist as well for $\sigma = \mathit{gr}$ and resp.~ $\sigma=\mathit{pr}$.) To this end, we show that  $\mathcal{E}_M \in \mathit{co}(AF)$.  

(\emph{Conflict free}.) Suppose, towards a contradiction, that $(A,B) \in \mathit{Def}(\mathcal{N})$ for some $A,B \in \mathit{Arg}(\mathcal{N}_M)$. Since the latter set is closed under subarguments, we can assume w.l.o.g.~that $A$ attacks $B$ at $B$ and that either $A \succeq B$ (direct defeat) or $A \prec B$ (reverse defeat). Note that all the rules occurring in $A, B$ are triggered by $I(M^c, \mathcal{K})$. Based on this, it can be checked that for any possible pair $\tau, \tau' \in \{b,c,r,p\}$ satisfying $A\in \mathcal{A}^\tau$ and $B \in \mathcal{A}^{\tau'}$, one of the consistency conditions for the sets $I(M^c, \mathcal{K}) \cup O(M^c, M^r, \mathcal{K})$, or $O(M^c, M^r, \mathcal{K}) \cup \{\phi\}$ for some $\phi\in P(M^c, \mathcal{R}^p, \mathcal{K})$, is violated (by the top rules of $A, B$ being in $M$). But this contradicts the initial assumption that $M$ is a norm extension. 

\smallskip
(\emph{Admissibility}.) Let $(B, A) \in \mathit{Def}(\mathcal{N})$ be such that  $B \in \mathit{Arg}(\mathcal{N}) \setminus \mathcal{E}_M$ and $A \in \mathcal{E}_M$. As before, and by the symmetry of $\bar{ \hspace{1mm} }$, we can just assume that $B$ attacks $A$ at $A$ and $B\succeq A$.  

(Case $B \in \mathcal{A}^b$.) This is impossible because $A,B  \in \recommender{E}_M$ is in contradiction with the direct consistency of $\recommender{E}_M$ from Lemma~\ref{rrpp}. 

(Case $B \in \mathcal{A}^c$.) From $B \notin \mathcal{E}_M$ and the maximality of $M^c$, there is either a triggered rule $r = \psi \Rightarrow^c \phi \in M^c$ or simply $\{\phi\} \in  \mathcal{A}^b$ such that $\phi \in \overline{\mathrm{Conc}(B)}$ (by the symmetry of $\bar{ \hspace{1mm} }$  we can ignore the opposite $\mathrm{Conc}(B) \in \overline{\phi}$). In either case, there is an argument $C = C' \Rightarrow^c \phi$ or resp.~$C= \{\phi\}$ in $\mathcal{E}_M$. In the former case, we have $C' \Rightarrow^c \phi \in \mathcal{A}^c$ and so $C \succeq B$; in the latter case, we have $C \succ B$. In any case, argument $C$ satisfies $(C,B) \in \mathit{Def}(\mathcal{N})$. 

(Case $B\in \mathcal{A}^r$.) By Def.~\ref{def:prior}, we also have that $A\in \mathcal{A}^r$. Let $B = B' \Rightarrow^r \phi$ for some $r = \psi \Rightarrow^r \phi$. By symmetry, also $A$ attacks $B$ at $B$ and $A \succeq B$. Hence, $(A,B) \in \mathit{Def}(\mathcal{N})$.

(Case $B\in \mathcal{A}^p$.) By Def.~\ref{def:prior}, $A \in \mathcal{A}^r$. And, by symmetry, let $B = B' \Rightarrow^p \phi$ be built using some rule $r = \psi \Rightarrow^p \phi$ with  $\phi \in\overline{\mathrm{Conc}(A)}$. We first show that $r$ cannot be triggered by $I(M^c,\mathcal{K})$. Suppose the contrary: then we have $B' \in \mathcal{E}_M$ and, with $r \in M^p = \mathcal{R}^p$, conclude that $\phi \in P(M^c, M^p, \mathcal{K})$. But this contradicts the assumption that $O(M^c, M^r, \mathcal{K}) \cup \{\phi\}$ is consistent for this particular $\phi$. Now, from $r$ not being triggered, we obtain that $B' \notin \mathcal{E}_M$. Hence, also some $\mathcal{R}^c$-rule occurring in $B'$ is not triggered by $I(M^c, \mathcal{K})$. Let $r' = \theta \Rightarrow^c \psi'$ be the earliest such rule in $B'$; that is, let $r'$ occur as the top rule of some subargument $C = C'  \Rightarrow^c \psi'$ of $B'$ such that $C' \in \mathcal{E}_M$. Then, by the maximality of $M^c$ with the consistency of $I(M^c, \mathcal{K})$, this set $I(M^c, \mathcal{K})$ contains (by symmetry) a contrary, say $\alpha \in \overline{\psi'}$. In other words, $M^c$ contains a triggered rule $r''$ of the form $\ldots\Rightarrow^c \alpha$, and so an argument $D = \ldots\Rightarrow^c \alpha$ exists in  $\mathcal{E}_M$ that attacks  $C$ at $C$. Since $C', D\in\mathcal{A}^c$, we have $D \succeq C$ and thus conclude that $(D,C)\in\mathit{Def}(\mathcal{N})$. 

In summary, no matter the nature of $B$, it is the case that $\mathcal{E}_M$ defends $A\in \mathcal{E}_M$.   

\smallskip
(\emph{Closure under defended arguments}.) This claim is trivial for brute fact (defended) arguments $A= \{\phi\}$, so suppose $\mathcal{E}_M$ defends an argument $A\in \mathit{Arg}(\mathcal{N})$ of the form $A = A' \Rightarrow^\tau \phi$. Towards a contradiction, suppose that $A\notin\mathcal{E}_M$. Moreover, let such an $A$ be minimal with this property: if  $A'$ (or any other subargument) is defended by $\mathcal{E}_M$, then $A' \in \mathcal{E}_M$. Since $A'$ is indeed defended by $\mathcal{E}_M$ (provided that $A$ is), we obtain that $A' \in \mathcal{E}_M$. 
Since $A \notin \mathcal{E}_M$ but $A' \in \mathcal{E}_M$, the top rule of $A$, say  $r = \ldots \Rightarrow^{\tau}\phi$, is not in $M^\tau$. And since $A' \in \mathcal{E}_M$, the rule $r$ is triggered by $I(M^c, \mathcal{K})$, and so by the maximality of $M^\tau$, there must be a rule $r' = \ldots \Rightarrow^{\tau'} \theta$ in $M^{\tau'}$ with $\theta \in \overline{\phi}$, and that rule is also triggered by $I(M^c, \mathcal{K})$. The latter implies that an argument $B = B' \Rightarrow^{\tau'} \theta$ with top rule $r'$ exists in $\mathit{Arg}(\mathcal{N})$ and is such that $B' \in \mathcal{E}_M$. Note that $B$ is a defeater of $A$ at $A$.  

(Case $\tau' = r$.) Then, a defeater $C$ of $B$ at $B$ is either a brute fact argument (in contradiction with $r'$ being in $M^{\tau'}$) or it is built with top rule $r'' \in M^c \cup M^r$, say $r'' = \ldots\Rightarrow^c \psi$ or $r'' = \ldots\Rightarrow^r \psi$ for some $\psi \in \overline{\theta}$. In either case, the membership $r'' \in M$ contradicts the assumption that $I(M^c, \mathcal{K}) \cup O(M^c, M^r, \mathcal{K})$ is consistent, given that $r' \in M^r$.  

(Case $\tau' = c$.) The proof is similar, with only the two cases: $C \in \mathcal{A}^b$, or $r''$ as a rule in $M^c$.

(Case $\tau' = p$.) A defeater $C$ of $B$ at $B$ cannot exist, contradicting that $\mathcal{E}_M$ defends $A$.

\noindent
In all cases of $\tau'$, we reached a contradiction, so $r\in M^\tau$ and, finally, $A \in \mathcal{E}_M$.
\end{proof}

We again compare a $P$-maximal norm extension $M = (M^c, M^r, \mathcal{R}^p)$ and the argumentation theory induced by it: $\mathcal{N}_M = (\mathcal{L}, \bar{ \hspace{1mm} }, M^c \cup M^r \cup \mathcal{R}^p, \mathcal{K})$.

\medskip
\noindent
\textbf{Proposition~\ref{naive-prop}.} \emph{Let $\mathcal{N} = (\mathcal{L}, \bar{ \hspace{1mm} }, \mathcal{R}, \mathcal{K})$ be an argumentation theory inducing the argumentation framework $AF = (\mathit{Arg}(\mathcal{N}),\mathit{Def}(\mathcal{N}))$. 
(1) For any naive extension $\mathcal{E} \in \mathit{na}(AF)$, it holds that $\mathcal{E} = \mathit{Arg}(\mathcal{N}_M)$ for some norm extension $M$ in $\mathcal{K}$. 
(2) For any $P$-maximal norm extension $M$ in $\mathcal{K}$, the set $\mathcal{E}_M = \mathit{Arg}(\mathcal{N}_M)$ is a naive extension: $\mathcal{E}_M
\in \mathit{na}(AF)$.}

\begin{proof}
(1) Let $\mathcal{E} \in \mathit{na}(AF)$. We define $M = (M^c, M^r, M^p)$. For each $\tau \in \{c, r, p\}$, let \begin{center}$M^\tau = \bigcup_{A\in \mathcal{E}} (\mathrm{Rules}(A) \cap \mathcal{R}^\tau) \cup \{\psi \Rightarrow^\tau \phi \in \mathcal{R}^\tau : \psi \notin \mathrm{Conc}(\mathcal{E} \cap \mathcal{A}^c)\}.$\end{center} Let us check that $M$ is a ($P$-maximal) norm extension in $\mathcal{K}$.  (Consistency.) Clearly, each of the sets $I(M^c,\mathcal{K})$ and $I(M^c,\mathcal{K}) \cup O(M^c, M^r, \mathcal{K})$ and $O(M^c, M^r, \mathcal{K}) \cup \{\phi'\}$ for some $\phi' \in P(M^c, M^p, \mathcal{K})$ is consistent, as otherwise a defeat would occur within $\mathcal{E}$, namely inside $\mathcal{E}\cap \mathcal{A}^c$ or $\mathcal{E} \cap (\mathcal{A}^c \cup \mathcal{A}^r)$ or resp.~$\mathcal{E} \cap (\mathcal{A}^r \cup \mathcal{A}^p)$.  (Maximal consistency.) Suppose towards a contradiction that some of the sets $M^\tau$ is not maximal with this property, i.e., suppose that all the above sets $I(\cdot,\cdot)$, \ldots, $O(\cdot, \cdot) \cup\{\phi'\}$~are also consistent when we add some rule $r \in \mathcal{R}^\tau$ to $M^\tau$, say $r = \psi \Rightarrow^\tau \phi$.

(Case $\tau = c$.) From $r\notin M^c$ and the definition of $M^c$, we obtain that $r$ is triggered: $\psi \in \mathrm{Conc}(\mathcal{E} \cap \mathcal{A}^c)$ and so $\psi \in I(M^c, \mathcal{K})$. Thus an argument $A' = \ldots \Rightarrow^c \psi$ or $A' = \{\psi\}$ exists in $\mathcal{E}$, from which we can build the argument $A = A'\Rightarrow^c \phi$. Since the addition of $r$ to $M^c$ preserves the consistency of the set $I(M^c \cup \{r\}, \mathcal{K}) \cup O(M^c, M^r, \mathcal{K})$, the argument $A$ neither attacks nor is attacked by any element of $\mathcal{E} \cap (\mathcal{A}^c \cup \mathcal{A}^r)$; in addition, a defeat from/to $\mathcal{E} \cap \mathcal{A}^p$ cannot exist. Thus, $\mathcal{E}\cup\{A\}$ is conflict-free, so from the assumption that $\mathcal{E}\in\mathit{na}(AF)$, we conclude that $A\in\mathcal{E}$, which now implies that $r\in M^\tau$ and thus contradicts the assumption that $r\notin M^\tau$.

(Case $\tau = r$.) The reasoning is analogous, but we now add some rule $r \in \mathcal{R}^r \setminus M^r$ and use the consistency of $I(M^c, \mathcal{K}) \cup O(M^c, M^r  \cup \{r\}, \mathcal{K})$ and all sets of the form $O(M^c, M^r  \cup \{r\}, \mathcal{K}) \cup \{\phi'\}$ to exclude any defeat involving the new argument $A = A'\Rightarrow^r \phi$ and $\mathcal{E}$.

(Case $\tau = p$.) Since permission arguments only attack obligation arguments, the proof is the same as in the previous case. Because of this, the above definition of $M^p$ implies that $M^p = \mathcal{R}^p$.

(2) We check that $\mathit{Arg}(\mathcal{N}_M)$ is a naive extension whenever $M = (M^c, M^r, \mathcal{R}^p)$ is a norm extension. Let $\mathcal{E} = \mathit{Arg}(\mathcal{N}_M)$. We prove by induction on the complexity of an arbitrary $A \in \mathit{Arg}(\mathcal{N})$ that if $\mathcal{E} \cup \{A\}$ is conflict-free, then $A \in \mathcal{E}$. (Base case.) For brute fact arguments $A$, clearly all of them are conflict-free and in $\mathcal{E}$. (Inductive case.) Assume that all subarguments $A'$ of some $A \in \mathit{Arg}(\mathcal{N})$ satisfy this rule: if $\mathcal{E} \cup \{A'\}$ is conflict-free, then $A'\in\mathcal{E}$. Now suppose that  $\mathcal{E} \cup \{A\}$ is also conflict-free. We show that $A \in \mathcal{E}$. Let $A = A' \Rightarrow^\tau \phi$ be built with $r = \psi \Rightarrow^\tau \phi'$ as its top rule. Since $\mathcal{E} \cup \{A\}$ is conflict-free, so is $\mathcal{E} \cup \{A'\}$ and so it holds that $A' \in \mathcal{E}$. Assume, towards a contradiction, that $A \notin \mathcal{E}$. Hence, since $r$ is triggered by $M$, $r\notin M^\tau$ (and so $\tau \neq p$ since $M^p = \mathcal{R}^p$). Thus, by the maximality of $M^\tau$ with the corresponding consistency condition, one of the following sets is inconsistent if we add $r$ to $M^\tau$: $I(M^c \cup \{r\}, \mathcal{K}) \cup O(M^c \cup \{r\}, M^r, \mathcal{K})$ (if $\tau = c$); or, if $\tau = r$, either $I(M^c, \mathcal{K}) \cup O(M^c, M^r \cup \{r\}, \mathcal{K})$ or $O(M^c, M^r \cup \{r\}, \mathcal{K}) \cup \{\phi'\}$ for some $\phi'\in P(M^c, \mathcal{R}^p, \mathcal{K})$. In any of these cases it can be checked that there is a defeat relation between $A$ and some $B\in \mathcal{A}^c$ or resp.~between $A$ and some $B \in \mathcal{A}^r \cup \mathcal{A}^p$. Since for any such $B$, $B \in \mathcal{E}$, we reached a contradiction with the assumption that $\mathcal{E} \cup \{A\}$ was conflict-free. 
\end{proof}

\medskip
\noindent
\textbf{Proposition~\ref{simpler-method}.} \emph{Let $\normsys_\mathcal{S} = (\mathcal{L}, \bar{ \hspace{1mm} }, \mathcal{R}, \mathcal{K})$ be an argumentation theory of a set of stakeholders $\mathcal{S}$ and let $\sigma \in\{\mathit{co},\mathit{gr},\mathit{pr},\mathit{st}\}$. For any priority extension $\mathcal{E} \in \priex{\sigma}(AF(\normsys_\mathcal{S}))$, its Jiminy fragment $\mathcal{E}^J$ is a priority extension of the Jiminy framework: $\mathcal{E}^J   \in \priex{\sigma}(AF(\normsys_j))$.}

\begin{proof} Let us abbreviate $AF(\normsys_{\mathcal{S}}) = (\mathit{Arg}(\normsys_{\mathcal{S}}), \mathit{Def}(\normsys_{\mathcal{S}}))$ as $AF  = (\mathit{Arg},\mathit{Def})$. By Def.~\ref{def:prior}, 
\begin{itemize}
\item[$(\star)$] all defeats of the arguments in $\mathit{Arg}(\normsys_j)$ are from arguments in the set $\mathit{Arg}(\normsys_j)$; in other words, $\mathit{Def}' \cap (\mathit{Arg}(\mathcal{N}_{\mathcal{S}}) \times \mathit{Arg}(\mathcal{N}_{j})) = \mathit{Def}' \cap (\mathit{Arg}(\mathcal{N}_{j}) \times \mathit{Arg}(\mathcal{N}_{j}))$ for any $\mathit{Def}' \in \{\mathit{Def},\mathit{Def}^{\mathcal{E}}, \ldots\}$.
\end{itemize}
(Case $\sigma = \mathit{co}$.) Let $\mathcal{E} \in \priex{\mathit{co}}(AF(\normsys_{\mathcal{S}}))$ and $A \in \mathit{Arg}$ be arbitrary. That is, (1) $\mathcal{E}^J \in \mathit{co}(AF(\mathcal{N}_{j}))$ and (2) $\mathcal{E} \in \mathit{co}(AF^{\mathcal{E}})$. We have to show that: 
\begin{center}
(1') $\mathcal{E}^J \in \mathit{co}(AF(\mathcal{N}_{j}))$ \qquad and \qquad (2') $\mathcal{E}^J \in \mathit{co}(AF^{\mathcal{E}^J})$. 
\end{center}
But (1') is just (1), while (2') is equivalent to $\mathcal{E}^J \in \mathit{co}(AF^{\mathcal{E}})$, so we prove the latter. Using (2), $A \in \mathcal{E}$ ~$\Leftrightarrow$~ $\mathcal{E}$ defends $A$ (in $AF^\mathcal{E}$), i.e., for all $(B, A) \in \mathit{Def}^{\mathcal{E}}$, there is $(C, B) \in \mathit{Def}^{\mathcal{E}}$. Now consider $AF^{\mathcal{E}}(\normsys_{j}) = (\mathit{Arg}(\normsys_{j}), \mathit{Def}(\normsys_{j}))$. Clearly $\mathcal{E}^J$ is conflict-free since $\mathcal{E}$ is conflict-free. It remains to be shown the above $\Leftrightarrow$-equivalence for $\mathcal{E}^J$ and an arbitrary $A \in \mathit{Arg}(\normsys_{j})$. ($\Leftarrow$)  If $\mathcal{E}^J$ defends $A$ (in $AF^{\mathcal{E}}(\mathcal{N}_j)$), then by $\mathcal{E}^J \subseteq \mathcal{E}$ and $(\star)$, it is the case that $\mathcal{E}$ also defends $A$ (in $AF^\mathcal{E}$). By (2), $A \in \mathcal{E}$. Finally, by the definition of $\mathcal{E}^J$, it also holds that $A \in \mathcal{E}^J$.  ($\Rightarrow$) Suppose that $A \in \mathcal{E}^J$. Hence $A \in \mathcal{E}$. By (2), we have that $\mathcal{E}$ defends $A$ (in $AF^{\mathcal{E}}$). By $(\star)$, we obtain that $\mathcal{E}^J$ also defends $A$.     

(Case $\sigma = \mathit{gr}$.) 
Let $\mathcal{E} \in \priex{gr}(AF)$. Thus, (1) $\mathcal{E}^J$ is $\subseteq$-minimal within $AF(\mathcal{N}_j)$ w.r.t. the properties of~(a) conflict-freeness, (b) being able to defend itself and (c) closure under defended arguments; and (2) $\mathcal{E}$ is $\subseteq$-minimal within $AF^{\mathcal{E}}$ w.r.t.~properties (a)--(c). Since the Jiminy fragment of $\mathcal{E}^J$ is $\mathcal{E}^J$ itself, it obviously satisfies properties (a)--(c) w.r.t.~the same framework $AF(\mathcal{N}_j)$. It only remains to be shown that (2') $\mathcal{E}^J$ is in $\mathit{gr}(AF^{\mathcal{E}^J}(\mathcal{N}_j)) = \mathit{gr}(AF^\mathcal{E}(\mathcal{N}_j))$. Clearly, $\mathcal{E}^J \in \mathit{co}(AF(\mathcal{N}_j))$ because: (a) $\mathcal{E}^J$ is conflict-free since $\mathcal{E}$ is conflict-free; (b) the fact that $\mathcal{E}^J$ defends itself follows from ($\star$) and the fact that $\mathcal{E}$ defends itself; (c) to see that $\mathcal{E}^J$ is closed under defended arguments, any defended argument in $\mathit{Arg}(\mathcal{N}_j)$ would also be defended by $\mathcal{E}$ given $(\star)$, hence it would belong to $\mathcal{E}$ and then $\mathcal{E}^J$. Finally suppose, towards a contradiction, that $\mathcal{E}^J$ is not $\subseteq$-minimal with these properties, so there is some $\mathcal{E}^{J-} \varsubsetneq \mathcal{E}^J$ that satisfies properties (a)--(c) as well. We then define a set $\mathcal{E}^- = \mathcal{E} \setminus \bigcup_n \mathcal{F}_n$ by removing from $\mathcal{E}$ the following inductive construction: (i) $\mathcal{F}_0 =$ the set of arguments in or built from $\mathcal{E}^J \setminus \mathcal{E}^{J-}$ and (ii) $\mathcal{F}_{n+1} =$ the set of  arguments defended only by arguments in $\bigcup_{m \leq n} \mathcal{F}_{m}$. It can  be checked that: $\mathcal{E}^-$ is a proper subset of $\mathcal{E}$  (since $\mathcal{E}^{J-} \varsubsetneq \mathcal{E}^J$), and also that $\mathcal{E}^-$ is a complete extension. This contradicts the initial assumption of (2) that $\mathcal{E} \in \mathit{gr}(AF^\mathcal{E})$.

(Case $\sigma = \mathit{pr}$.) Let $\mathcal{E} \in \priex{pr}(AF)$, so that (1) $\mathcal{E}^J \in \mathit{pr}(AF(\mathcal{N}_j))$ and (2) $\mathcal{E} \in \mathit{pr}(AF^{\mathcal{E}})$. By (1), it only remains to be proven that (2') $\mathcal{E}^J \in \mathit{pr}(AF^{\mathcal{E}^J}) = \mathit{pr}(AF^{\mathcal{E}})$. As in the previous proof for $\sigma=\mathit{gr}$, it is the case that $\mathcal{E}^J \in \mathit{co}(AF^{\mathcal{E}})$. Suppose, towards a contradiction, that $\mathcal{E}$ is not $\subseteq$-maximal with (a) conflict-freeness, (b) defending itself and (c) closure under defended arguments. It follows that there is some $A \in \mathit{Arg}(\mathcal{N}_j) \setminus \mathcal{E}^J$ such that $\mathcal{E}^J \cup \{A\}$  satisfies properties (a)--(c). We then inductively define $\mathcal{F}_0 =  \mathcal{E}^J \cup \{A\}$ and $\mathcal{F}_{n+1} = $ the set of arguments in $\mathit{Arg}$ defended  by $\bigcup_{m \leq n} \mathcal{F}_{m}$. Then the set $\mathcal{F} = \bigcup_n \mathcal{F}_m$ satisfies (a)--(c) and properly extends $\mathcal{E}$ with (at least) $A$, in contradiction with the assumption that $\mathcal{E}\in \mathit{pr}(AF^{\mathcal{E}})$.

(Case $\sigma = \mathit{st}$.) Let $\mathcal{E} \in \mathit{st}^\ast(AF)$. That is, (1) $\mathcal{E}^J \in \mathit{st}(AF(\mathcal{N}_j))$ and (2) $\mathcal{E}$ is conflict-free and for any $A \in \mathit{Arg}$, it holds that $A \in \mathit{Arg}\setminus \mathcal{E}$~$\Leftrightarrow$~$(B,A) \in \mathit{Def}^{\mathcal{E}}$ for some $B \in \mathcal{E}$. We need to prove that (1') $\mathcal{E}^J \in \mathit{st}(AF(\mathcal{N}_j))$, which is just (1), and that (2') $\mathcal{E}^J \in \mathit{st}(AF^{\mathcal{E}^J}(\mathcal{N}_j))$ or equivalently that $\mathcal{E}^J$ is also conflict-free (which is immediate) and that $\mathcal{E}^J \in \mathit{st}(AF^{\mathcal{E}^J}(\mathcal{N}_j))$. We prove a similar $\Leftrightarrow$ equivalence to the one above. ($\Rightarrow$). Suppose $A \in \mathit{Arg}(\mathcal{N}_j) \setminus \mathcal{E}^J$. By the definition of these two sets, we also have $A \in \mathit{Arg}(\mathcal{N}_j) \setminus \mathcal{E}$. By (2), $\mathcal{E}$ defeats $A$ (in $\mathit{Arg}$) and so, by $(\star)$, we now have that $\mathcal{E}^J$ also defeats $A$. ($\Leftarrow$). Now suppose that $\mathcal{E}^J$ also defeats some $A \in \mathit{Arg}(\mathcal{N}_{j})$. Clearly, $\mathcal{E}$ defeats $A$ as well, and by (2), $A\in \mathit{Arg}\setminus \mathcal{E}$. Finally, the latter fact together with $A \in \mathit{Arg}(\mathcal{N}_{j})$ and the definition of $\mathcal{E}^J$ imply that $A \notin \mathcal{E}^J$.  
\end{proof}

\acks{A shorter version of this paper was presented at the AAAI/ACM Conference on Artificial Intelligence, Ethics and Society, January 27--28, 2019. That paper presents the general idea of a Jiminy advisor, but not the details of the formal argumentation mechanism. The authors are grateful to Dr. Louise Dennis for providing some of the examples in this paper. Beishui Liao was partially supported by the `2030 Megaproject'-New Generation Artificial Intelligence of China (2018AAA0100904), the National Key Research and Development Program of China (2022YFC3340900) and the National Social Science Foundation Major Project of China (20 \& ZD047). L.~van der Torre is financially supported by Luxembourg's National Research Fund (FNR) through the project Deontic Logic for Epistemic Rights (OPEN O20/14776480) and the (Horizon 2020 funded) CHIST-ERA grant CHIST-ERA19-XAI (G.A. INTER/CHIST/19/14589586), as well as by the European Union’s Justice programme under grant agreement 101007420 (ADELE).}

\bibliography{biblio}
\bibliographystyle{theapa}
\end{document}